\documentclass[10pt,journal,compsoc]{IEEEtran}

\usepackage{times}
\usepackage{epsfig}
\usepackage{graphicx}
\usepackage{amsmath}
\usepackage{amsthm}
\usepackage{amssymb}
\usepackage{dsfont}
\usepackage[ruled,vlined]{algorithm2e}
\usepackage{algpseudocode}
\usepackage{color}
\usepackage{soul}
\usepackage{caption}
\usepackage{subcaption}
\usepackage{comment}
\usepackage{siunitx}
\usepackage{booktabs}
\usepackage[table]{xcolor}

\usepackage[pagebackref=false,breaklinks=true,letterpaper=true,colorlinks,bookmarks=false]{hyperref}
\newif\ifdraft
\drafttrue

\ifdraft

\newcommand{\VC}[1]{{\color{blue}{\bf vc: #1}}}

\else

\newcommand{\VC}[1]{{\color{green}{}}}

\fi

\newcommand{\vb}{\mathbf{b}}

\newcommand{\mB}{\mathbf{B}}

\newcommand{\mE}{\mathbf{E}}

\newcommand{\mI}{\mathbf{I}}

\newcommand{\mS}{\mathbf{S}}

\newcommand{\cD}{\mathcal D}

\newcommand{\cF}{\mathcal F}

\newcommand{\cI}{\mathcal I}

\newcommand{\cS}{\mathcal S}
\newcommand{\cT}{\mathcal T}
\newcommand{\cU}{\mathcal U}

\definecolor{gray}{RGB}{127,127,127}

\newcommand{\R}{\mathbb{R}}

\newcommand{\loss}{L}
\newcommand{\objFG}{O}
\newcommand{\objBG}{G}

\newcommand{\parag}[1]{\vspace{0.5mm}\noindent{\bf #1}}

\newcommand{\iceskating}[0]{{\bf FS-Singles}}
\newcommand{\handheld}[0]{{\bf Handheld190k}}
\newcommand{\ski}[0]{{\bf Ski-PTZ}}
\newcommand{\human}[0]{{\bf H36M}}

\begin{document}

\author{Isinsu~Katircioglu,
			Helge~Rhodin,
			Victor~Constantin,
			J\"{o}rg~Sp\"{o}rri,
			Mathieu~Salzmann,
			Pascal~Fua,~\IEEEmembership{Fellow,~IEEE}%
	\IEEEcompsocitemizethanks{
		\IEEEcompsocthanksitem I.Katircioglu, V.Constantin, M.Salzmann and P. Fua are with the Computer Vision Laboratory, \'{E}cole Polytechnique F\'{e}d\'{e}rale de Lausanne, Switzerland. E-mail: \{isinsu.katircioglu, victor.constantin, mathieu.salzmann, pascal.fua\}@epfl.ch.
	\IEEEcompsocthanksitem H.Rhodin is with the Imager Lab, The University of British Columbia, Canada. Email: rhodin@cs.ubc.ca.
	\IEEEcompsocthanksitem J.Sp\"{o}rri is with the Department of Orthopaedics, Balgrist University Hospital, University of Zurich, Switzerland. Email: Joerg.Spoerri@balgrist.ch.
}
}

\title{Self-supervised Segmentation via Background Inpainting}	

\IEEEtitleabstractindextext{

\begin{abstract}

While supervised object detection and segmentation methods achieve impressive accuracy, they generalize poorly to images whose appearance significantly differs from the data they have been trained on. To address this when annotating data is prohibitively expensive, we introduce a self-supervised detection and segmentation approach that can work with single images captured by a potentially moving camera. At the heart of our approach lies the observation that object segmentation and background reconstruction are linked tasks, and that, for structured scenes, background regions can be re-synthesized from their surroundings, whereas regions depicting the moving object cannot.
We encode this intuition into a self-supervised loss function that we exploit to train a proposal-based segmentation network. To account for the discrete nature of the proposals, we develop a Monte Carlo-based training strategy that allows the algorithm to explore the large space of object proposals. We apply our method to human detection and segmentation in images that visually depart from those of standard benchmarks and outperform  existing self-supervised methods.
\end{abstract}

}

\maketitle

\section{Introduction}

\begin{figure}
  \centering
\resizebox{1\linewidth}{!}{%
 \setlength{\tabcolsep}{1px}
  \begin{tabular}{cccc}%
  \includegraphics[width=0.253\textwidth]{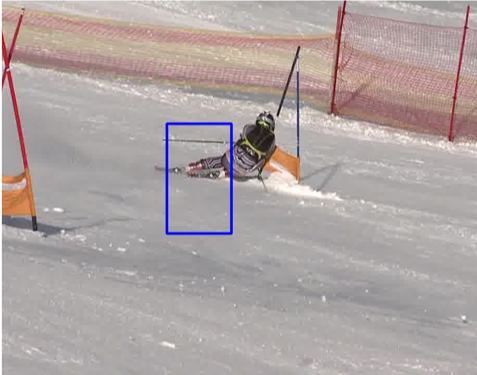} & 
  \includegraphics[width=0.253\textwidth]{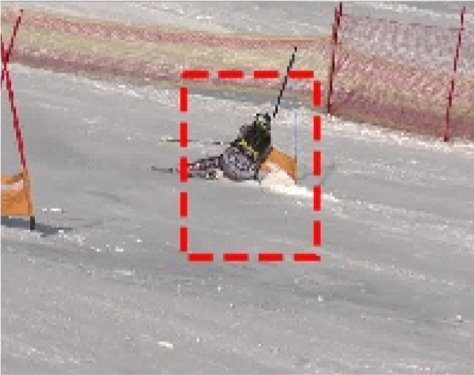} & 
  \includegraphics[width=0.262\textwidth]{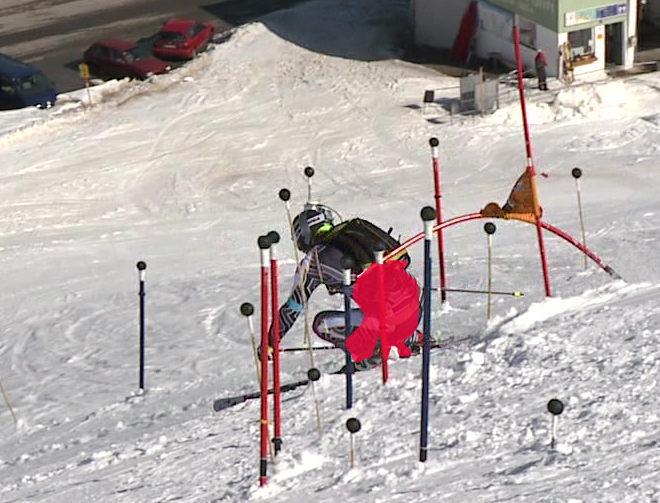} & 
  \includegraphics[width=0.262\textwidth]{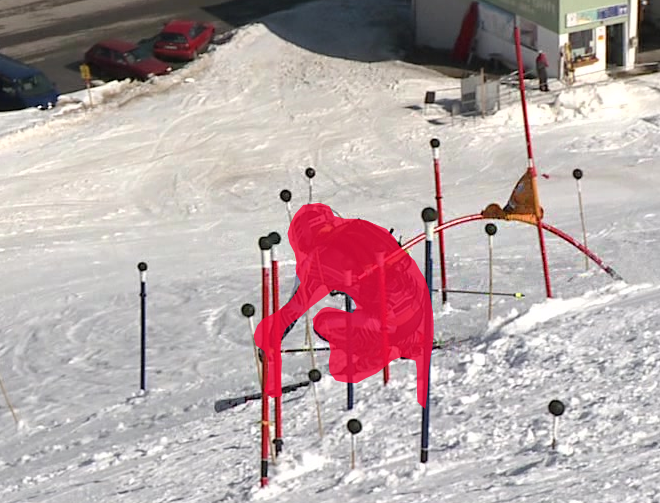} \\ 
  {\small (a) YOLOv3} & {\small (b) Ours }& {\small (c) MaskRCNN} & {\small (d) Ours}
  \end{tabular}
}

  \caption{\textbf{Domain specific detection and segmentation.} 
  Our self-supervised method detects the skier well, while YOLO trained on a general dataset does not generalize to this challenging domain. 
  Similarly, MaskRCNN trained on a general dataset sometimes misses body parts such as the upper body of the skier in (c).
  }
  \label{fig:teaser}
\end{figure}

Object detection and segmentation methods now deliver impressive precision and recall rates when trained and tested on large annotated datasets~\cite{Lin14a}. However, they perform less well in situations for which they have not been specifically trained. For example, the popular MaskRCNN network~\cite{He17a} that has been trained on the large MS-COCO database~\cite{Lin14a} often performs very well. However, it can easily fail in unusual scenarios such as the one depicted by Fig.~\ref{fig:teaser}, which features a slalom skier in a position rarely encountered in the MS-COCO data. 

Therefore, weakly- and self-supervised detection and segmentation of salient foreground objects in complex scenes have recently gained attention~\cite{Bielski19,Crawford19,Croitoru19,Eslami16,Rhodin19a}. These methods promise effortless processing of community videos with little human intervention. However, a close look at these techniques reveals that they make strong assumptions, such as the target objects being seen against a static background, rely on pre-trained supervised object localization and object-boundary detection, or can only operate on video streams. %

To develop a more generic approach, we start from the observation that in most images the background forms a consistent, natural scene. Therefore, the appearance of any background patch can be predicted from its surroundings. By contrast, a moving object's appearance is unpredictable from the neighboring scene content and can be expected to be very different from what an inpainting algorithm would produce. We incorporate this insight into a proposal-generating deep network whose architecture is inspired by those of YOLO~\cite{Redmon16} and MaskRCNN~\cite{He17a} but does not require explicit supervision.

For each proposal, we synthesize a background image by masking out the corresponding region and inpainting it from the rest of the image. The loss function we minimize favors the largest possible distance between this reconstructed background and the input image. This encourages the network to select regions that cannot be explained from their surrounding and are therefore salient. To handle the discrete nature of the proposals, we develop a Monte Carlo-based strategy to train our network. It operates on a discrete distribution, is unbiased, exhibits low variance, and is end-to-end trainable.

We focus on the well-defined, yet challenging and widely studied task of detecting and segmenting a person in single images. Our approach overcomes limitations in existing self-supervised human pose estimation methods requiring static cameras~\cite{Rhodin18a, Rhodin19a} or monochromatic background~\cite{Crawford19}.
We demonstrate the effectiveness of our unsupervised method on several human-tracking datasets captured with cameras that are static, pan-tilt-zoom, or hand-held. We focus on images acquired in 
realistic conditions, different from those of standard benchmarks, without requiring \emph{any} manual annotation. As shown in Fig.~\ref{fig:teaser}, in such situations with unusual activities, our method approaches the quality and sometimes 
outperforms the state-of-the-art detectors and instance segmentation methods that have been trained on large annotated datasets. Furthermore, it outperforms the existing self-supervised segmentation techniques tailored for detecting small motion in short videos. Our experiments also evidence that our approach not only enables reliable human segmentation in videos with %
large camera motion, but also generalizes to segmenting animals and moving rigid objects, as long as the target object moves by at least one object width in the training video.
In particular, we demonstrate the superiority of our method over MaskRCNN on videos depicting objects that look significantly different from the ones in MS-COCO.
We will make our code and ground truth segmentations for the existing \ski{} and the new \handheld{} and \iceskating{} datasets publicly available upon acceptance of the paper.

\section{Related Work}
\label{sec:related}

Most salient object detection and segmentation algorithms are fully-supervised~\cite{Cheng17a,He17a,Redmon16,Song18} and require large annotated datasets with paired images and labels. Our goal is to train a purely self-supervised method without either segmentation or object bounding box annotations. Note that this differs from the so-called \emph{unsupervised object segmentation} methods~\cite{Perazzi15,Hu18b,Jain17,Li18g,Li18j,Lu19,Yang19b}, that require domain-specific annotations during training but not at test time, or 
the label of the first frame at inference time~\cite{Wang19g}. We focus our discussion on self- and  weakly-supervised methods with regard to the type of training data used and refer to~\cite{Koh17b} for a complete discussion of methods using hand-crafted optimization.

\parag{Weakly-supervised methods.}
An early weakly-supervised method is the Hough Matching algorithm~\cite{Cho15}. It uses an object classification dataset and identifies foreground as the image regions that have re-occurring Hough features within images of the same class.
Similar principles have been followed to train deep networks for object detection~\cite{Jain17,Wei17}, optical flow estimation~\cite{Tokmakov17b,Tokmakov17a}, and object saliency~\cite{Li18g}.
These methods make the implicit assumption that the background varies across the examples and can therefore be excluded as noise. This assumption is violated when training on domain-specific images, where foreground and background are similar across the examples.

\parag{Motion-based methods.}
Conventional methods~\cite{Lee11,Papazoglou13,Factor14,Wang15d,Koh17b} explore the motion information mainly by resorting to hand-crafted features.~\cite{Wang15d} proposed a spatial-temporal energy function applied to optical flow field to obtain spatiotemporally consistent saliency maps that are further improved by using global appearance and location models. Similarly~\cite{Papazoglou13} computed the optical flow  to detect motion boundaries and refines them through ray-casting strategy. An alternative temporal solution relies on the recurrence property of the primary object in a video~\cite{Koh17b}. It finds the recurring candidate regions in the entire sequence by extracting color and motion cues through ultrametric contour maps. Identifying the matching segment tracks in different frames is done by minimizing a chi-square distance temporally in the feature space.
Given video sequences, the temporal information can be exploited by assuming that the background changes slowly~\cite{Barnich11} or linearly~\cite{Stretcu15}. However, even a static scene induces non-homogeneous deformations under camera translation, and it can be difficult to handle all types of camera motion within a single video, and to distinguish articulated human motion from background motion~\cite{Russell14}.
Some of the resulting errors can be corrected by iteratively refining the crude background subtraction results of~\cite{Stretcu15} with an ensemble of student and teacher networks~\cite{Croitoru19}. This, however, induces a strong dependence on the teacher used for bootstrapping. Recently, ~\cite{Lu20} showed that leveraging the temporal information at different granularities through forward-backward patch tracking and cross-frame semantic matching can be used to learn video object patterns from unlabeled videos. Note that these methods can only operate on video streams and exploit a strong temporal dependency, which our model doesn't.

Our approach is conceptually related to VideoPCA~\cite{Stretcu15}, which models the background as the part of the scene that can be explained by a low-dimensional linear basis. This implicitly assumes that the foreground is harder to model than the background and can therefore be separated as the non-linear residual. Here, instead of using motion cues, we propose to rely on the predictability of image patches from their spatial neighborhood using deep neural networks. This gives us an advantage over VideoPCA that only works with videos and comparably little background motion and complexity. Another closely related work~\cite{Yang19c} employs a similar inpainting network to ours on flow fields. It relies on an adversarial model that tries to hallucinate the optical flow from its surrounding while generating the mask of a supposedly moving object in the region where the inpainting network yields poor reconstruction. ~\cite{Yang19c} is based on PWC network~\cite{Sun18a} that is trained with supervision on a large object database to predict flow with clear object boundaries. Methods based on deep optical flow are not strictly self-supervised and can suffer from degenerate cases when applied to still images with no or little movement.

\parag{Self-supervised methods.}
Most similar to our approach are the self-supervised ones to object detection~\cite{Bielski19,Crawford19,Eslami16,Rhodin19a} that complement auto-encoder networks by an attention mechanism. These methods first detect one or several bounding boxes, whose content is extracted using a spatial transformer~\cite{Jaderberg15}. This content is then passed through an auto-encoder and re-composited with a background. In~\cite{Rhodin19a}, the background is assumed to be static and in~\cite{Crawford19,Eslami16} even single colored, a severe restriction in practice. \cite{Crawford19} uses a proposal-based network similar to ours, but resorts to approximating the proposal distribution with a continuous one to make the model differentiable. Here, we demonstrate that much simpler importance sampling is sufficient.~\cite{Pathak17} uses the noisy segmentation masks predicted by an unsupervised version~\cite{Factor14} as pseudo labels to train a ConvNet to segment moving objects from single images.  \cite{Bielski19} uses a generative model relying on the assumption that the image region strictly covering the salient object can be subject to random shifts without affecting the realism of the scene. Similarly,~\cite{Chen19a} relies on an adversarial network whose generator extracts the object mask and redraws the object by assigning different color or texture features to that region. By contrast,~\cite{Arandjelovic19} aims to discover the foreground object by compositing it into another image so that the discriminator fails to classify the resulting image as fake. These methods can easily be deceived by other background objects whose random displacement or texture change can still yield realistic images. In contrast to all of these techniques, our approach works with images acquired using a moving camera and with an arbitrary background.

In addition to object detection, the algorithm of~\cite{Rhodin19a} also returns instance segmentation masks by reasoning about the extent and depth ordering of multiple people in a multi-camera scene. However, this requires multiple static cameras and a static background at training time, as does the approach of~\cite{Baque17b} that performs instance segmentation in crowded scenes.

\section{Method}

Our goal is to learn a salient person detector and segmentor from unlabeled videos acquired in as practical a setup as possible. We therefore require only bare videos or images as input and do not impose restrictions on the camera motion between frames.

\begin{figure*}[t]
	\begin{center}
		\includegraphics[width=0.8\linewidth]{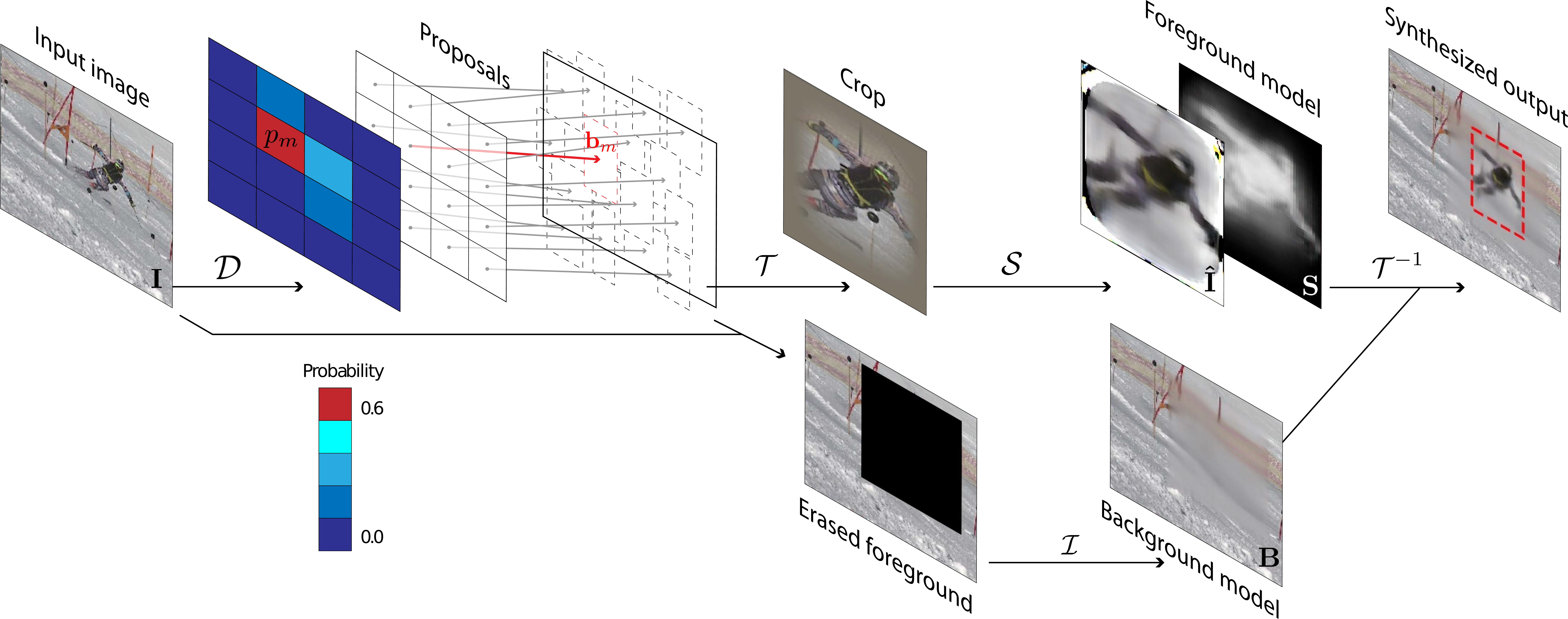}
	\end{center}

	\caption{\textbf{Method overview.} Encoder-decoder network ($\cS$) with an attention mechanism defined by proposal-based detection ($\cD$) and spatial transformers ($\cT,\cT^{-1})$. Via the use of an inpainting network ($\cI$), our approach makes it possible to train this network in an entirely self-supervised manner on unknown scenes with a moving background and captured with a hand-held camera.
	}
	\label{fig:pipeline}
\end{figure*}

Our algorithm takes a single image $\mI \in \R^{W \times H \times 3}$ as input and outputs a bounding box $\vb_m \in \R^{4}$, expressed as a center location, width, and height, and a segmentation mask $\mS \in \R^{128 \times 128}$ within that window. This is achieved by the end-to-end training of a network comprising detection and segmentation blocks depicted by Fig.~\ref{fig:pipeline}. Our detection network architecture follows that of YOLO~\cite{Redmon16} and the segmentation network has an encoder-decoder structure. Our model first predicts a set of $C$ candidate object locations $(\vb_{c})^C_{c=1}$ and corresponding probabilities $(p_{c})^C_{c=1}$. To this end, we use a fully-convolutional architecture that divides the image into a grid, with each cell $c$ yielding one probability, $p_c$, and an offset from the cell center to compute $\vb_c$.
 Then, the $\vb_{m}$ with highest probability $p_m \geq p_c, \; \forall c$, is chosen and its content is decoded into foreground $\hat{\mI} \in \R^{128 \times 128}$, segmentation mask $\mS$, and background $\mB \in \R^{W \times H}$ with separate
encoder-decoder branches.

In contrast to most prior work, we derive a probabilistic formulation in which not a single, but multiple candidates can give rise to plausible reconstructions. This has turned out to be crucial to succeed on images with unknown and dynamic backgrounds.
However, in the probabilistic case, self-supervised training requires to develop dedicated optimization techniques because the ground-truth candidate is unknown and gradients need to be backpropagated through the sampling from the discrete proposal distribution for end-to-end training. In the following two sections, we explain step-by-step our solution to this problem.
\vspace{-0.5mm}
\subsection{Self-Supervised Training with Known Background}
\label{sec:knownbg}
Let us first consider the simplified case of finding objects against a known background, e.g., obtained by background subtraction when the camera is static. This case has already been addressed by \cite{Crawford19,Rhodin19a}, but we propose an alternative that yields better results.
The different steps in our approach are depicted by Fig.~\ref{fig:pipeline}. Given a set of unlabeled images $(\mI_i)^N_{i=1}$,
our goal is to train a neural network $\cD(\mI) \mapsto ((\vb_{c}), (p_{c}))^C_{c=1}$ to produce suitable bounding box candidates $\vb_{c}$ and to assign to each one a probability $p_{c}$ to contain the object. We introduce a reconstruction objective, $\loss\left(\cF(\mI, \vb_c, \mB), \mI\right)$, that measures
 how well the autoencoder $\cF$ reproduces the input image $\mI$ on top of a background $\mB$, with attention on the bounding box $\vb_c$.
Because $\vb_c$ selects a part of the image for decoding, this loss encourages $\cD$ to focus on the object so as to model the foreground in front of $\mB$.  Moreover, the autoencoder only approximates the actual image, forcing the segmentation mask not to contain the parts already captured by the background.
 
As in \cite{Crawford19,Rhodin19a}, we implement a spatial transformer $\cT$ that crops the area of interest, followed by a bottle-neck autoencoder $\cS$ that produces the foreground and segmentation mask, and a second spatial transformer $\cT^{-1}$ that undoes the cropping and blends the synthesized foreground and background. Thus, we write
 \begin{equation}
 \begin{aligned}
 \cF(\mI,\vb_c, \mB) = \cT^{-1}(\hat{\mI}) \circ \cT^{-1}(\mS) + \mB \circ (1-(\cT^{-1}(\mS)), \\ 
 \text{ with } (\hat{\mI},\mS) =  \cS(\cT(\mI,\vb_c)),
 \end{aligned}
 \end{equation}
where $\circ$ is the elementwise multiplication.
The loss $\loss$ can then be a least-square distance between all pixel values, and we discuss alternatives at the end of the section.

Because we have a finite set of candidates, the reconstruction objective can be expressed deterministically as a weighted sum over all candidates as
\begin{equation}
\objFG(\mI) = \sum_{c=1}^{C} p(c) \loss\left(\cF(\mI,\vb_c, \mB) , \mI\right).
\end{equation}
Note that minimizing this objective will jointly optimize the detection network $\cD$ that generates proposals, and the synthesis network $\cS$ that models the object appearance and the segmentation mask.
This objective is an explicit function of $p$ and could be optimized by gradient descent.
However, summing over $C$ is inefficient when $C$ is large. For example, in our experiments $C=64$ and the resulting batches do not fit in GPU memory.
Our goal now is to overcome this limitation by incorporating the sampling of candidates in the
end-to-end training process.
We start by reasoning about the expected loss across all likely candidates, which can be written as
\begin{equation}
\objFG(\mI) = \mE_c \left[\loss\left(\cF(\mI,\vb_c, \mB) , \mI\right) \right], \text{ with} \ c \sim p,\;  \text{where} \ p(c) = p_c,
\label{eq:expected objective}
\end{equation}
and $\mE_c$ denotes the expectation over $c$ drawn from the categorical proposal distribution output by the network $\cD(\mI)$.

This probabilistic approach was also analyzed by~\cite{Crawford19}, by resorting to using a continuous approximation of the discrete distribution $p$ to facilitate end-to-end training. We propose a simpler alternative to compute the expected loss, exploiting Monte Carlo and importance sampling. We defer the mathematical details of this approach to Section~\ref{section:mc}.

\subsection{Self-Supervised Training with Moving Background}
\label{sec:separation}

Having derived an efficient training scheme for proposal-based segmentation when $\mB$ is given, we now turn to the moving camera scenario, where $\mB$ is generally unknown.
This more difficult case could be reduced to the former if $\mB$ could be predicted.
This, however, appears even harder than foreground segmentation. %

To overcome this difficulty, we redefine the training strategy and replace background prediction with the simpler task of inpainting local regions instead of the entire background $\mB$. This can easily be trained by removing a region and inpainting it from its immediate surrounding~\cite{Pathak16,Yu2018}. 
We therefore cast foreground segmentation as the task of finding the area $\vb_c$ that, when inpainted, yields the largest image reconstruction error.
Using an off-the-shelf inpainting model for this purpose led to hallucinated objects. Therefore, we train the inpainting network $\cI$ in a self-supervised manner on the available input videos by randomly selecting and cutting the area to be inpainted. Under the assumption that the foreground object moves, the network $\cI$ would not manage to reconstruct the foreground objects if fully removed in the input because the surrounding background gives no cues of their presence. This requires the following assumptions:
\begin{itemize}
\item{\textbf{Assumption A1:}} \textbf{Assumption A1:}Every part of the background must be uncovered more often than covered. In contrast to related works that require motion in every frame~\cite{Tokmakov17b,Tokmakov17a,Koh17b,Hu18b,Yang19c}, this assumption is naturally fulfilled in long videos depicting moving people.
\item{\textbf{Assumption A2:}} \textbf{Assumption A2:}The foreground and background are distinguishable in color or texture. As discussed in Section~\ref{section:extension}, this can be relaxed by using optical flow.
\end{itemize}

Instead of performing an expensive search at inference time, we train $\cD$ to predict the foreground location. We search probabilistically and introduce the background objective
\begin{equation}
\objBG(\mI) = \mE_c \left[-\frac{\loss(\cI(\mI, \vb_c),  \mI)}{a(\vb_c)} \right], \text{ with } c \sim p,
\label{eq:background objective}
\end{equation}
where $\cI$ takes the image $\mI$ and the region $\vb_c$ to inpaint as input, $\loss$ is the same as the loss in the foreground objective in Eq.~\eqref{eq:expected objective},
$a(\vb_c)$ normalizes the loss by the window area, and $p$ and $\vb_c$ are computed using $\cD$ as before.
As detailed in Section~\ref{section:mc}, we use an importance sampling strategy to optimize the discrete distribution. Note the negative sign compared to the foreground objective. This negative reconstruction error favors selecting the regions where the true image is dissimilar to the reconstructed background when minimizing Eq.~\eqref{eq:background objective}.

\parag{Disentangled training strategy.} The objective $\objBG$ has trivial solutions. It favors locations with high error density, irrespectively of their size, as illustrated by Fig.~\ref{fig:h36m ablation}(c). Alternatively, removing the normalization by the area favors erasing extensively large regions, whether they contain an object or not, because a larger number of reconstructed pixels leads to a higher inpainting error, as shown in Fig.~\ref{fig:h36m ablation}(b).

To eliminate these degenerate cases, we combine the new background objective $\objBG$ with the
foreground objective $\objFG$ of Eq.~\eqref{eq:expected objective}, substituting the known $\mB$ with the learned inpainting $\cI(\mI,\vb_c)$. The reason behind this is that these two terms are complementary:
While $\objBG$ prefers locations that cover the object neither precisely nor entirely, $\objFG$ favors a tight fit over partial coverage but has a trivial solution when $\vb_c$ is on a background region, i.e., not covering the object and having nothing to encode.

Finding a balance between these two adversarial objectives by relative weighting has turned out to be difficult, if not impossible.
Instead, we separate them such that their influence on the individual network components is mutually exclusive.
The probabilities $p_c$ are only optimized according to $\objBG$.
Therefore, $\objFG$ cannot impose a bias to the background regions, where it has a trivial solution.
In turn, $\vb_c$ is optimized according to $\objFG$ to find a tight fit without the bias of $\objBG$ to excessively large or small proposals $\vb_c$. Furthermore, $\cS$ is solely optimized according to $\objFG$ to output the best possible reconstruction, instead of the largest distance to the background as induced by $\objBG$. This can all be computed in a single forward-backward pass by treating the excluded variables as constants in the respective objectives, that is, by
cutting their gradient flow.

Note that stable training through effective separation into coarse and fine localization 
is only possible with the chosen proposal-based detection framework;
not with direct regression. %

\subsection{Monte Carlo (MC) Sampling} 
\label{section:mc}
To be efficient, the expectation in Eq.~\eqref{eq:expected objective} could be estimated by sampling a small set of $J$ candidates from $p$.
Unfortunately, sampling from such a discrete distribution is not differentiable with respect to its parameters, which would preclude end-to-end gradient-based optimization.
Instead, we rewrite Eq.~\eqref{eq:expected objective} using an arbitrary distribution $q$, by reweighting with the quotient of both distributions. That is,
\begin{align}
\objFG(\mI)
&= \mE_c\left[ \frac{p(c)}{q(c)} \loss\left(\cF(\mI,\vb_c, \mB) , \mI\right) \right], \text{ with } c \sim q \;.
\label{eq:expected objective importance}
\end{align}
This change of distribution and relative weighting hold for any two probability distributions, as explained in the appendix.
We optimize Eq.~\eqref{eq:expected objective} in Section~\ref{sec:knownbg} using stochastic gradient descent and mini-batches of $(\mI_i)^N_{i=1}$ where $N = 16$ with a single sample drawn from $q$ per image.

\parag{Importance sampling.} While moving the distribution into the expectation sum provides differentiability, Monte Carlo sampling comes at the cost of a potentially large variance, i.e., high approximation error for a few samples. For instance, by choosing $q$ to be the uniform sampling distribution $\cU$, most of the uniformly drawn samples will have a low probability of $p$ and, therefore, negligible influence.
To reduce this variance, we leverage importance sampling which provides a low-variance unbiased estimator, and set the sampling distribution $q$ to be similar or equivalent to $p$.
To prevent division by very small values that could lead to numerical instability, we define the new distribution of $q$ as
\begin{equation}\label{eq:epsilon}
q(c) = {p(c)(1-C\epsilon) + \epsilon}\;.
\end{equation}
As a side effect, $\epsilon$ controls the probability that an unlikely case is chosen, which induces a form of exploration that is helpful in the early training stages of the network.
Note that the fraction $\frac{p(c)}{q(c)}$ cancels numerically for $q(c) = p(c)$.
However, while values cancel, their derivatives do not. The differentiability of $p$ is maintained as
the sampling distribution $q$ must be treated as a constant.
In this case, the gradient of Eq.~\eqref{eq:expected objective importance} equals that of the likelihood ratio method~\cite{Glynn90} used in the REINFORCE algorithm~\cite{Williams92}.
We favor the MC interpretation that explains the effect of the gradient quotient as importance sampling and generalizes to arbitrary distributions $q$. While the relation has been known for a long time~\cite{Glynn90}, it has been overlooked in the recent literature.

\subsection{Extensions to the Base Model}
\label{section:extension}
\parag{Regularization.}
To further boost performance,
we introduce an $\loss_2$ prior on the probability output $p$ of the detection network $\cD$, and a v-shaped prior $\loss_v$ on $\mS$. The former promotes a large margin between the probability values of the bounding box candidates by producing a significantly high probability value for the candidate box with the foreground object and suppressing the rest. The latter stabilizes the early training iterations by encouraging the average value of the segmentation mask at the beginning of training to be larger than a value $\lambda$, yet sparse when exceeding this threshold. Thus, we write
\begin{equation}
\loss_v = \left|\left(\frac{1}{WH}\sum^{W}_{x}\sum^{H}_{y}{\cT^{-1}(\mS)}_{xy} \right) - \lambda\right|+ \lambda \;,
\label{eq:v_prior}
\end{equation}
where $W$ and $H$ are the image width and height respectively and $\lambda$ is set to 0.005.

\begin{figure}
  \centering{
  \begin{tabular}{@{}ccc@{}}
   
\includegraphics[width=0.16\textwidth]{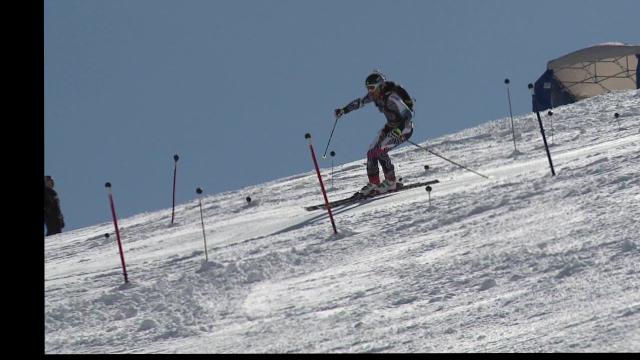} &\hspace{-4mm}
\includegraphics[width=0.16\textwidth]{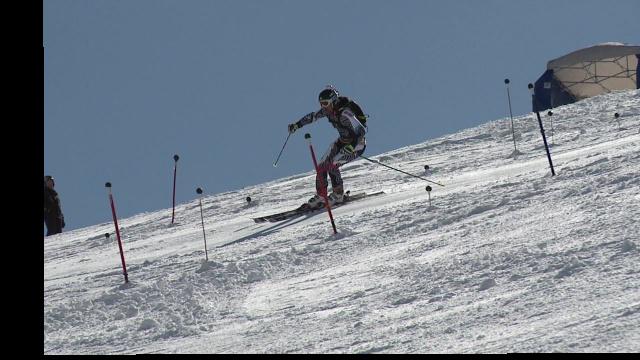} &\hspace{-4mm}
  \includegraphics[width=0.16\textwidth]{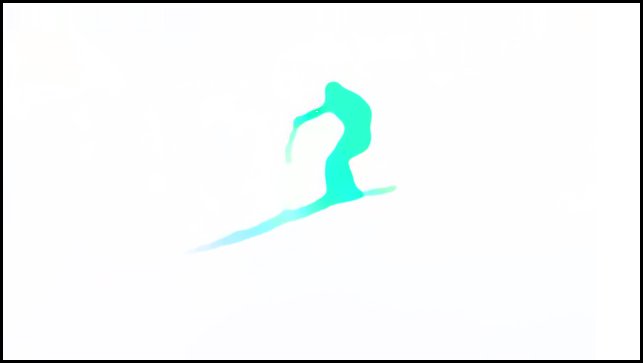} \\
  \includegraphics[width=0.16\textwidth]{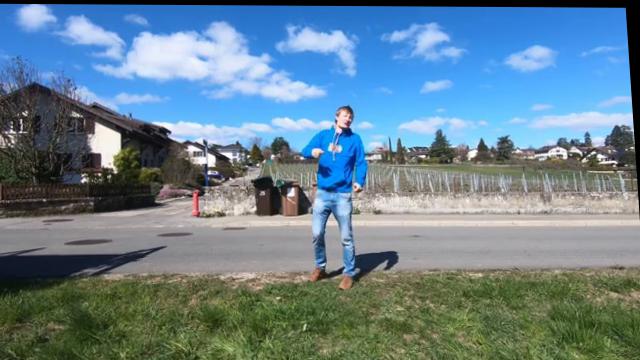} &\hspace{-4mm}
  \includegraphics[width=0.16\textwidth]{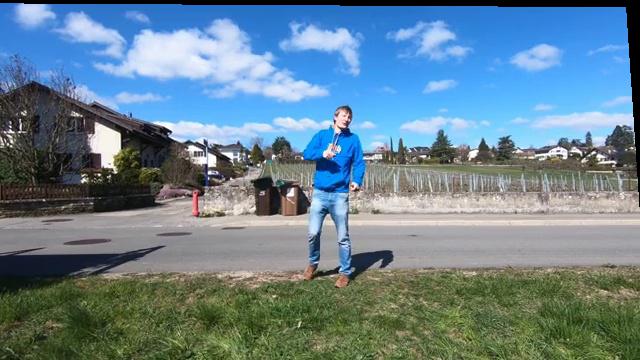} &\hspace{-4mm}
  \includegraphics[width=0.16\textwidth]{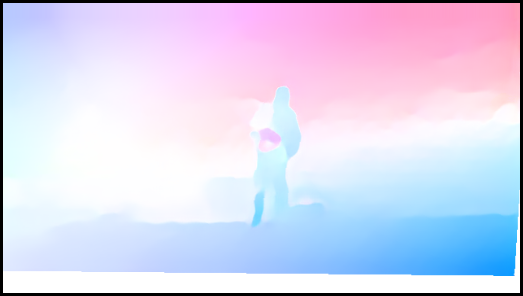} \\
  {\small (a) } & {\small (b) } & {\small (c) }
  \end{tabular}}

  \caption{\textbf{Optical flow image generation on \ski{} and \handheld{}.} We use a homography based on SIFT keypoints to compute rectified images that are provided as input to FlowNet 2.0. (a) Source image warped to the target scene;  (b) Target image; (c) Optical flow image highlighting the moving foreground region between the source image and the target image after the background motion is eliminated. In \ski{},  the optical flow images provide strong cues about the foreground object as the scene was captured by rotating cameras, making homography estimation effective. In  \handheld{}, because the camera undergoes translations, the homography and optical flow estimates are less accurate, but can nonetheless improve our segmentation performance.}
  \label{fig:optical_flow_gen}
\end{figure}

\parag{Using optical flow.}
To better handle the moving foreground subject, we use optical flow while extracting the moving region, when available. To this end, we use optical flow images obtained by running FlowNet 2.0~\cite{Ilg17} on stabilized pairs of consecutive frames obtained by computing a homography using SIFT keypoints to warp the source image to the target one. We use the resulting optical flow image $\mI_{f}$ as an intermediate supervision to our model. 
To this end, we train another inpainting network $\cI_f(\mI_{f},\vb_c)$, whose goal is to reconstruct the flow images. We then introduce an additional flow background objective $\objBG(\mI_f)$ that favors the $\vb_c$s with higher inpainting loss. This objective regulates the bounding box detection by assigning higher confidence to foreground regions through backpropagation. As shown in Fig.~\ref{fig:optical_flow_gen}, this allows us to discard the motion in the background due to a moving camera. Note that, because we use flow images only as intermediate supervision, at inference time our model still works with single images. 

\parag{Implementation details.}
We use ImageNet-trained weights for initialization of our encoder component but also report the results of our full model with no pre-training and unsupervised pre-training in our ablation study in Section~\ref{sec:ski_dataset_analysis}.
We rely on a perceptual loss on top of the per pixel $\loss_2$ loss for $\loss$; exploit the Focal Spatial Transformers (FST) of~\cite{Rhodin19a} to speed up convergence; and scale the erased region in $\cI$ to be 1.1 times the size of that predicted by $\cD$ to increase the chances of covering the object.
Moreover, we discard location offsets outside the image and limit the offset to 1.5 times the 
bounding box width, as larger ones are already fully covered by the neighboring 
bounding boxes. We performed a grid search on the relative weights of the terms, the offset limits, and $\lambda$. The pixel reconstruction and perceptual losses are weighted 1:2, and the priors on the probability values and the segmentation mask have a weight of 0.1 and 0.25, respectively, to compensate for their different magnitudes. More details are given in the appendix.

Following common practice in the self-supervised segmentation literature~\cite{Li18g,Li18j,Yang19c}, the final segmentation masks are generated by a CRF~\cite{Kraehenbuehl11} post-processing step that uses both the unary and pairwise bilateral potential terms. We evaluate the influence of this post-processing quantitatively and qualitatively in Table~\ref{tbl:results_ski_quantitative} and in the appendix, respectively.

\begin{figure}
  \centering
  \begin{tabular}{@{}cccc@{}}
\includegraphics[width=0.2\linewidth,height=0.13\linewidth]{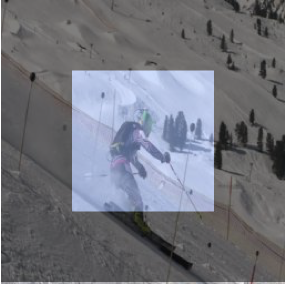}&%
\includegraphics[width=0.2\linewidth,height=0.13\linewidth]{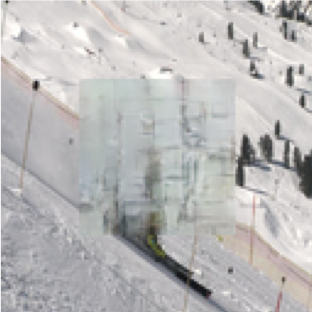}&%
\includegraphics[width=0.2\linewidth,height=0.13\linewidth]{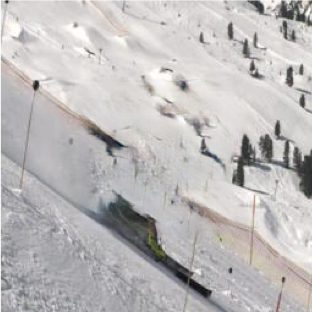}&%
\includegraphics[width=0.2\linewidth,height=0.13\linewidth]{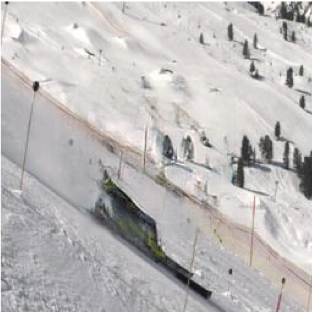}\\{\small (a)} & {\small (b) }& {\small (c) } & {\small (d) }
  \end{tabular}
  \caption{\textbf{Off-the-shelf inpainting results on \ski{}}. (a) Input image with the middle part hidden.  We show the inpainting results of (b) \cite{Pathak16}, (c)~\cite{Yu2018} trained on ImageNet and (d)~\cite{Yu2018} trained on Places2.}
  \label{fig:inpainting others}
\end{figure}

\parag{Inpainting network.}
In principle, any off-the-shelf inpainting network trained on large and generic background datasets could be used. For instance, the methods of~\cite{Pathak16,Yu2018} can produce very plausible results. However, in domain-specific images, they tend to hallucinate objects as shown in Fig.~\ref{fig:inpainting others} and are therefore ill-suited to our goal.
Instead, we train $\cI$ from scratch, by reconstructing randomly removed rectangular image regions and later use it off-the-shelf in our full pipeline as shown in Fig.~\ref{fig:pipeline}. Note that it is fine for this network not to generalize well to new scenes as it is not needed at test time.

\section{Experiments}
\label{sec:eval}

In this section, we first demonstrate the effectiveness of our approach at dealing with unusual motions acquired with PTZ cameras using the \ski{} dataset of~\cite{Rhodin18a}. We then introduce a novel \handheld{} dataset depicting people performing 14 everyday activities and a figure skating \iceskating{} dataset with different step, spin and jump combinations to demonstrate that our method can handle general moving cameras. For evaluation purposes, we provide ground-truth segmentations for both. Finally, we present the experiments with different loss functions and hyper-parameter study on the \ski{} dataset and analyze the influence of different aspects of our approach on the well-known \human{} dataset. Additional
qualitative results and details on the network architectures and the parameter search are provided in the appendix. Altogether our results show that our approach outperforms the existing self-supervised segmentation techniques, including the ones that exploit temporal cues at inference time~\cite{Koh17b, Yang19c}, approaches the accuracy of supervised methods on objects they have been trained for but seen in different conditions, and outperforms them on previously-unseen objects.

\subsection{Unusual Activity Filmed Using PTZ-Cameras}

Let us first consider the \ski{} dataset of~\cite{Rhodin18a} featuring six skiers on a slalom course. We split the videos of six skiers as four/one/one to form training, validation, and test sets, with, respectively, $7800$, $1818$ and $1908$ frames. The intrinsic and extrinsic parameters of the pan-tilt-zoom cameras are constantly adjusted to follow the skier. As a result, nothing is static in the images, the background changes quickly, and there are additional people standing as part of the background. We use the full image as input, evaluate detection accuracy using the available 2D pose annotations and segmentation accuracy by manually segmenting $16$ frames from each of the six cameras, which add up to $192$ frames in two test sequences. To determine the hyperparameter values, we use $3$ manually segmented frames from each of the six cameras, for a total $36$ frames in two validation sequences.

\newcommand*\rot{\rotatebox{90}}
\renewcommand{\arraystretch}{1}
\renewcommand{\tabcolsep}{2mm}
\begin{table*}[t]
\small
\begin{center}
\resizebox{2\columnwidth}{!}{
\begin{tabular}{ccc}
\begin{tabular}{@{}lcc@{}}
&\multicolumn{2}{c}{\ski}\\
\toprule
Method              &J Measure &F Measure\\
\midrule
VideoPCA \cite{Stretcu15}       			  & 0.54   &  0.61      \\
ARP \cite{Koh17b}								& 0.72	 &0.82		\\
ReDO \cite{Chen19a}													&0.43    &0.49  \\
Unsup-DilateU-Net \cite{Croitoru19}		& 0.63	 &0.73	\\
Unsup-Mov-Obj w/o CRF \cite{Yang19c} & 0.61 & 0.71 \\
Unsup-Mov-Obj \cite{Yang19c}			& 0.66 	 & 0.76 \\
Ours w/o optical flow													 &  0.62   &  0.69    \\
Ours w/ optical flow										 &0.70		&0.77		\\
Ours w/ optical flow + CRF								& \bf{0.73} 		&\bf{0.83}		\\
\bottomrule
\end{tabular}&

\begin{tabular}{@{}lcccc@{}}
\multicolumn{2}{c}{\handheld}\\
\toprule
& J measure & F measure \\
\midrule
 & 0.47        & 0.49         \\
 &0.60			&0.68			\\
 	& 0.33				& 0.38 		\\
	& 0.67	   & 0.75		\\
	&0.60		&0.68 \\
	&0.75		&0.83 \\
& 0.75    &  0.87 \\
& 0.70			&0.79			\\
&\bf{0.76}			    &\bfseries{0.85}				\\
\bottomrule
\end{tabular} & 
\begin{tabular}{@{}lcccc@{}}
	\multicolumn{2}{c}{\iceskating}\\
	\toprule
	& J measure & F measure \\
	\midrule
	& 0.55        & 0.69        \\
	& 0.56		&0.69			\\
	& 0.68				& 0.77 		\\
	& 0.53	   & 0.53		\\
	&0.53		&0.73 \\
	&0.68		&0.85 \\
	&0.66   & 0.72 \\
	&0.69			&0.80			\\
	&\bf{0.71}			    &\bf{0.86}				\\
	\bottomrule
\end{tabular} \\
\end{tabular}
}
\end{center}
\caption{\textbf{Segmentation results on the \ski{}, \handheld{} and \iceskating{} datasets}. Our method with optical flow consistently outperforms the other self-supervised methods, and ours without flow exceeds or is on par with the other baselines on all three datasets. The best results in each column are shown in bold.}
\label{tbl:results_ski_quantitative}
\end{table*}

In Table~\ref{tbl:results_ski_quantitative}(left), we compare our approach to several state-of-the-art self-supervised segmentation baselines in terms of 
J- and F-measure as defined in~\cite{Perazzi16}. The former is defined as the intersection-over-union between the ground truth segmentation mask and the prediction, while the latter is the harmonic average between the precision and the recall at the mask boundaries.
To be fair, we compensate for different segmentation masks quantification levels by a grid search (at 0.05 intervals) to select the best J-measure threshold for each method.
Our approach with optical flow outperforms all the baselines in terms of both J- and F-measure. When not using optical flow for training purposes, our approach remains on par with other self-supervised methods despite their use of explicit temporal dependencies. In particular, the comparison to~\cite{Yang19c} without CRF post-processing shows that our method can achieve the same performance against an optical flow based method without needing a flow-based intermediate supervision. Note that all the baselines are trained on our datasets from scratch using same amount of data, except for~\cite{Croitoru19} that additionally uses a segmentation mask discriminator trained on the combination of the ImageNet VID and YouTube Objects datasets. In other words, while this method is trained in a self-supervised fashion, it relies on a significantly larger amount of data than ours.

\begin{figure*}[t]
  \centering{
\renewcommand{\arraystretch}{0.4}
  \begin{tabular}{@{}cccccc@{}}

  \includegraphics[width=0.16\linewidth]{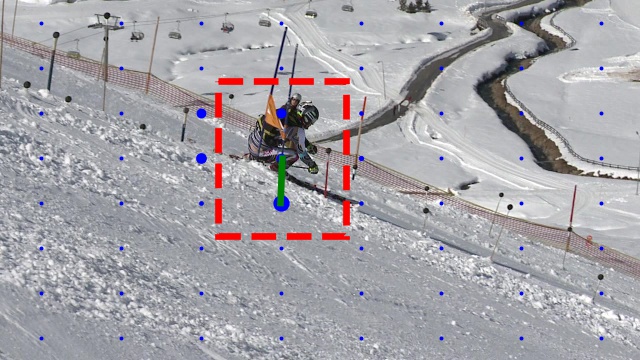} &\hspace{-4.0mm}
  \includegraphics[width=0.16\linewidth]{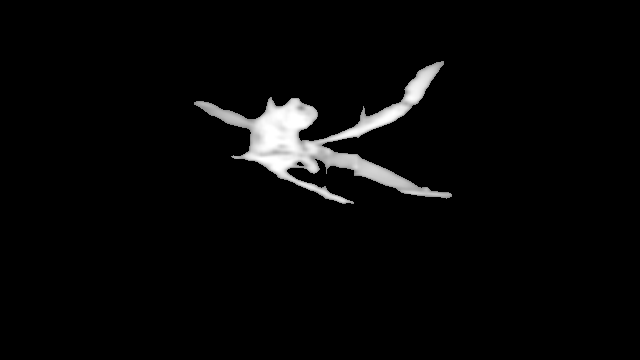}& \hspace{-4mm}
  \includegraphics[width=0.16\linewidth]{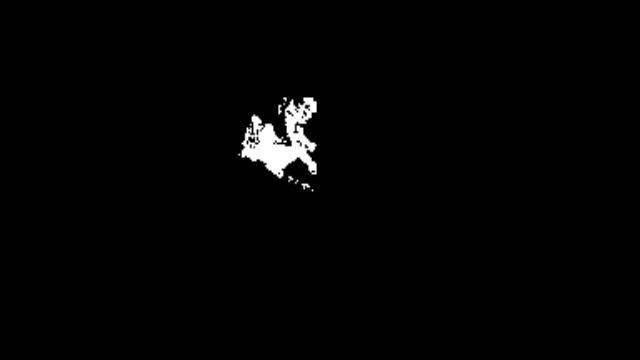}& \hspace{-4mm}
     \includegraphics[width=0.16\linewidth]{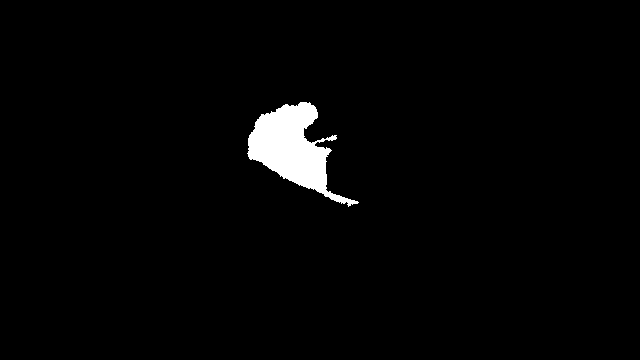}& \hspace{-4mm}
  \includegraphics[width=0.16\linewidth]{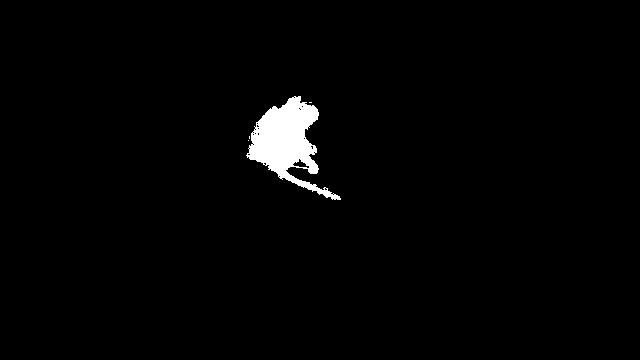}& \hspace{-4mm}
  \includegraphics[width=0.16\linewidth]{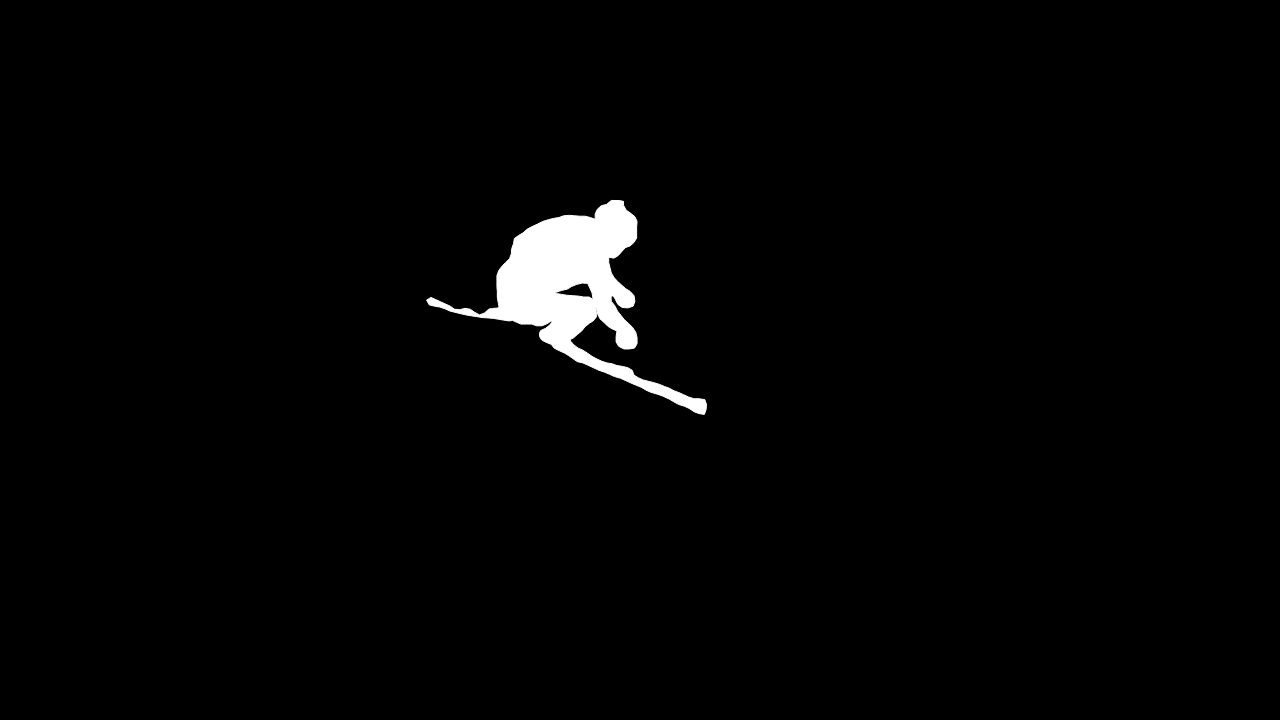}  \\

\includegraphics[width=0.16\linewidth]{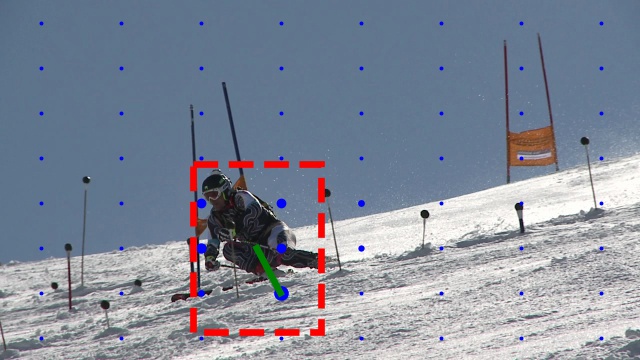} &\hspace{-4.0mm}
\includegraphics[width=0.16\linewidth]{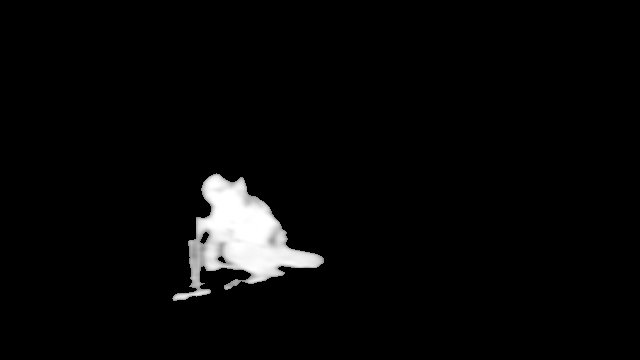} &\hspace{-4mm}
\includegraphics[width=0.16\linewidth]{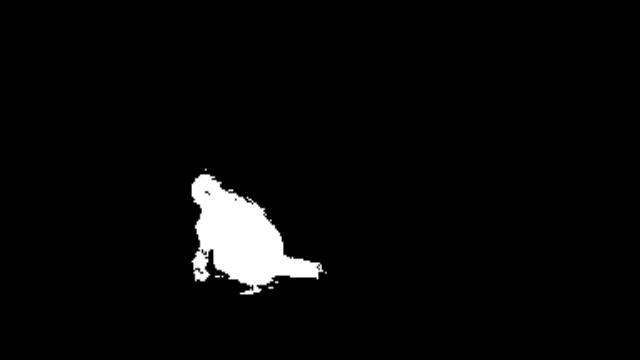} &\hspace{-4mm}
\includegraphics[width=0.16\linewidth]{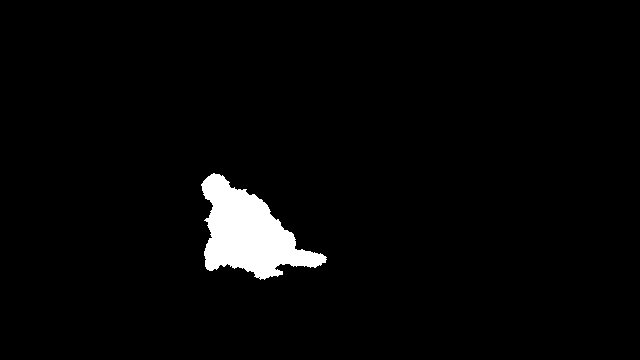} &\hspace{-4mm}
\includegraphics[width=0.16\linewidth]{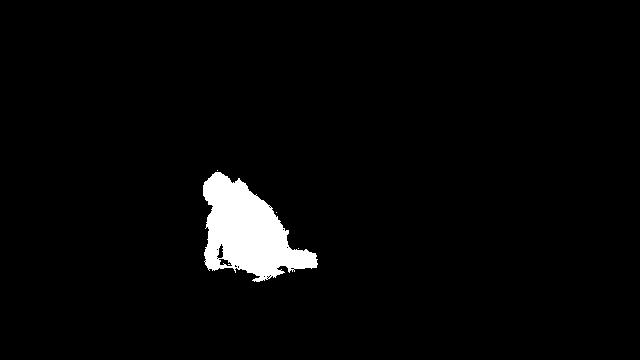} &\hspace{-4mm}
\includegraphics[width=0.16\linewidth]{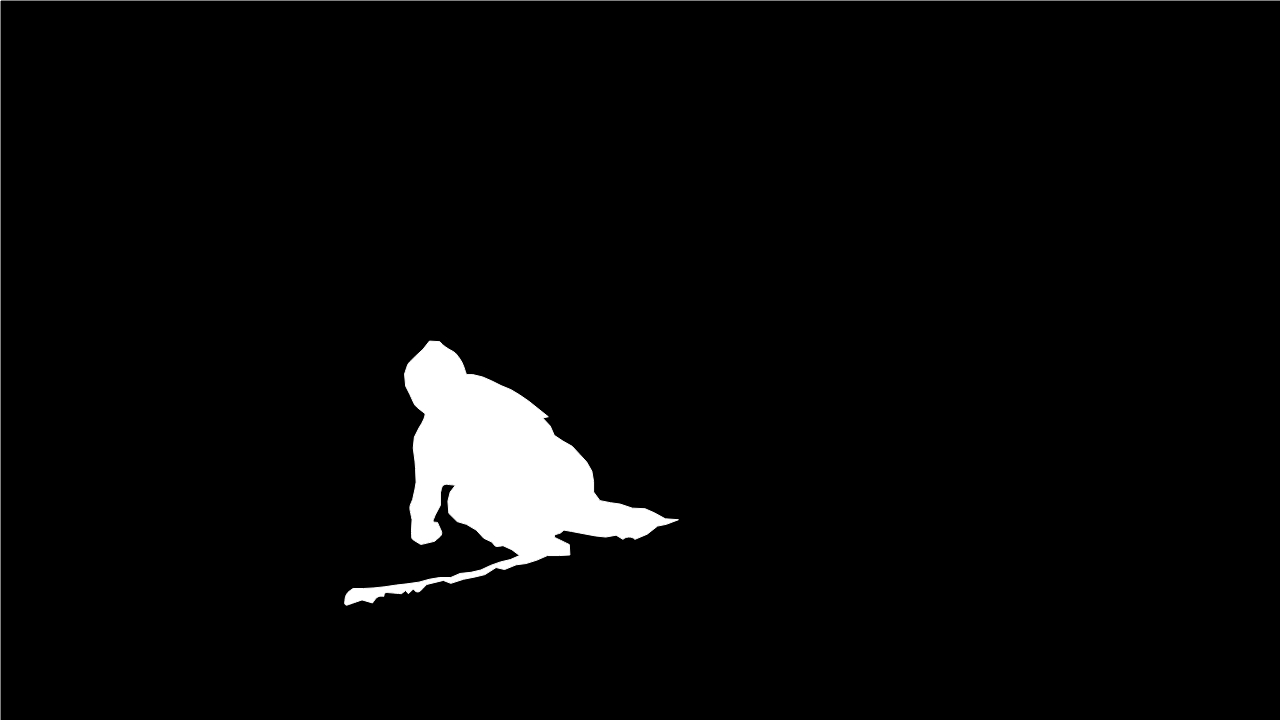} \\

\includegraphics[width=0.16\linewidth]{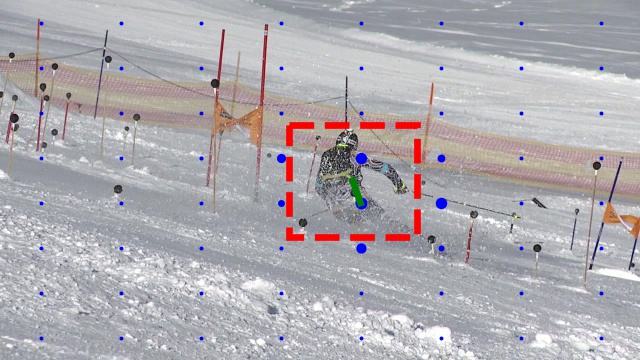} &\hspace{-4.0mm}
\includegraphics[width=0.16\linewidth]{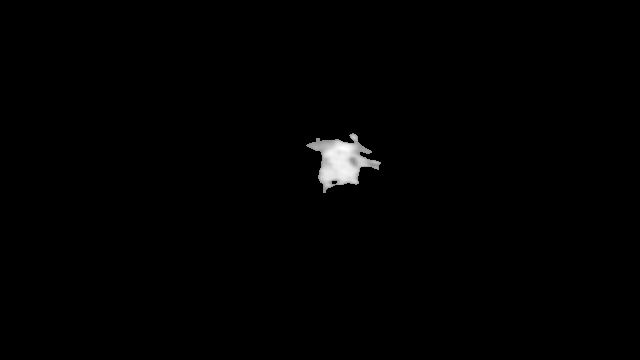} &\hspace{-4mm}
\includegraphics[width=0.16\linewidth]{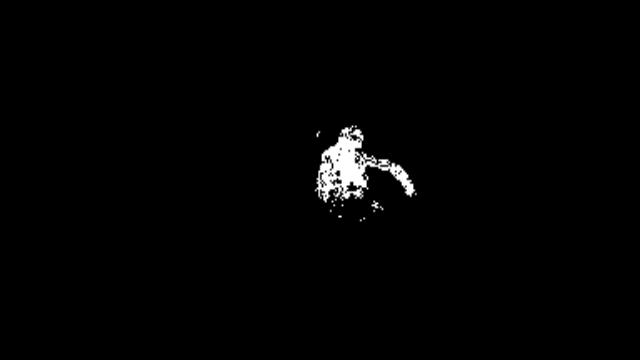} &\hspace{-4mm}
\includegraphics[width=0.16\linewidth]{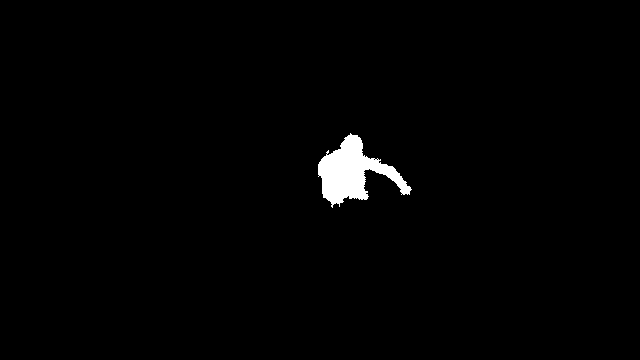} &\hspace{-4mm}
\includegraphics[width=0.16\linewidth]{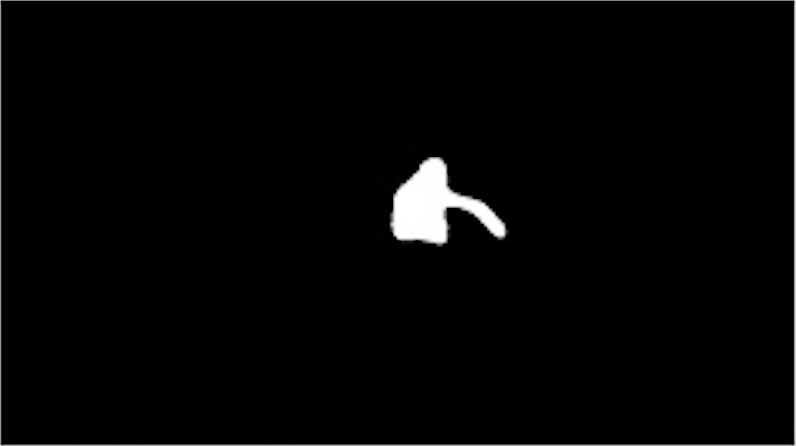} &\hspace{-4mm}
\includegraphics[width=0.16\linewidth]{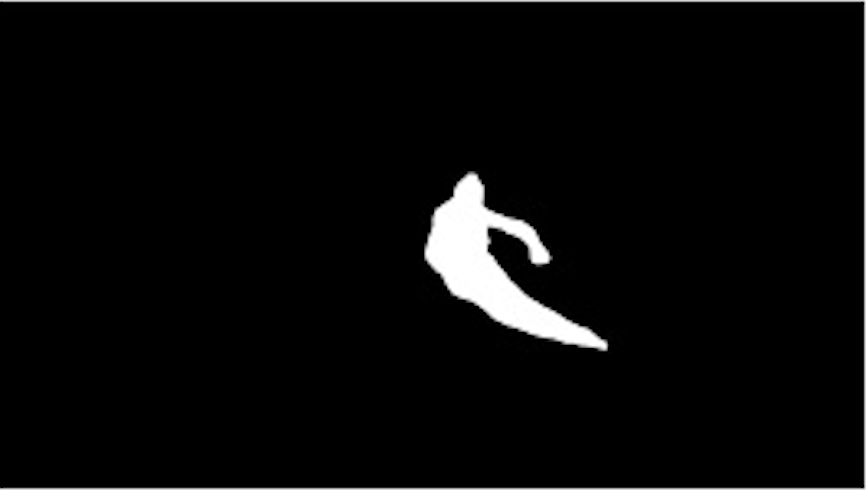} \\

\includegraphics[width=0.16\linewidth]{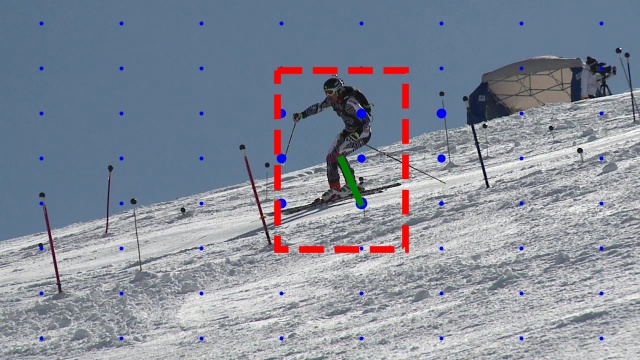} &\hspace{-4.0mm}
\includegraphics[width=0.16\linewidth]{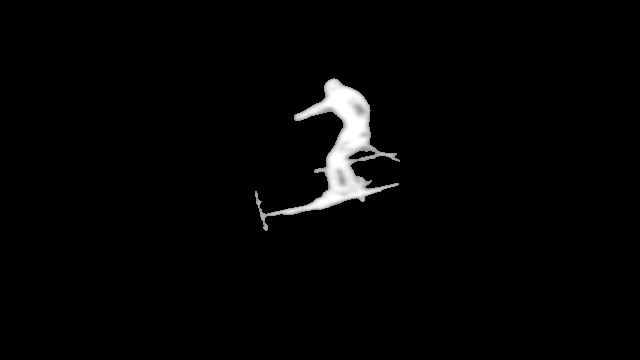} &\hspace{-4mm}
\includegraphics[width=0.16\linewidth]{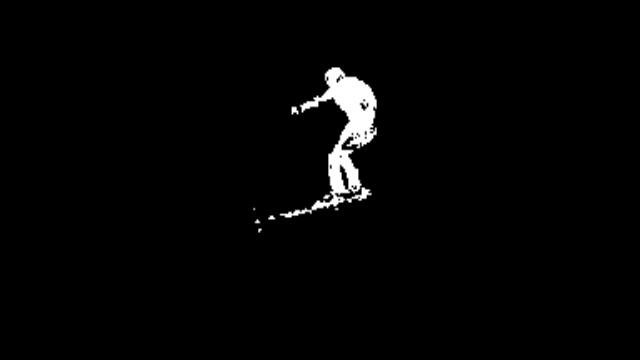} &\hspace{-4mm}
\includegraphics[width=0.16\linewidth]{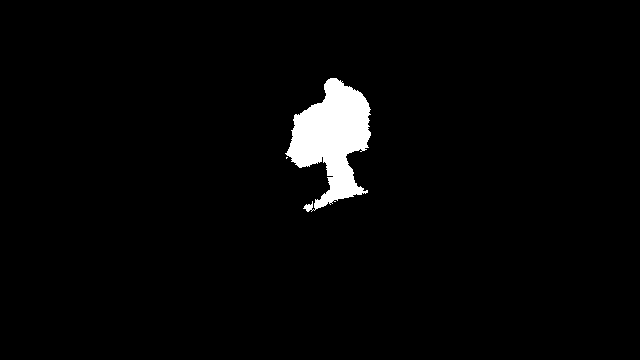} &\hspace{-4mm}
\includegraphics[width=0.16\linewidth]{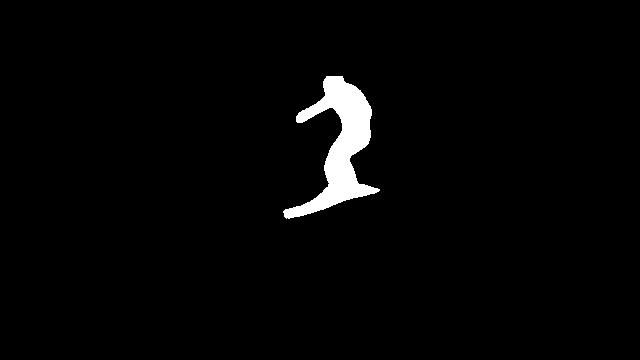} &\hspace{-4mm}
\includegraphics[width=0.16\linewidth]{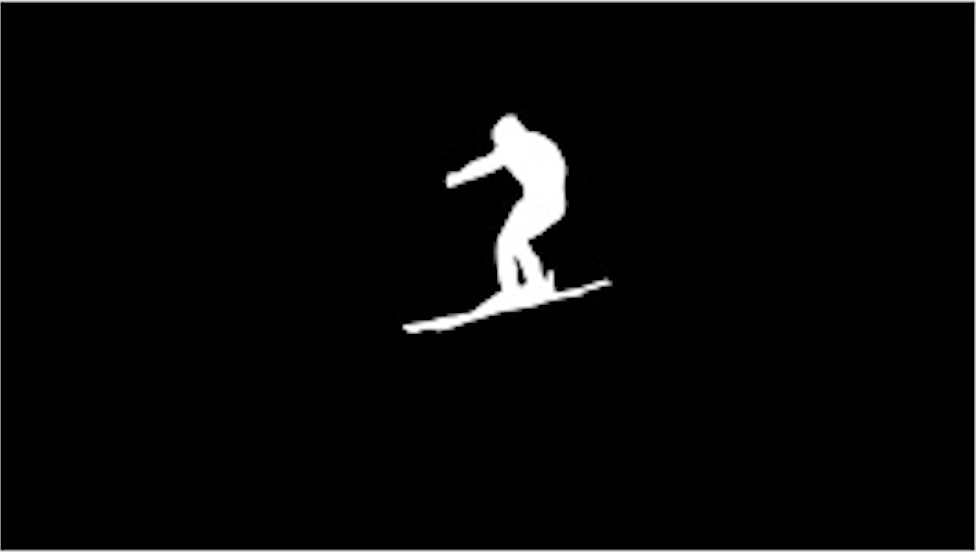} \\

\includegraphics[width=0.16\linewidth]{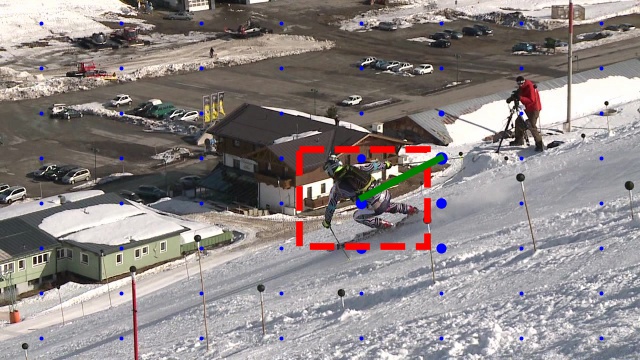} &\hspace{-4.0mm}

\includegraphics[width=0.16\linewidth]{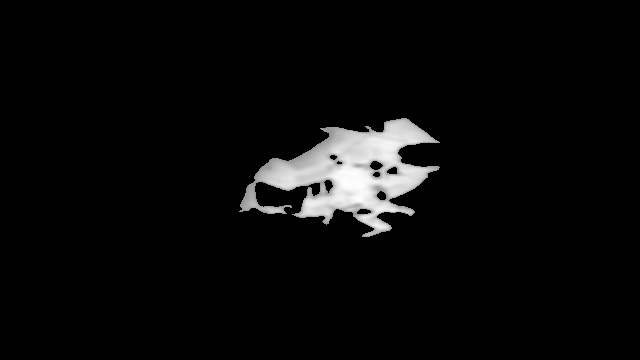} &\hspace{-4mm}
\includegraphics[width=0.16\linewidth]{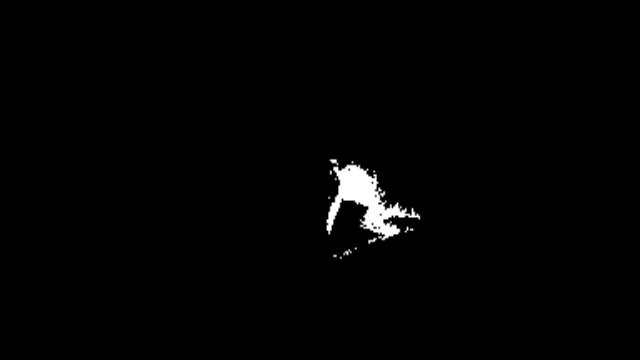} &\hspace{-4mm}
\includegraphics[width=0.16\linewidth]{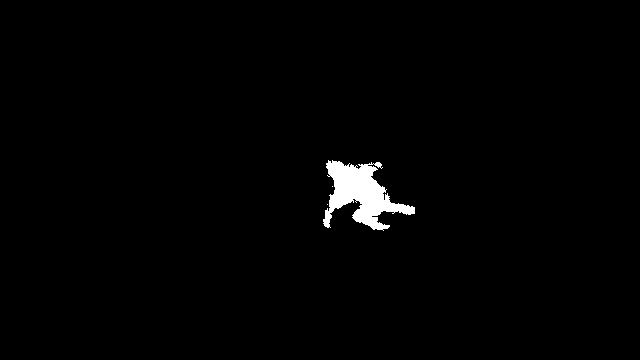} &\hspace{-4mm}
\includegraphics[width=0.16\linewidth]{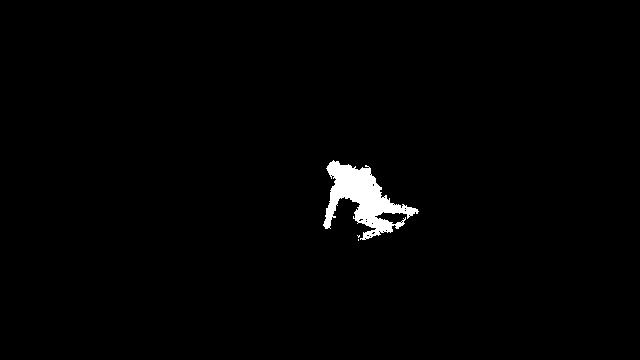} &\hspace{-4mm}
\includegraphics[width=0.16\linewidth]{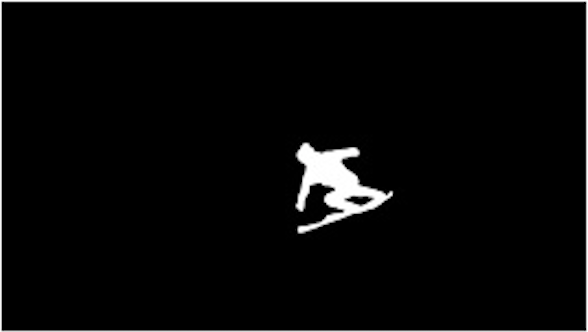} \\

\includegraphics[width=0.16\linewidth]{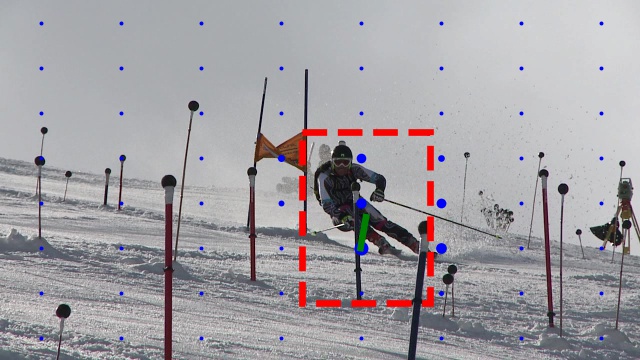} &\hspace{-4.0mm}
\includegraphics[width=0.16\linewidth]{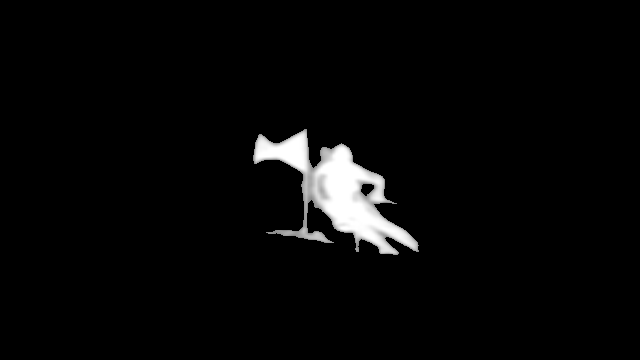} &\hspace{-4mm}
\includegraphics[width=0.16\linewidth]{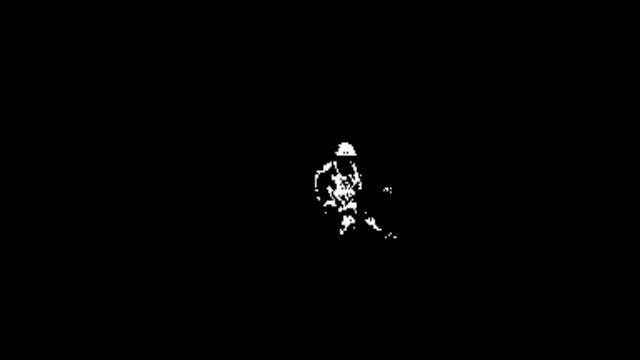} &\hspace{-4mm}
\includegraphics[width=0.16\linewidth]{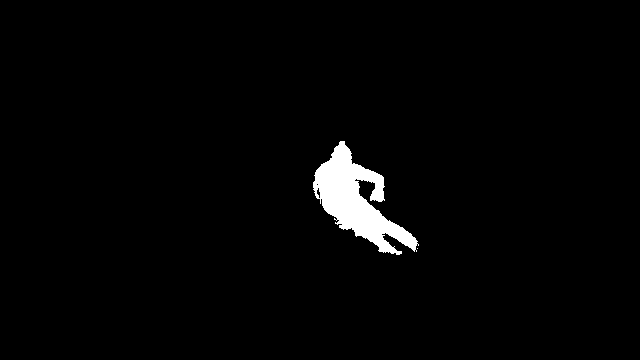} &\hspace{-4mm}
\includegraphics[width=0.16\linewidth]{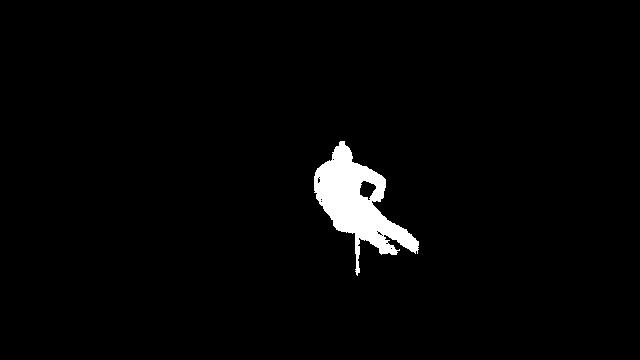} &\hspace{-4mm}
\includegraphics[width=0.16\linewidth]{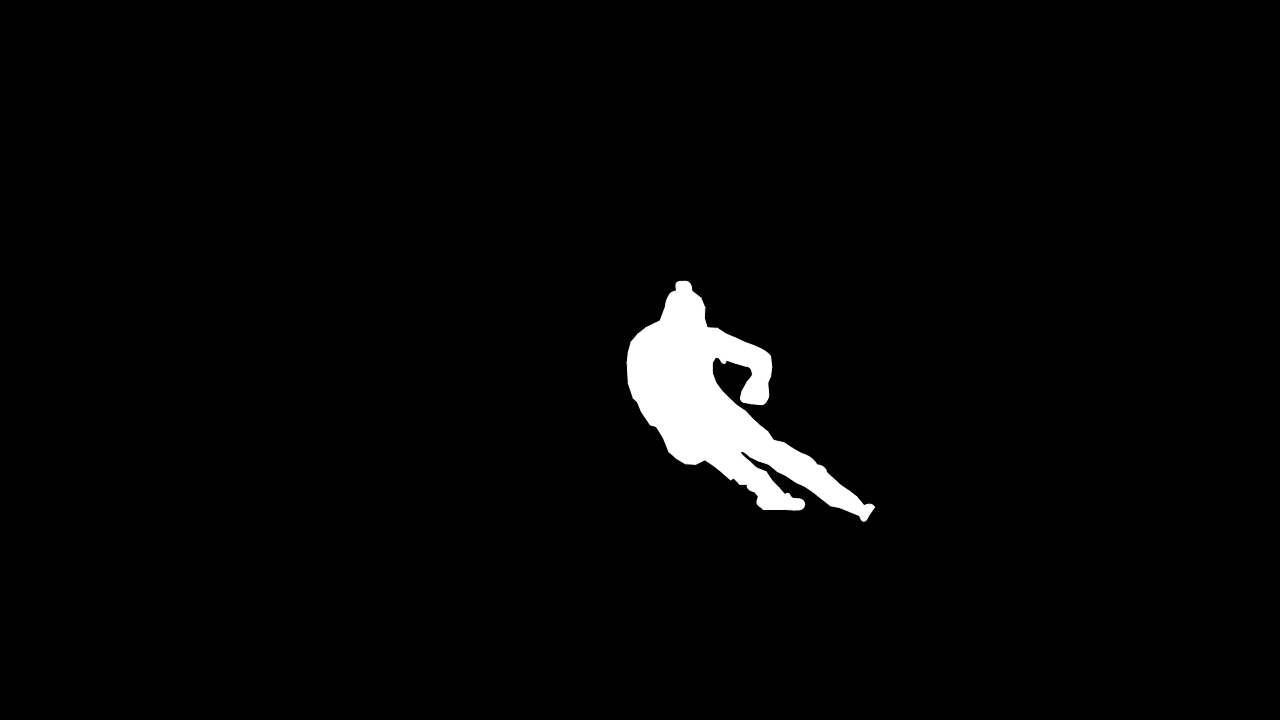} \\  \\

  {\scriptsize  \hspace{-1mm}(a) Input/Ours detection}& \hspace{-1mm}{\scriptsize (b) Unsup-DilateU-Net~\cite{Croitoru19}}&\hspace{-4mm}{\scriptsize (c) Unsup-Mov-Obj~\cite{Yang19c}} &\hspace{-4mm}{\scriptsize (d) ARP~\cite{Koh17b}}&\hspace{-4mm}{\scriptsize (e) Ours w/ optical flow}  &\hspace{-4mm}{\scriptsize (f) Ground truth }  \\ 
  \end{tabular}}

  \caption{\textbf{Qualitative results on the \ski{}.} Example results on the test images. (a) The detection results show the predicted bounding box with red dashed lines, the relative confidence of the grid cells with blue dots and the bounding box center offset with green lines (better viewed on screen). (b) Segmentation mask prediction of~\cite{Croitoru19}. (c)Segmentation mask prediction of~\cite{Yang19c}. (d) Segmentation mask prediction of~\cite{Koh17b}. (e) Our segmentation mask prediction. (f) Ground truth segmentation mask. Note that in the third row even though the skier is mostly occluded by snow, our method can detect and segment the visible part of the body. Our method is more accurate than~\cite{Croitoru19} in terms of background removal and outperforms~\cite{Yang19c} in terms of correctness of the object boundary. Note that in contrast to our method,~\cite{Koh17b} uses explicit temporal cues at inference time.}
  \label{fig:ski_qual}
\end{figure*}
We provide qualitative results in Fig.~\ref{fig:ski_qual}. The probability distribution, visualized as blue dots whose magnitude reflect the predicted likelihood, shows clear peaks on the persons. The limitations include occasional false positives, such as the gates on the slope in close proximity to the skier, reducing precision.

\begin{figure}[t]
\begin{tabular}{@{}ccc@{}}

	\includegraphics[width=0.31\linewidth]{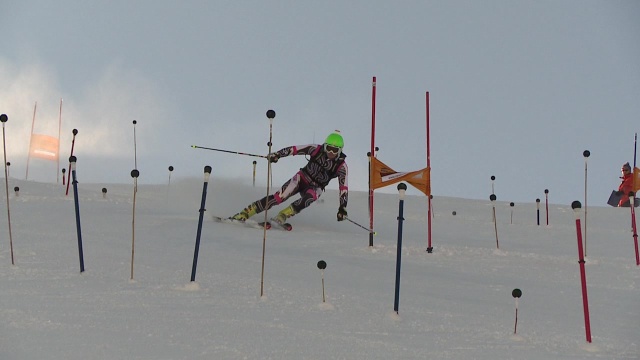} & \hspace{-4 mm}
	\includegraphics[width=0.31\linewidth]{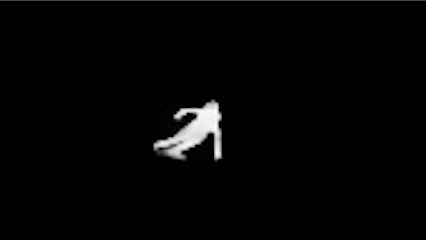}& \hspace{-4 mm}
	\includegraphics[width=0.31\linewidth]{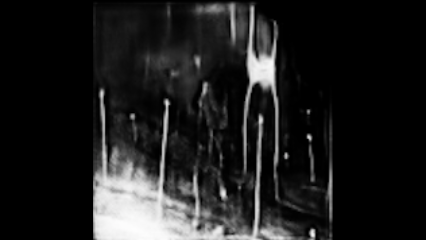}  \\
	
	\includegraphics[width=0.31\linewidth]{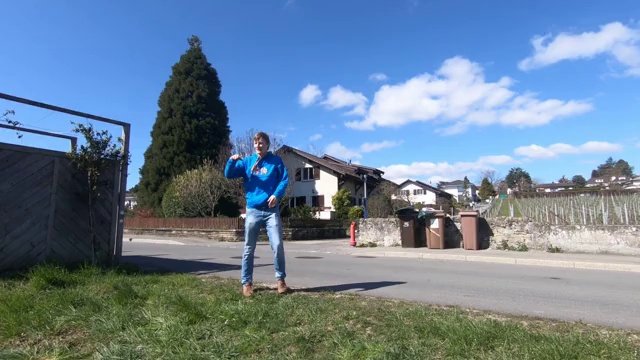} &\hspace{-4 mm}
	\includegraphics[width=0.31\linewidth]{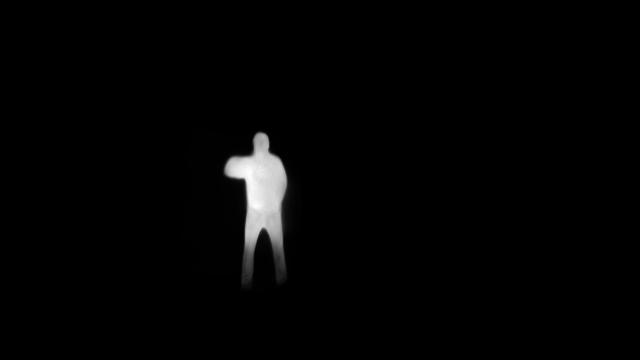}& \hspace{-4 mm}
	\includegraphics[width=0.31\linewidth]{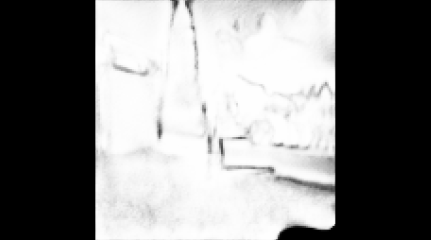}  \\

	{\small (a) Input}& {\small (b) Ours} & {\small (c) P-GAN~\cite{Bielski19}}  \\
\end{tabular}

\caption{\textbf{Soft segmentation masks generated by our method and PerturbedGAN (P-GAN)~\cite{Bielski19} on training examples.} Top row: P-GAN mask generated on the \ski{} dataset, the poles and snow patches are segmented as foreground. Bottom row: P-GAN mask generated on the \handheld{} dataset contains the foreground subject together with the ground they are standing on.}
\label{fig:gen_perturbed}
\end{figure}

In Fig.~\ref{fig:gen_perturbed}, we compare our method qualitatively to a recent self-supervised method~\cite{Bielski19}. Note that their generative model fails to segment the foreground object alone and instead segments background objects and sometimes even the ground. Therefore, we couldn't obtain any reasonable quantitative results for~\cite{Bielski19}. This method relies on the property that foreground regions can undergo random perturbations without altering the realism of the scene. However, in the \ski{} dataset, some background objects, such as poles, also satisfy this property, and the generator can choose to keep these regions. We also trained~\cite{Arandjelovic19}, another recent self-supervised method that discovers object masks by copying the selected region of the image onto another image with the goal of obtaining a realistic scene, on the \ski{} dataset and obtained implausible masks for the same reason. Since these methods performed poorly on the training samples, we do not provide their quantitative results on the test data. 

\subsection{Activities Captured Using Moving Cameras}

To demonstrate the effectiveness of our approach in the presence of general moving cameras, we introduce a new \handheld{} dataset captured by hand-held cameras. It features three training, one validation and one test sequences, comprising $120\,855$, $23\,076$ and $46\,326$ images, respectively, with a single actor performing actions mimicking those in \human{}. We manually annotated $112$ frames in the validation and $240$ frames in the test sequence to provide ground truth segmentation masks, which we believe will be useful for evaluating other self- and weakly-supervised methods.
The camera operators moved laterally, to test robustness to camera translation and hand-held rotation. We provide examples of our detection and segmentation results in Fig.~\ref{fig:handheld_qual}, and more are given in the appendix. Our method is robust to the undirected camera motion and to dynamic background motion,  such as branches swinging in the wind and clouds moving, and to salient textures in the background, such as that of the house facade.

\begin{figure*}[t]
  \centering{
\renewcommand{\arraystretch}{0.4}
  \begin{tabular}{@{}cccccc@{}}

  \includegraphics[width=0.16\linewidth]{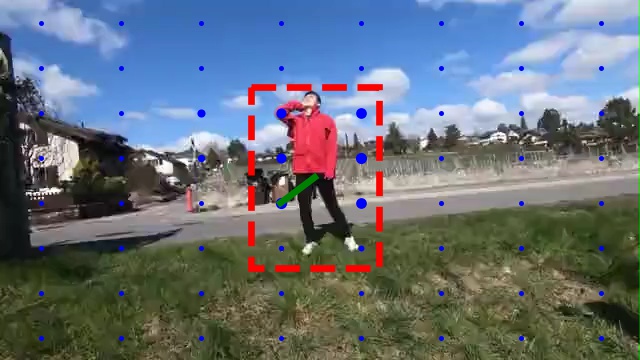} &\hspace{-4.0mm}
  \includegraphics[width=0.16\linewidth]{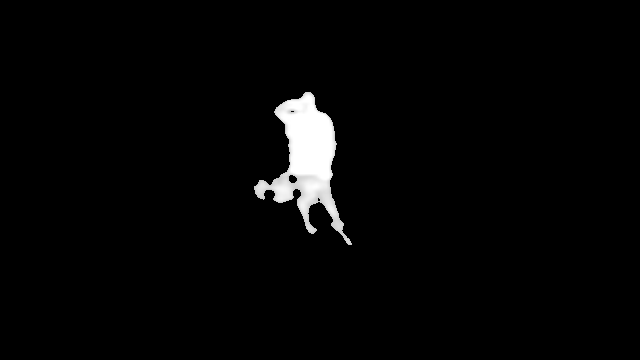}& \hspace{-4mm}
  \includegraphics[width=0.16\linewidth]{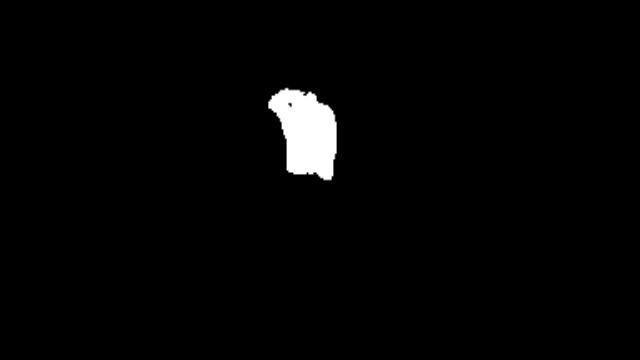}& \hspace{-4mm}
     \includegraphics[width=0.16\linewidth]{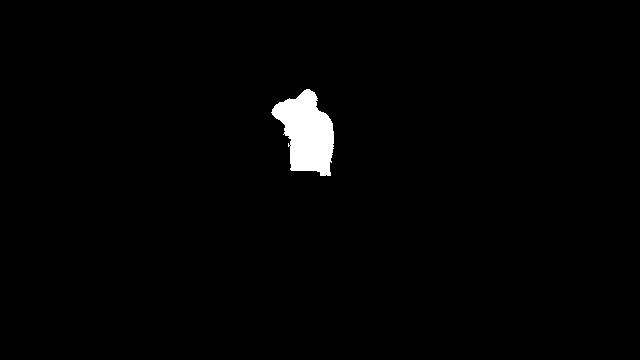}& \hspace{-4mm}
  \includegraphics[width=0.16\linewidth]{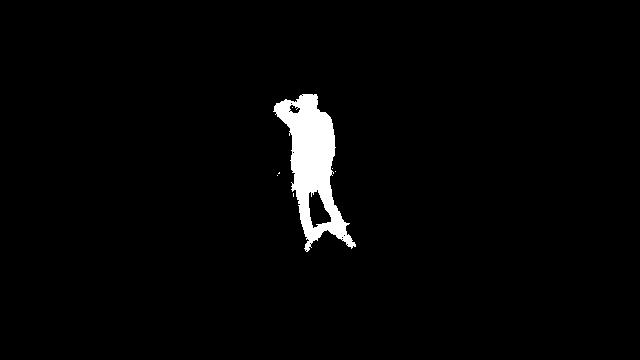}& \hspace{-4mm}
  \includegraphics[width=0.16\linewidth]{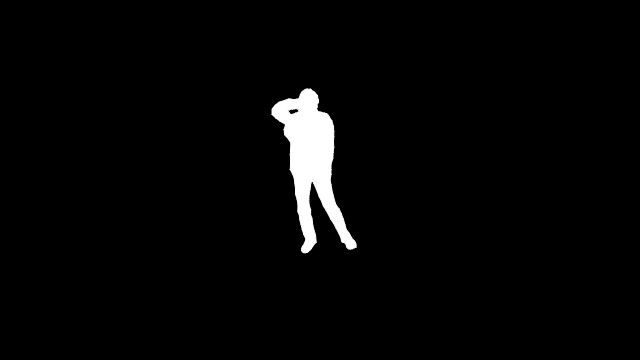}  \\

\includegraphics[width=0.16\linewidth]{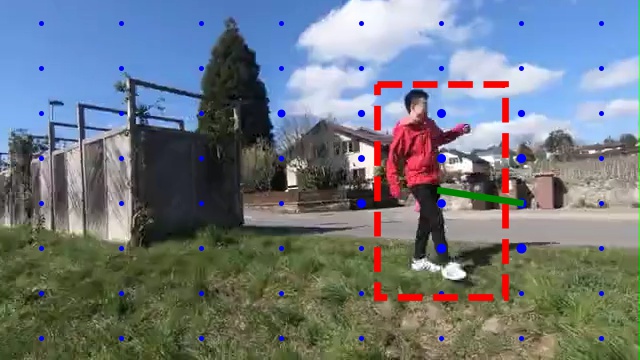} &\hspace{-4.0mm}
\includegraphics[width=0.16\linewidth]{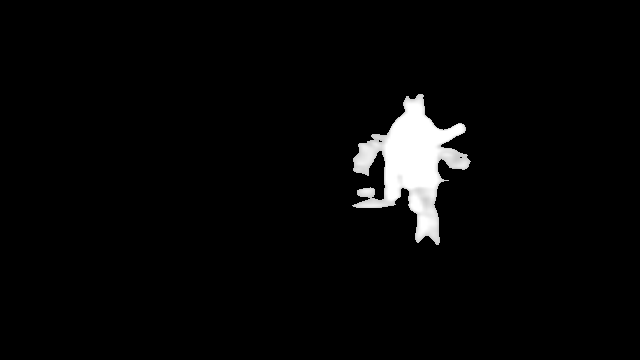} &\hspace{-4mm}
\includegraphics[width=0.16\linewidth]{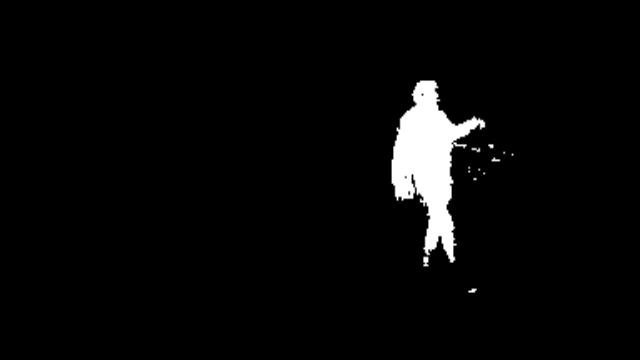} &\hspace{-4mm}
\includegraphics[width=0.16\linewidth]{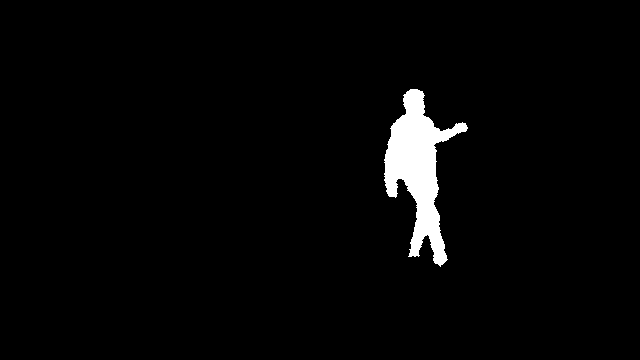} &\hspace{-4mm}
\includegraphics[width=0.16\linewidth]{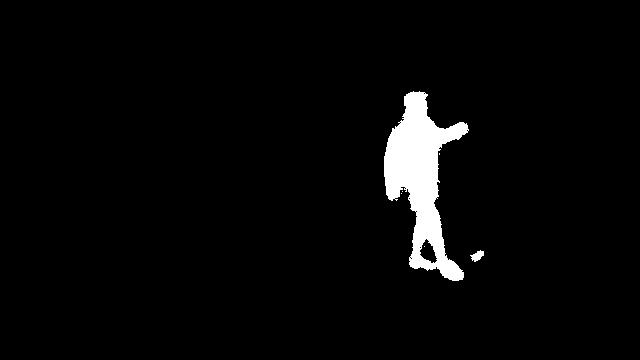} &\hspace{-4mm}
\includegraphics[width=0.16\linewidth]{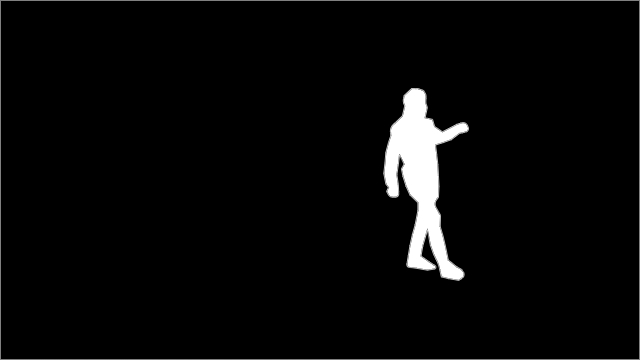} \\

\includegraphics[width=0.16\linewidth]{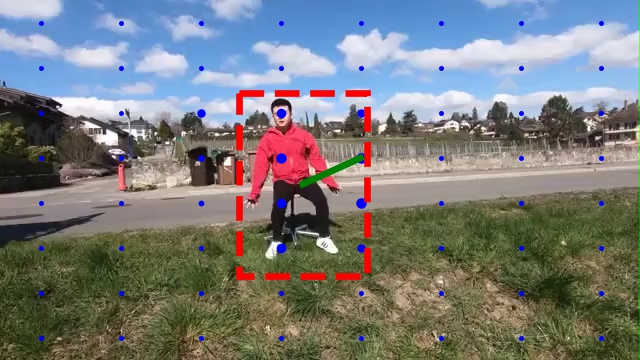} &\hspace{-4.0mm}
\includegraphics[width=0.16\linewidth]{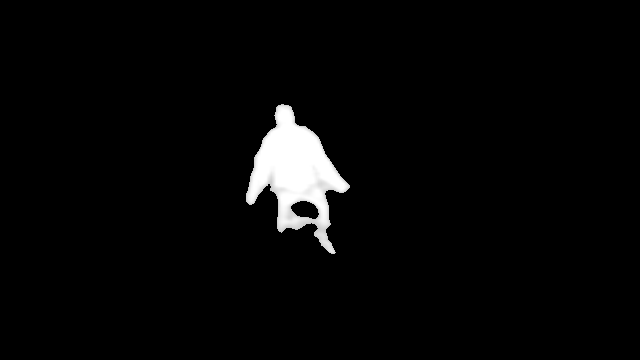} &\hspace{-4mm}
\includegraphics[width=0.16\linewidth]{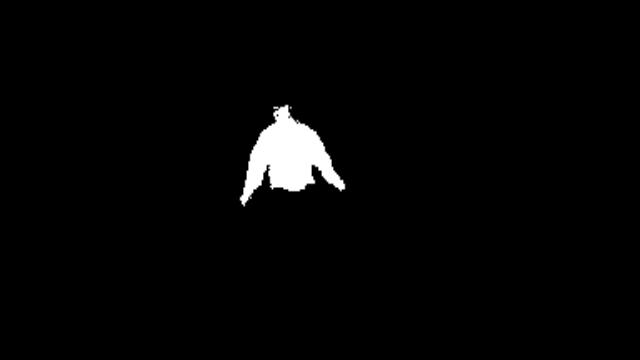} &\hspace{-4mm}
\includegraphics[width=0.16\linewidth]{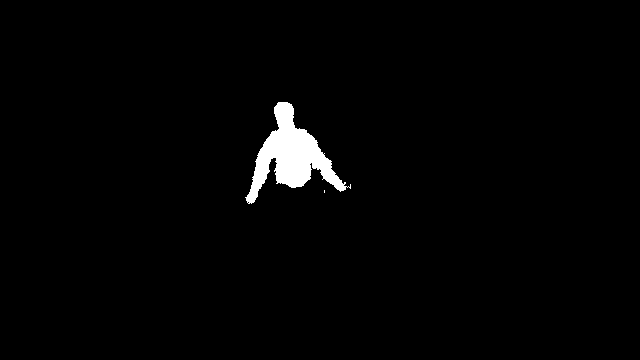} &\hspace{-4mm}
\includegraphics[width=0.16\linewidth]{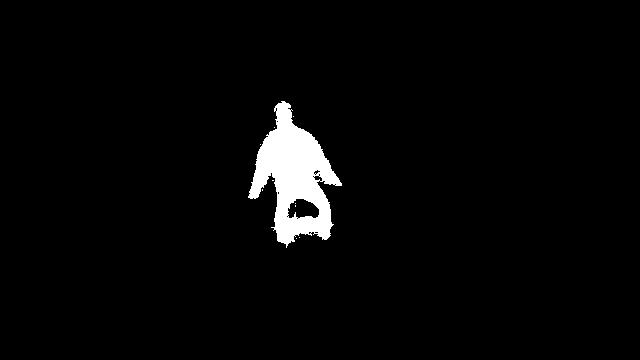} &\hspace{-4mm}
\includegraphics[width=0.16\linewidth]{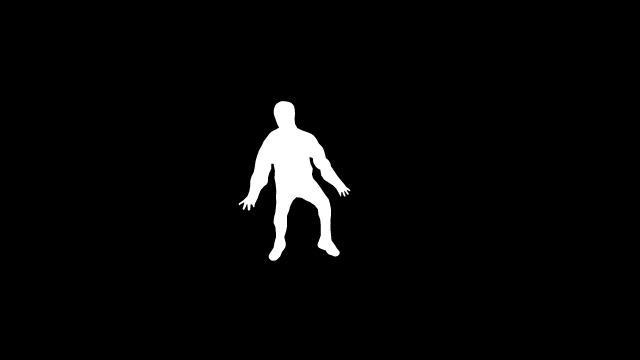} \\

\includegraphics[width=0.16\linewidth]{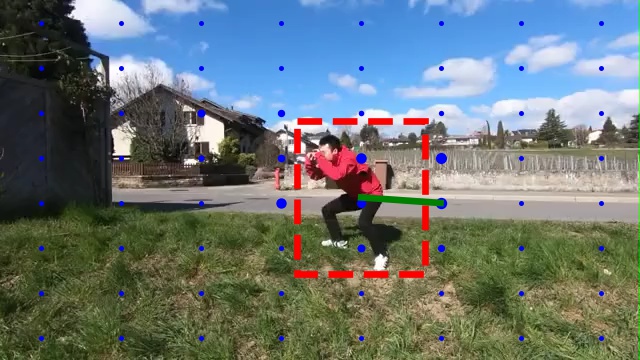} &\hspace{-4.0mm}
\includegraphics[width=0.16\linewidth]{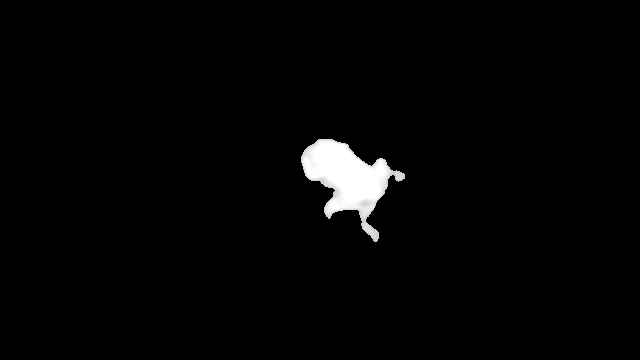} &\hspace{-4mm}
\includegraphics[width=0.16\linewidth]{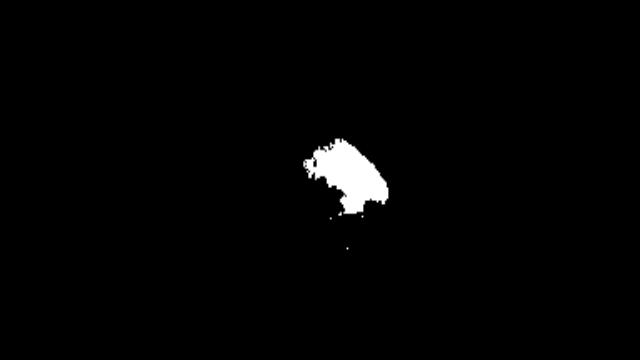} &\hspace{-4mm}
\includegraphics[width=0.16\linewidth]{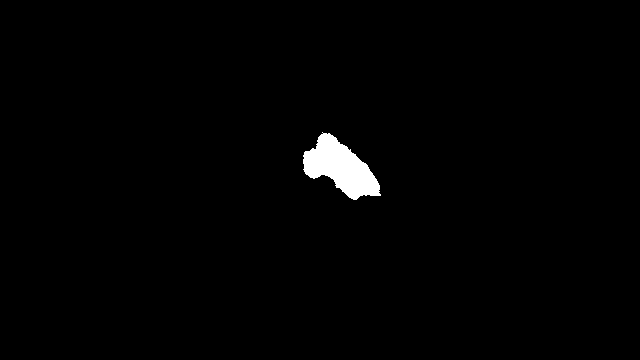} &\hspace{-4mm}
\includegraphics[width=0.16\linewidth]{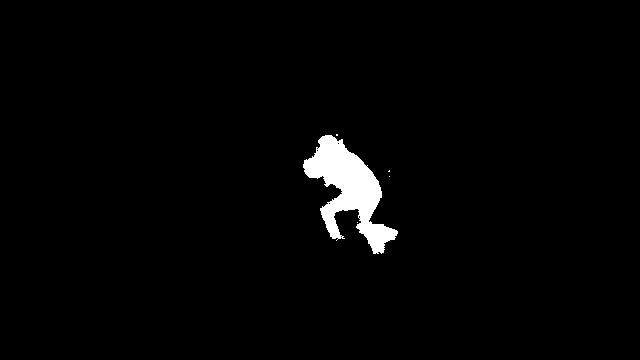} &\hspace{-4mm}
\includegraphics[width=0.16\linewidth]{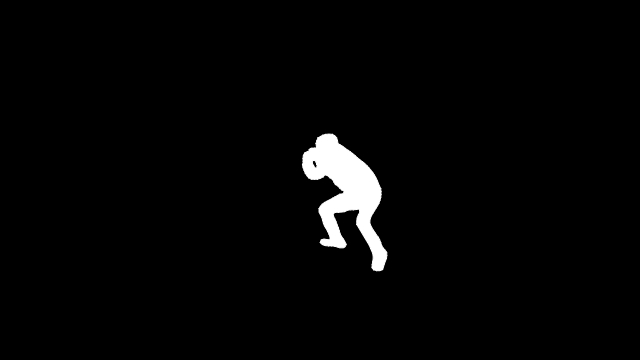} \\

\includegraphics[width=0.16\linewidth]{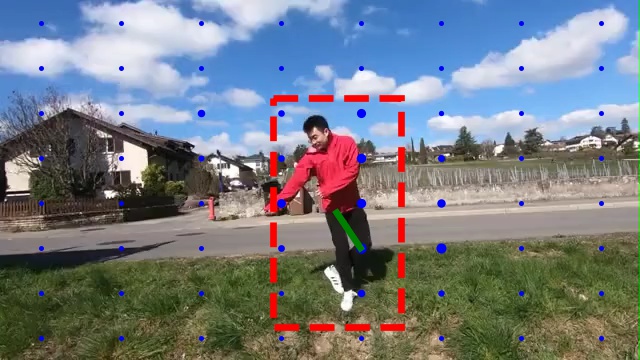} &\hspace{-4.0mm}
\includegraphics[width=0.16\linewidth]{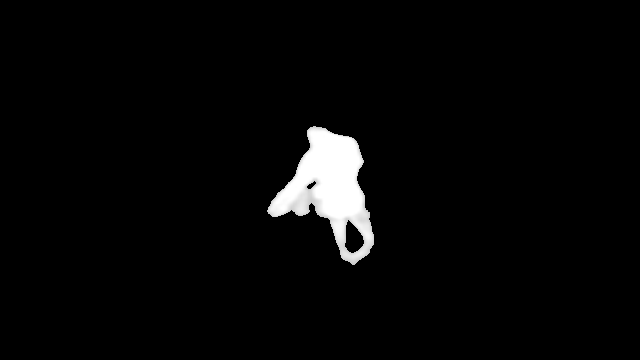} &\hspace{-4mm}
\includegraphics[width=0.16\linewidth]{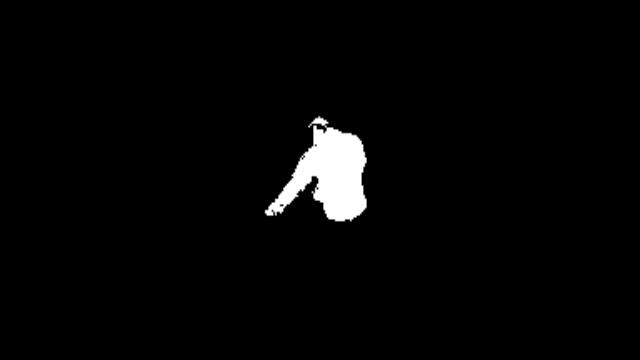} &\hspace{-4mm}
\includegraphics[width=0.16\linewidth]{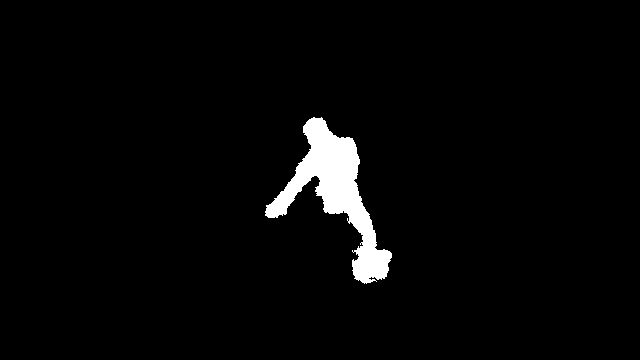} &\hspace{-4mm}
\includegraphics[width=0.16\linewidth]{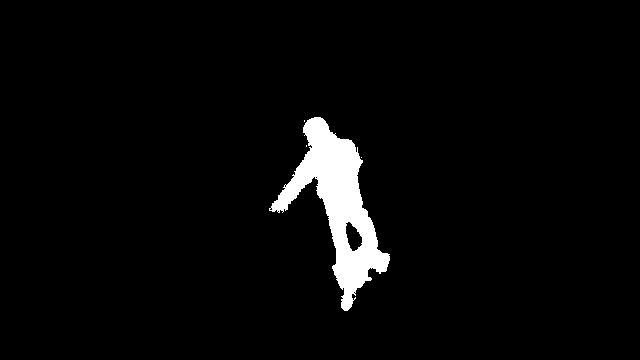} &\hspace{-4mm}
\includegraphics[width=0.16\linewidth]{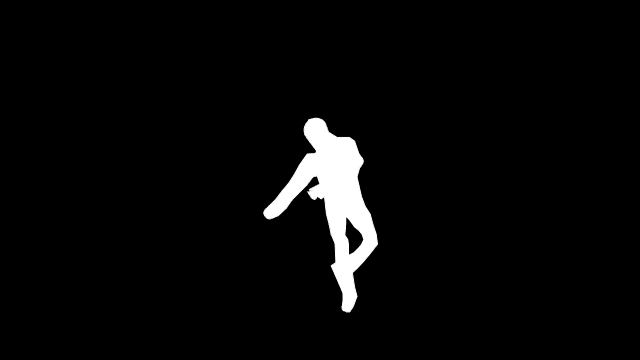} \\  \\

  {\scriptsize  \hspace{-1mm}(a) Input/Ours detection}& \hspace{-1mm}{\scriptsize (b) Unsup-DilateU-Net~\cite{Croitoru19}}&\hspace{-4mm}{\scriptsize (c) Unsup-Mov-Obj~\cite{Yang19c}} &\hspace{-4mm}{\scriptsize (d) ARP~\cite{Koh17b}}&\hspace{-4mm}{\scriptsize (e) Ours w/ optical flow}  &\hspace{-4mm}{\scriptsize (f) Ground truth }  \\ 
  \end{tabular}}

  \caption{\textbf{Qualitative results on the \handheld{}.} (a) Our detection result. The blue dots coincide with the grid cell centers and their size indicates the confidence of the bounding box proposals. The selected bounding box is illustrated with a red dashed line and the center of the grid cell yielding this proposal is connected to the center of the red box through the green line. (b) Segmentation mask prediction of~\cite{Croitoru19}. (c)Segmentation mask prediction of~\cite{Yang19c}. (d)Segmentation mask prediction of~\cite{Koh17b}. (e) Our segmentation mask prediction. (f) Ground truth segmentation mask. Our method can segment the full body of the actor more accurately than~\cite{Croitoru19,Yang19c,Koh17b} despite the other moving objects in the scene such as the clouds and occasionally appearing cars and pedestrians. In some frames, the shadow is also segmented since it moves with the primary object.}
  \label{fig:handheld_qual}
\end{figure*}

To perform a quantitative comparison, we use the $240$ manually-segmented test images taken from different motion classes with the subject in many different poses. In Table~\ref{tbl:results_ski_quantitative}(middle), we compare the results of our approach with those of the same methods as for the ski dataset. Our approach, both with and without optical flow, outperforms all the self-supervised baselines. This is even true for~\cite{Croitoru19} despite its use of a much larger dataset to train a discriminator in an unsupervised fashion and also for~\cite{Koh17b} that exploits strong temporal dependencies. 

We also evaluate our method on a new \iceskating{} dataset composed of single men's figure skating videos collected from YouTube. The videos are captured by general moving cameras and these cameras are usually adjusted fast enough to follow the movements of the skater to keep the subject in the footage. The \iceskating{} dataset contains $18$ training, $2$ validation and $3$ test sequences with $10\,613$,  $684$ and $1656$ frames and $6$, $2$ and $1$ skaters, respectively.

\begin{figure*}[t]
	\centering{
		\renewcommand{\arraystretch}{0.4}
		\begin{tabular}{@{}cccccc@{}}

			\includegraphics[width=0.16\linewidth]{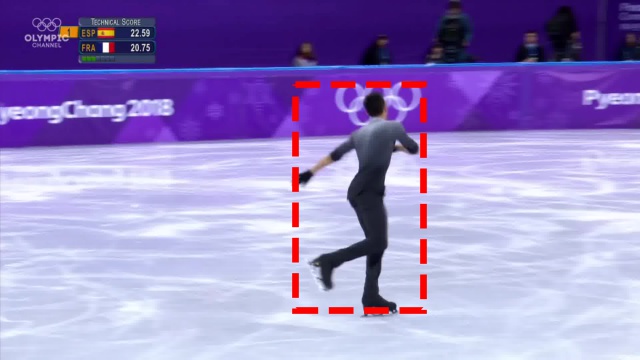} &\hspace{-4.0mm}
			\includegraphics[width=0.16\linewidth]{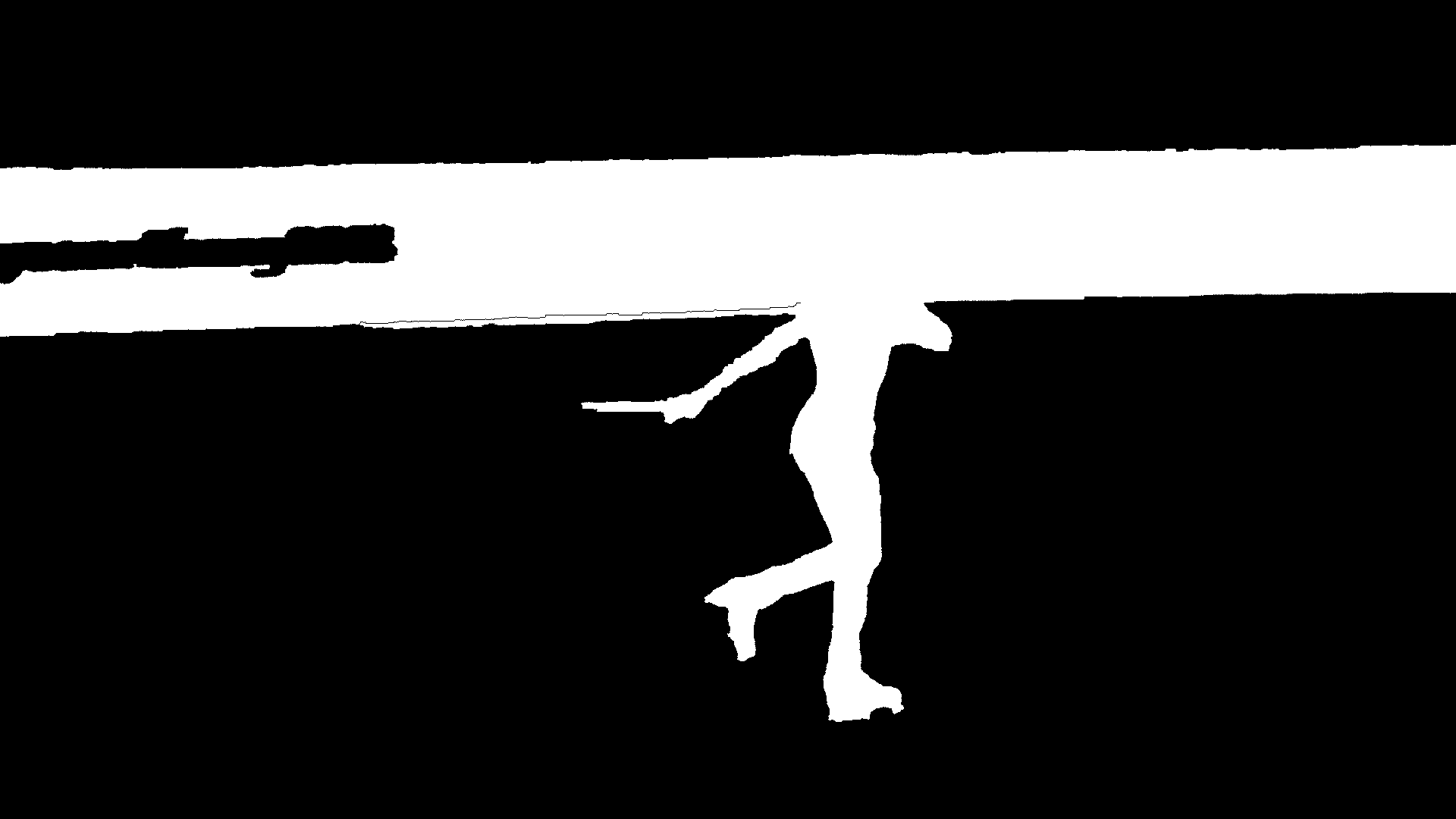} &\hspace{-4mm}
			\includegraphics[width=0.16\linewidth]{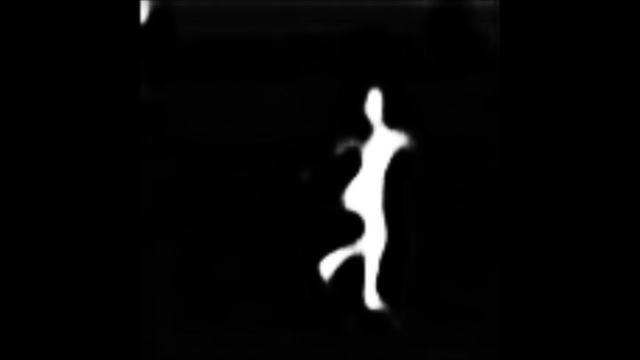} &\hspace{-4mm}
			\includegraphics[width=0.16\linewidth]{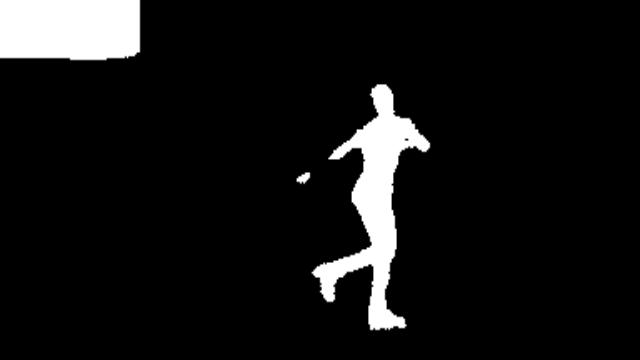} &\hspace{-4mm}
			\includegraphics[width=0.16\linewidth]{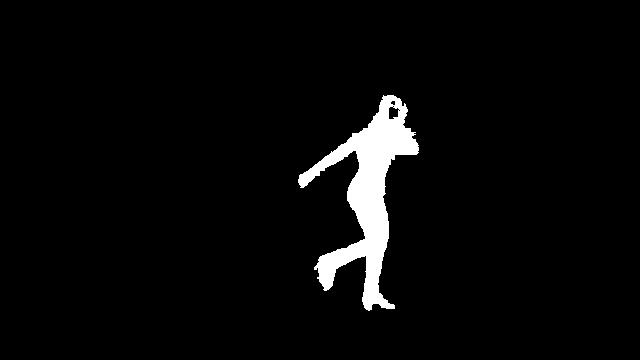} &\hspace{-4mm}
			\includegraphics[width=0.16\linewidth]{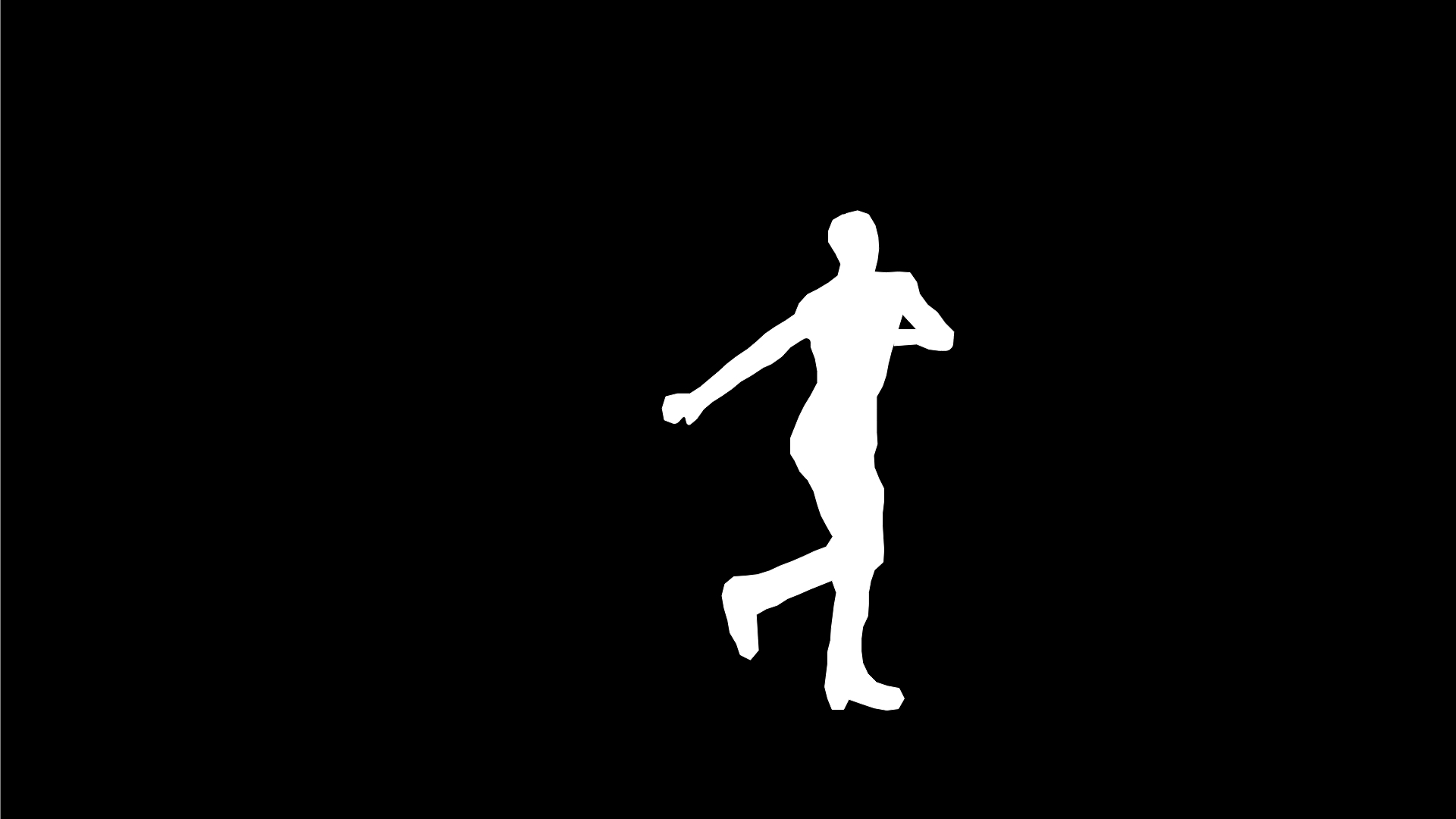} \\  
			
			\includegraphics[width=0.16\linewidth]{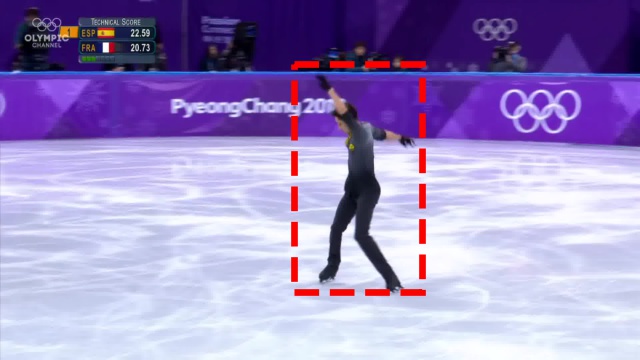} &\hspace{-4.0mm}
			\includegraphics[width=0.16\linewidth]{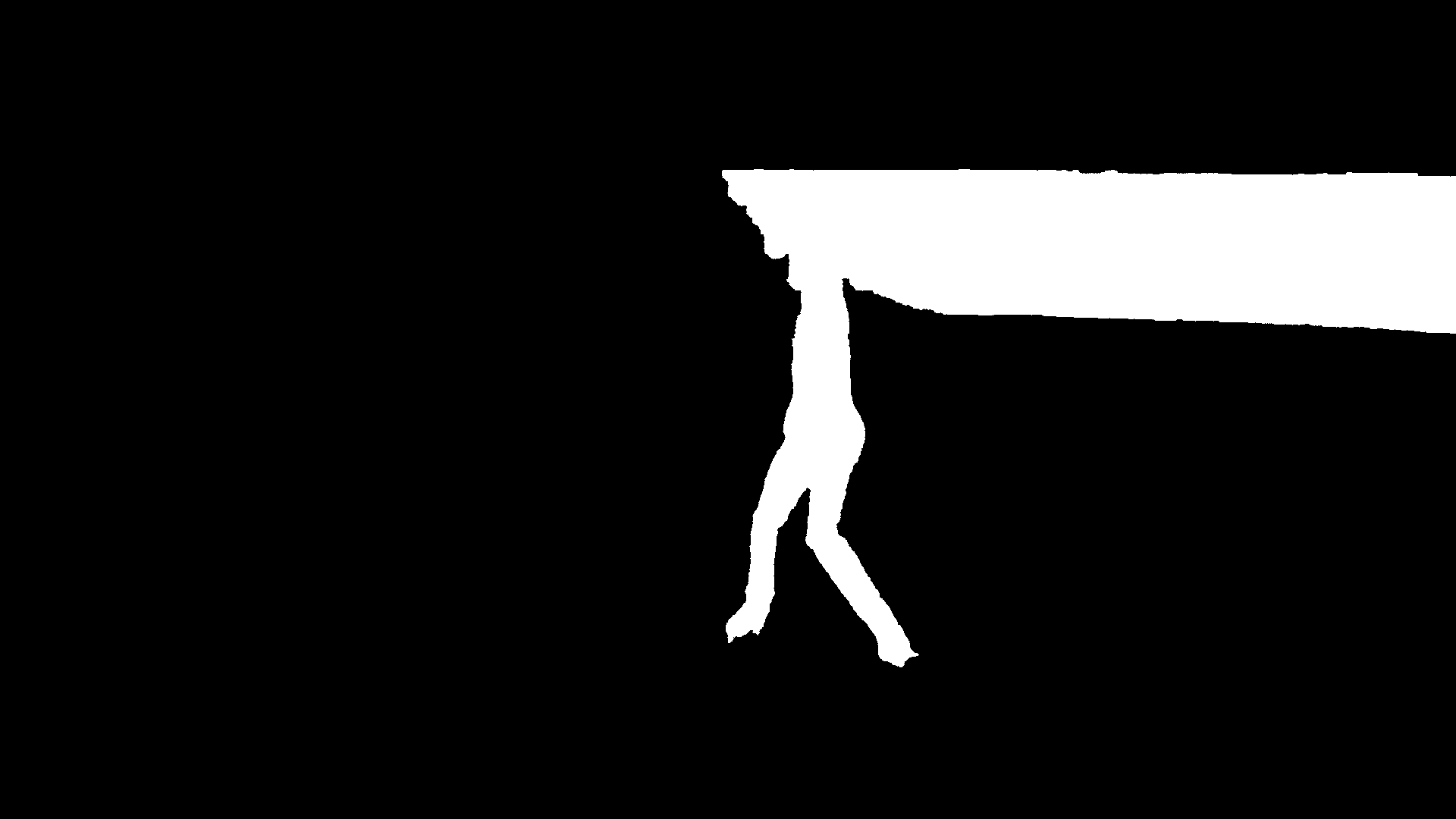} &\hspace{-4mm}
			\includegraphics[width=0.16\linewidth]{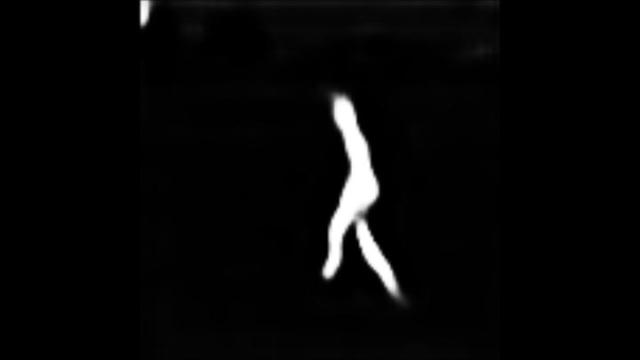} &\hspace{-4mm}
			\includegraphics[width=0.16\linewidth]{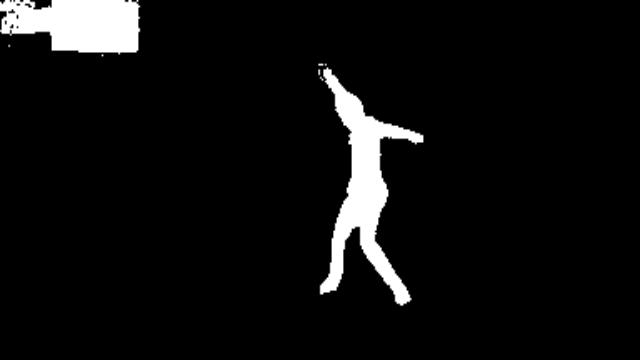} &\hspace{-4mm}
			\includegraphics[width=0.16\linewidth]{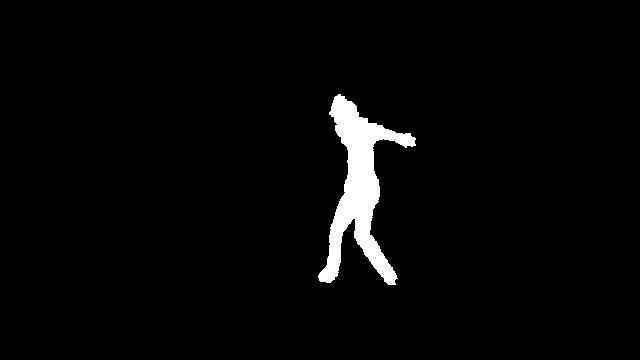} &\hspace{-4mm}
			\includegraphics[width=0.16\linewidth]{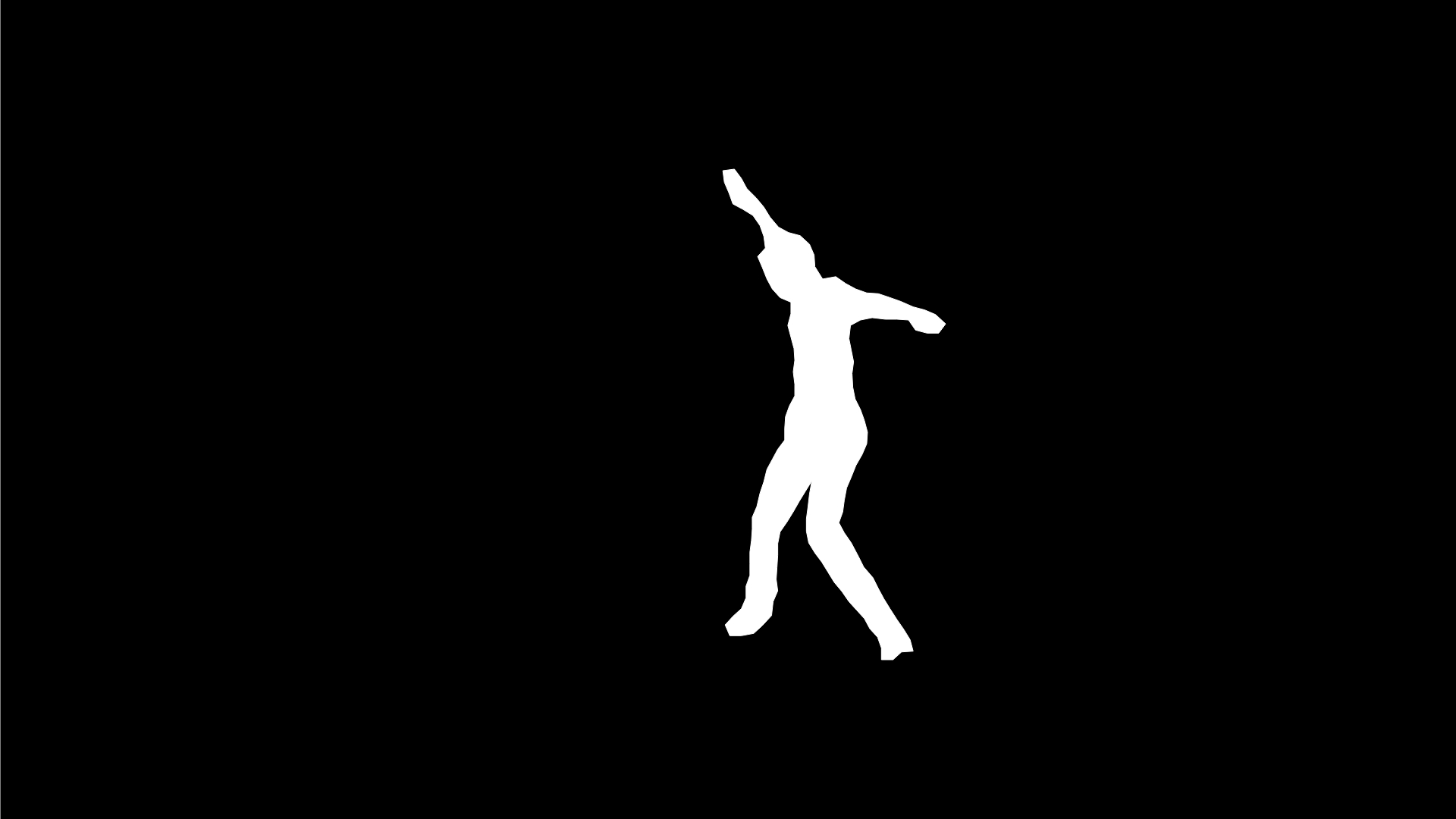} \\ 
			
			\includegraphics[width=0.16\linewidth]{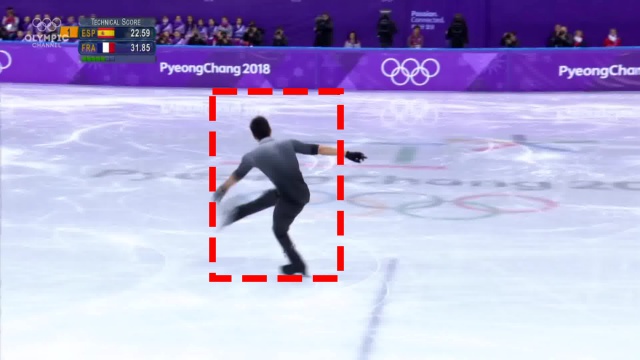} &\hspace{-4.0mm}
			\includegraphics[width=0.16\linewidth]{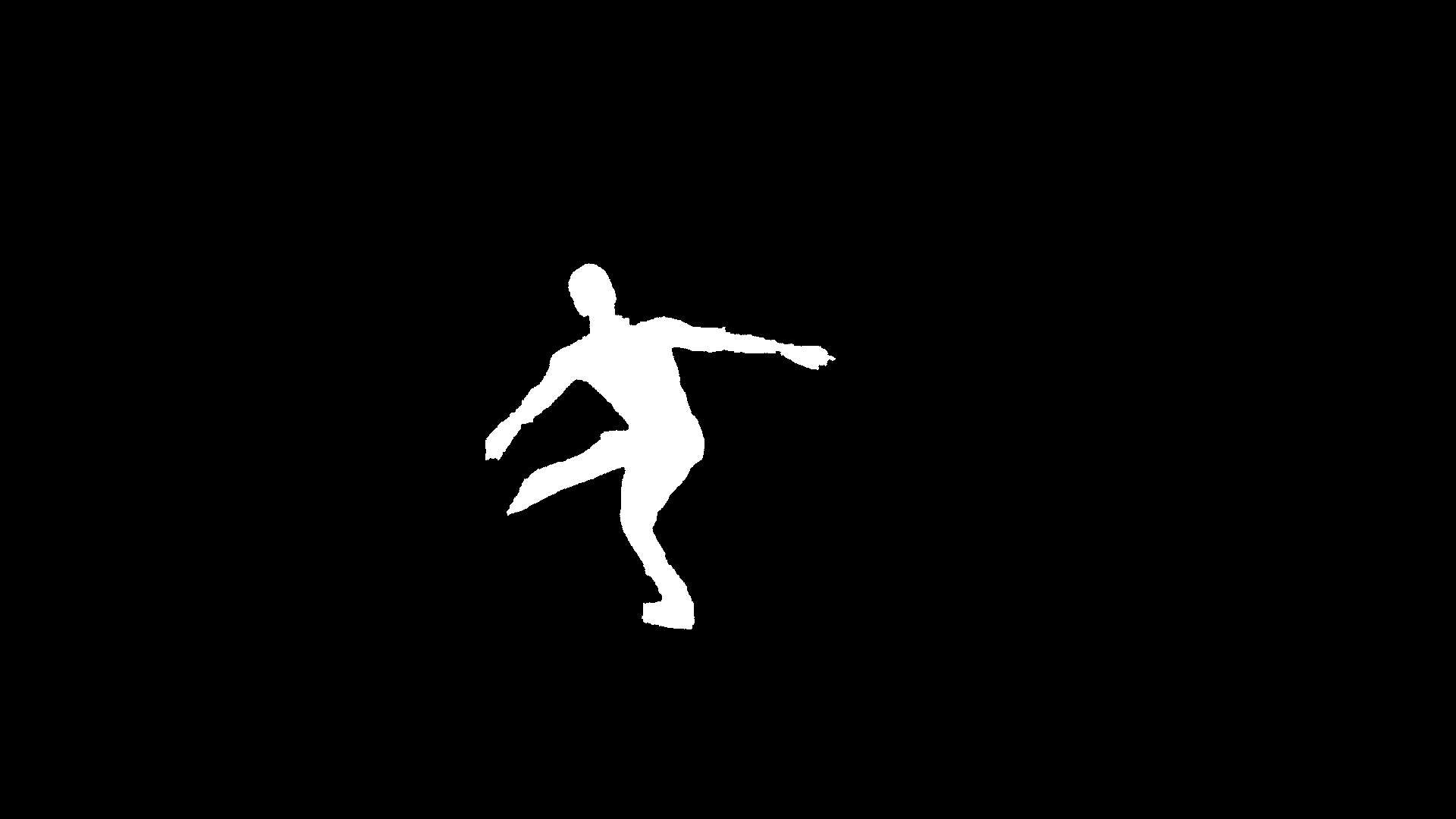} &\hspace{-4mm}
			\includegraphics[width=0.16\linewidth]{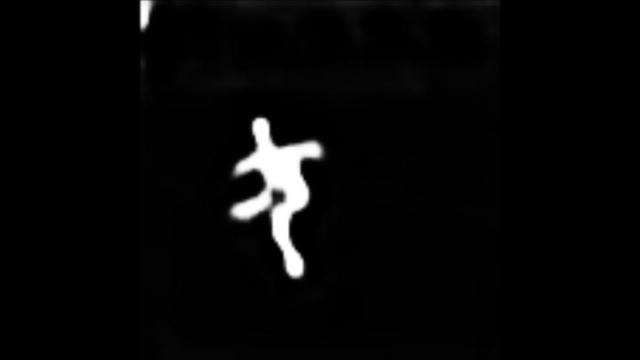} &\hspace{-4mm}
			\includegraphics[width=0.16\linewidth]{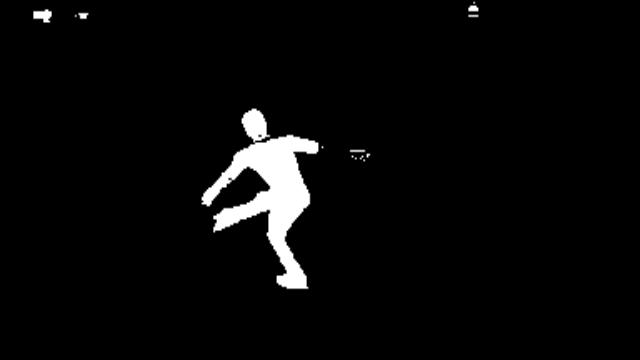} &\hspace{-4mm}
			\includegraphics[width=0.16\linewidth]{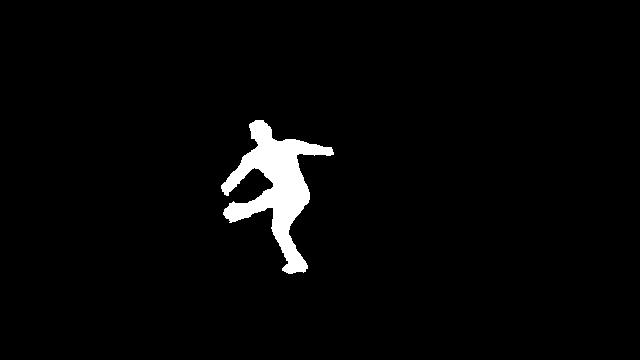} &\hspace{-4mm}
			\includegraphics[width=0.16\linewidth]{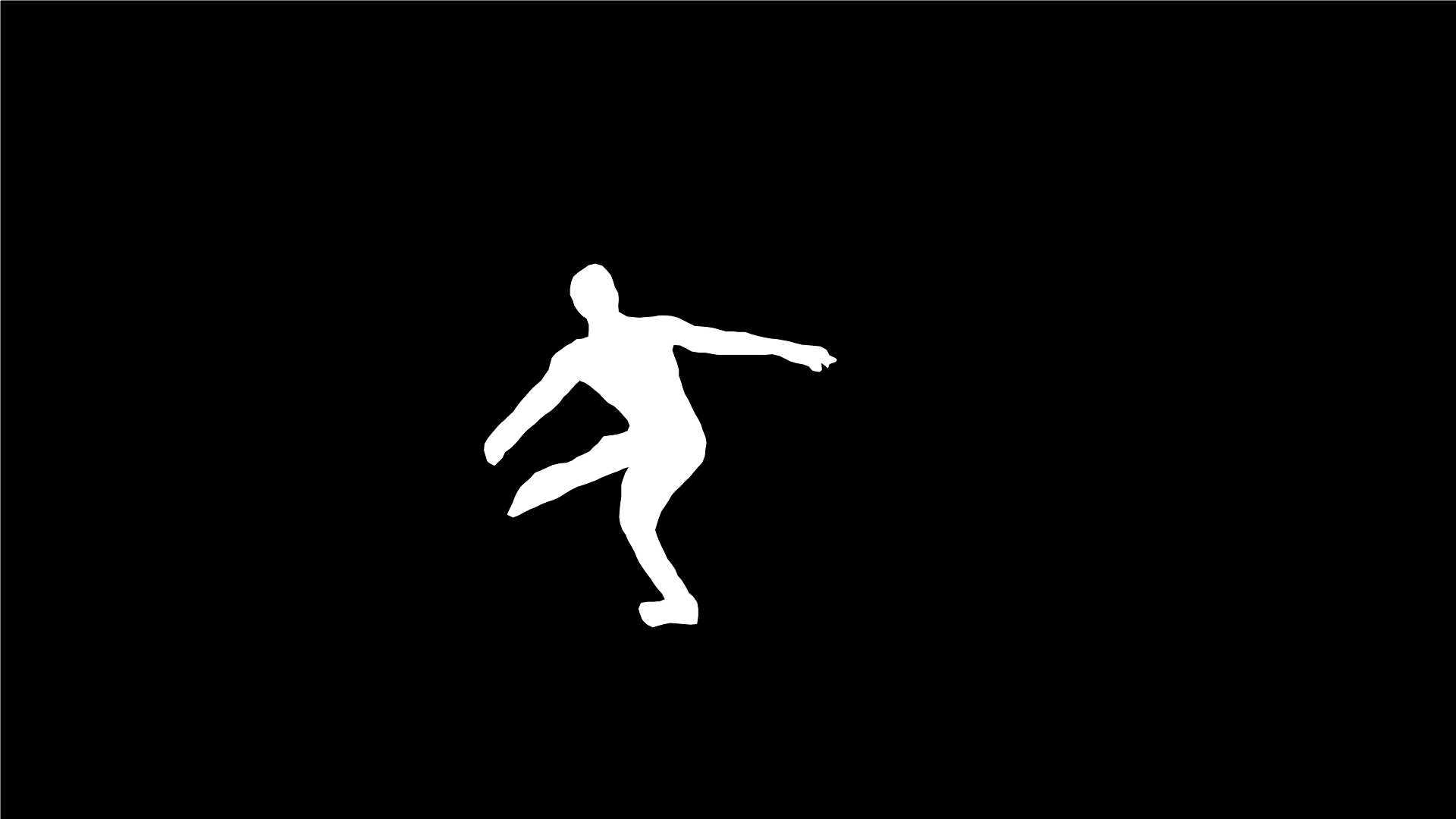} \\ \\

			{\scriptsize  \hspace{-1mm}(a) Input/Ours detection}& \hspace{-1mm}{\scriptsize (b) ARP~\cite{Koh17b}}&\hspace{-4mm}{\scriptsize (c) ReDO~\cite{Chen19a}} &\hspace{-4mm}{\scriptsize (d) Unsup-Mov-Obj~\cite{Yang19c}}&\hspace{-4mm}{\scriptsize (e) Ours w/ optical flow}  &\hspace{-4mm}{\scriptsize (f) Ground truth }  \\ 
	\end{tabular}}
	
\caption{\textbf{Qualitative results on the \iceskating{}.} (a) Our detection result. (b) Segmentation mask prediction of~\cite{Koh17b}. (c) Segmentation mask prediction of~\cite{Chen19a}.  (d) Segmentation mask prediction of~\cite{Yang19c}. (e) Our segmentation result.  (f) Ground truth segmentation mask. Our method is more accurate than~\cite{Koh17b} and~\cite{Yang19c} in terms of removing the background regions.}
\label{fig:figure_skating_qualitative}
\end{figure*}

The quantitative experiments on this dataset are conducted using $50$ manually-segmented test images including diverse and extreme figure skating motions such as axel jump, sit spin and camel spin. In Table~\ref{tbl:results_ski_quantitative}(right), we compare our approach to the self-supervised baselines. Our approach with optical flow outperforms all of them. The overall lower scores of the self-supervised methods on this dataset are due to the motion blur caused by the fast movements of the skaters, the low contrast between the ice and certain body parts and the audience in the background. In Fig.~\ref{fig:figure_skating_qualitative}, we compare the segmentation results of our method to those of the second, third and fourth best-performing methods. Note that our method can accurately detect the skater, even when the scene is cluttered with the audience in the background. The failure cases of our method are mainly due to the low contrast between the ice and the hands and feet of the skater, particularly in extreme spinning poses. Furthermore, the appearance of the skater occasionally matches that of the background people, making it difficult to detect the foreground subject precisely.

Overall, our method that relies on a single image at test time consistently yields the highest scores on all three datasets against other self-supervised methods that operate on single images~\cite{Arandjelovic19,Bielski19,Chen19a,Croitoru19} as well as the ones that require video and use temporal cues at inference time~\cite{Stretcu15,Koh17b,Yang19c}.

\subsection{Comparison to Supervised Models}

In this section we compare our method to MaskRCNN applied in an off-the-shelf manner. Table~\ref{tbl:maskrcnn} reports the results of MaskRCNN trained on the MS-COCO dataset~\cite{Lin14a}, which contains the person class in various sports and daily life scenarios, including skiing and skating. On the \ski{} dataset, our method outperforms MaskRCNN. This demonstrates the benefits of self-supervised learning to handle unusual scenarios, where the data differs significantly from that in the publicly-available datasets. On the \handheld{} and \iceskating{} datasets, MaskRCNN yields the highest scores, which is not surprising as the test sequences look similar to those in the MS-COCO training set. However, many other object categories are not present in the MS-COCO dataset. In those cases, simply exploiting MaskRCNN becomes non-trivial, because it provides class-specific segmentations, and thus cannot directly handle unknown objects.

\begin{figure*}[t]
  \centering
  \begin{tabular}{@{}ccccc@{}}

   \includegraphics[width=0.16\textwidth]{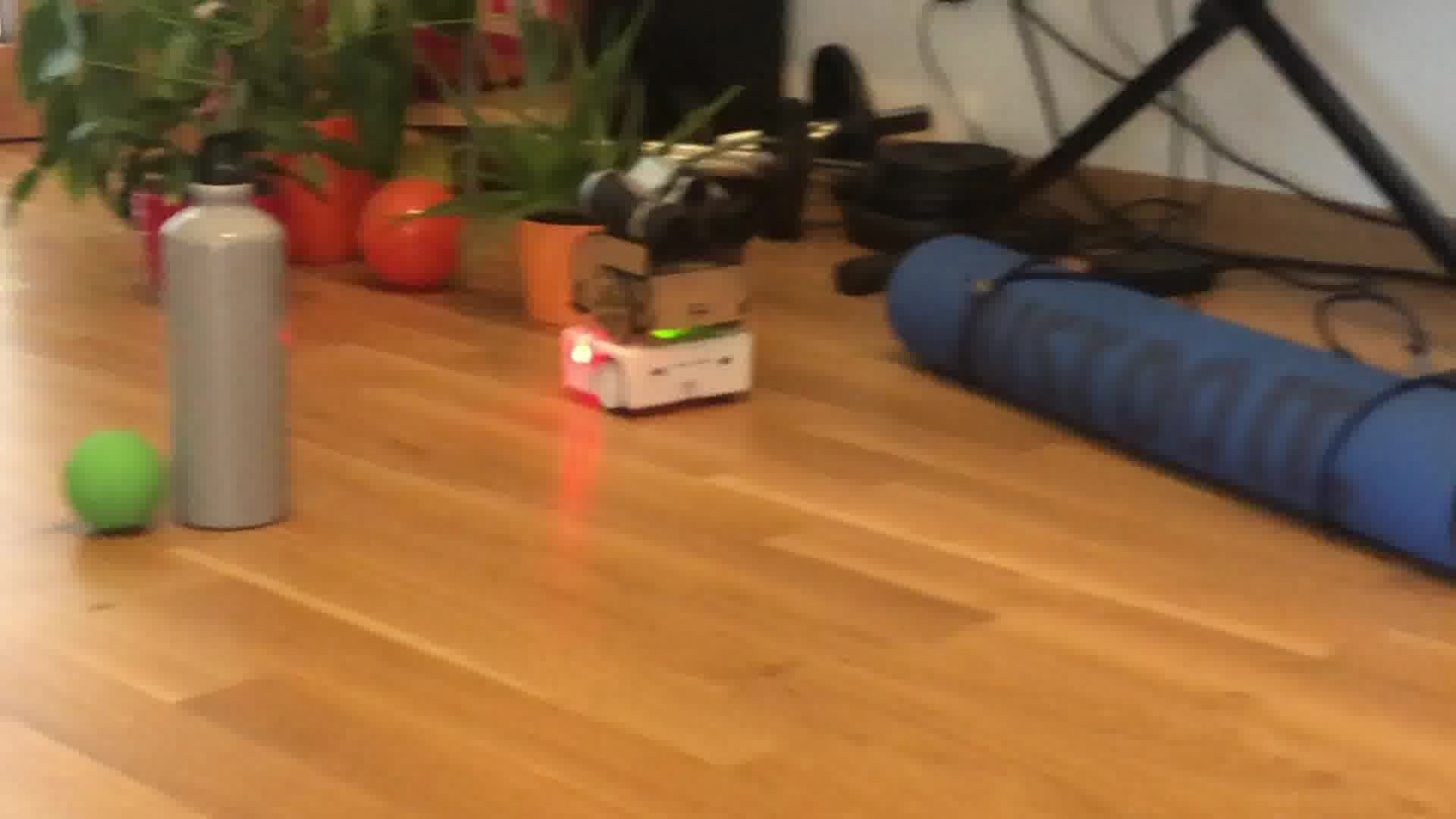} &\hspace{-1mm}
   \includegraphics[width=0.16\textwidth]{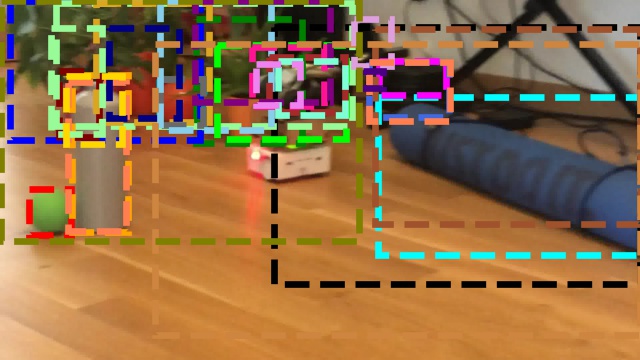} &\hspace{-6mm}
      \includegraphics[width=0.16\textwidth]{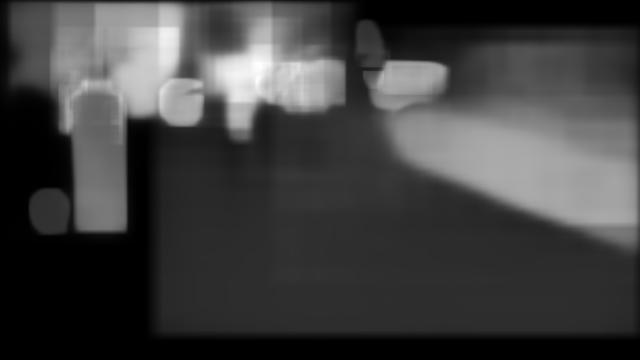} &\hspace{-7mm}
         \includegraphics[width=0.16\textwidth]{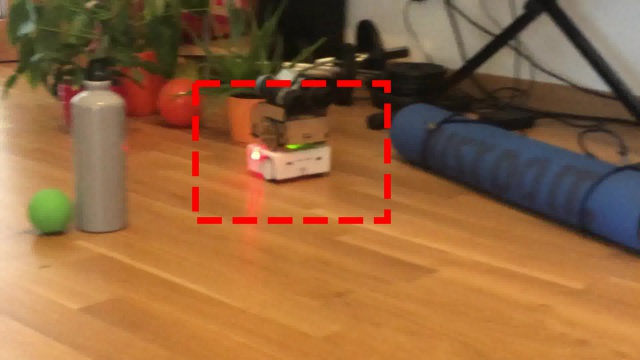} &\hspace{-2mm}
   \includegraphics[width=0.16\textwidth]{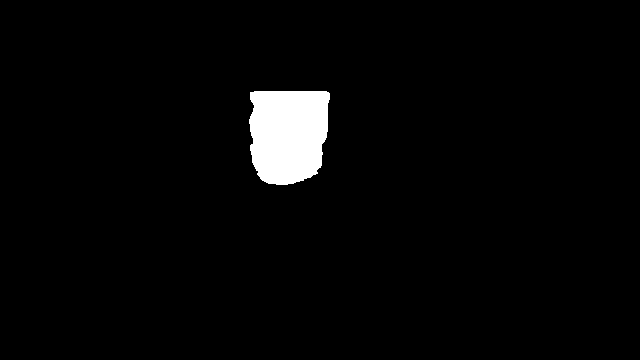} \\

 \includegraphics[width=0.16\textwidth]{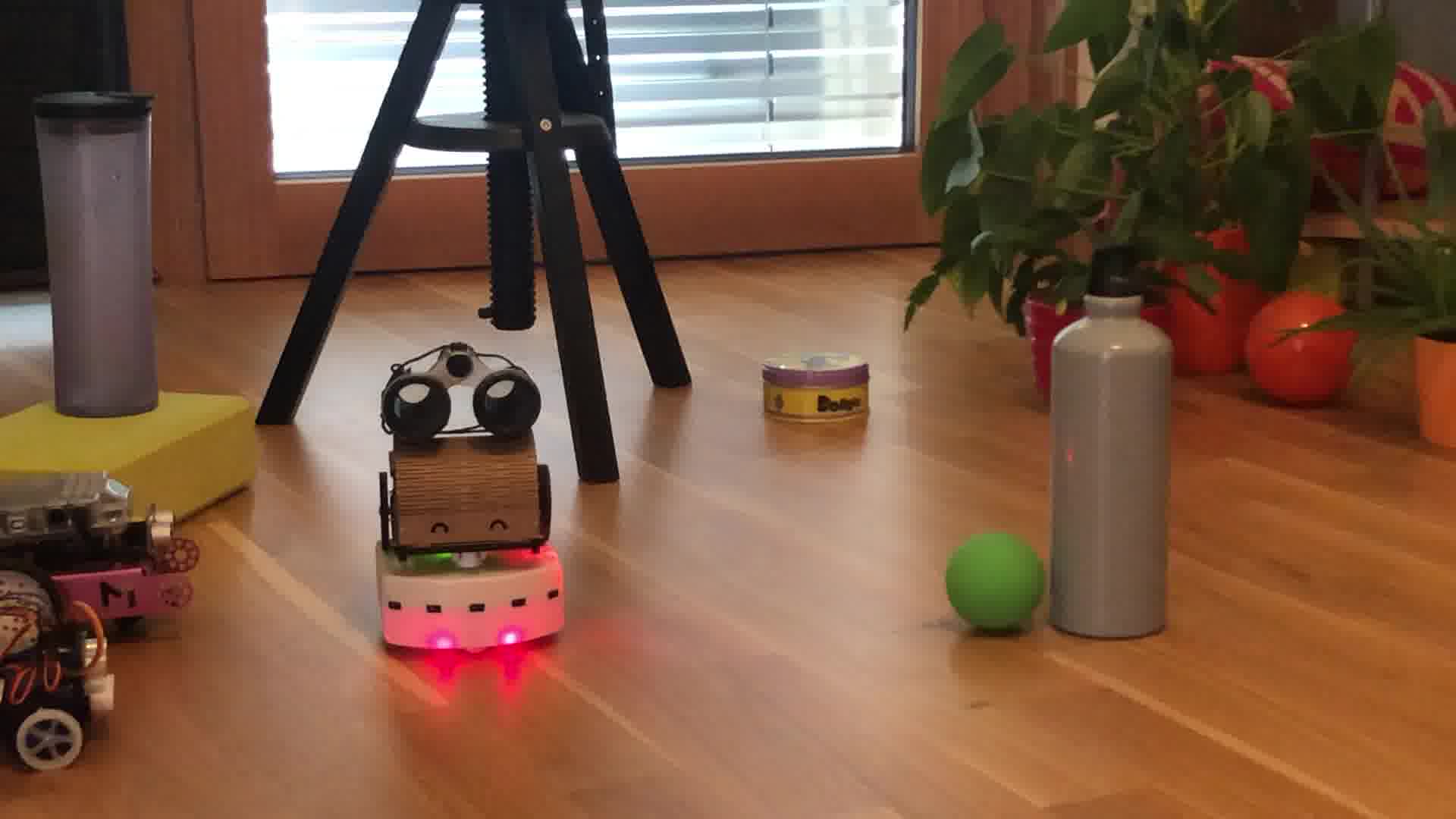} &\hspace{-1mm}
\includegraphics[width=0.16\textwidth]{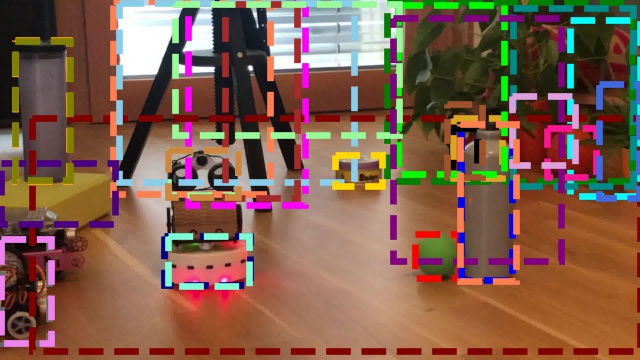} &\hspace{-6mm}
\includegraphics[width=0.16\textwidth]{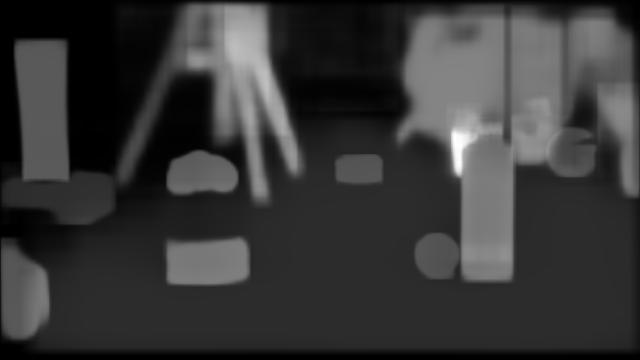} &\hspace{-7mm}
\includegraphics[width=0.163\textwidth]{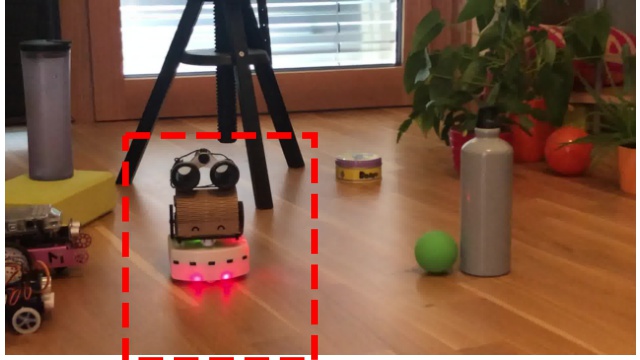} &\hspace{-2mm}
\includegraphics[width=0.16\textwidth]{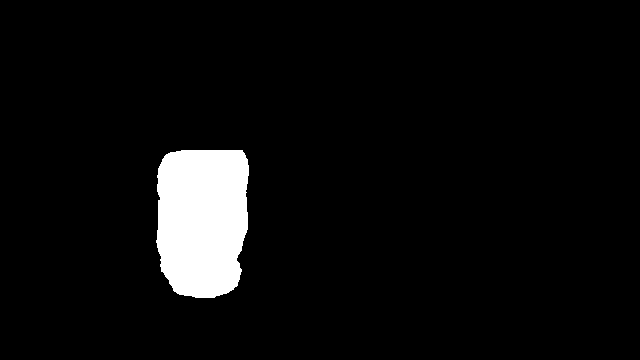} \\ 

 \includegraphics[width=0.16\textwidth]{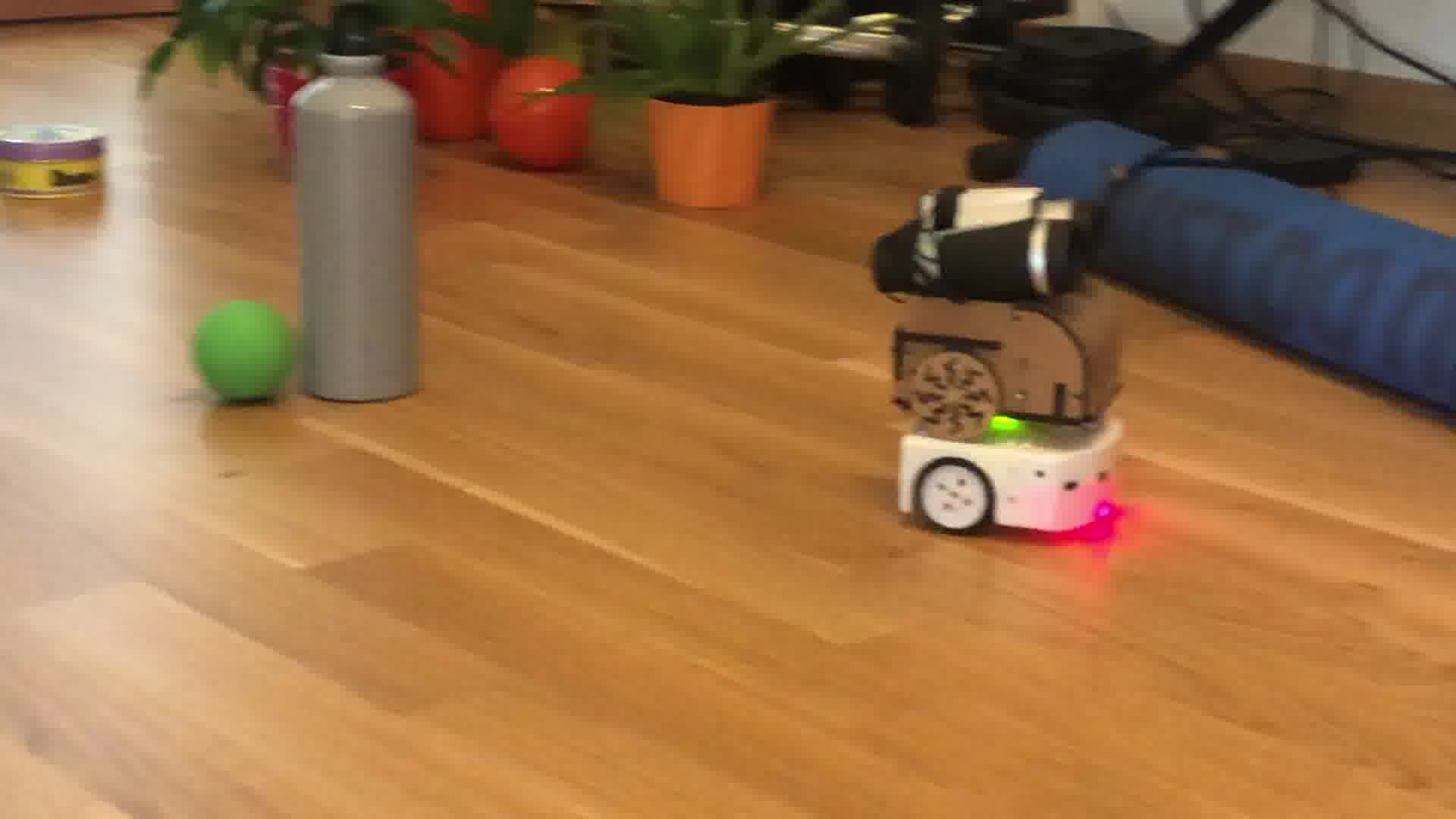} &\hspace{-1mm}
\includegraphics[width=0.16\textwidth]{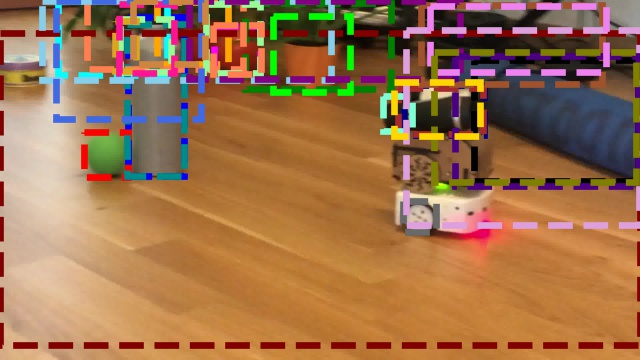} &\hspace{-6mm}
\includegraphics[width=0.16\textwidth]{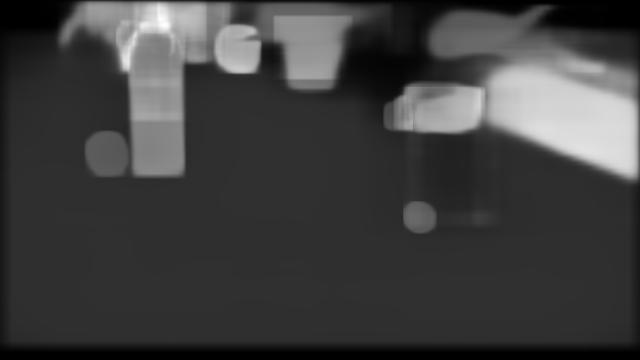} &\hspace{-7mm}
\includegraphics[width=0.16\textwidth]{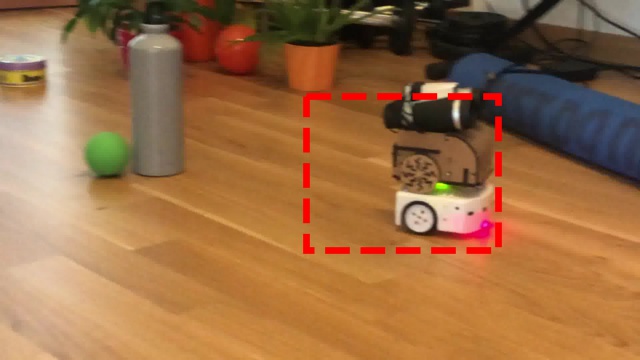} &\hspace{-2mm}
\includegraphics[width=0.16\textwidth]{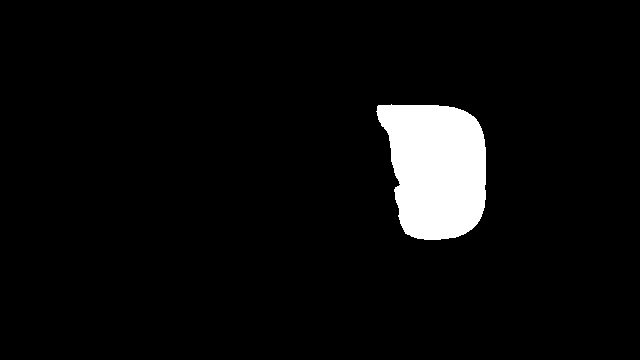} \\  \\
  
  {\small (a)Input} & \hspace{-4mm}{\small (b)MaskRCNN detection} & \hspace{-4mm}{\small (c)MaskRCNN segmentation} & \hspace{-6mm}{\small (d)Ours detection} & \hspace{-4mm}{(e)Ours segmentation}
  \end{tabular}

  \caption{\textbf{Qualitative results of MaskRCNN on a moving robot sequence captured with a handheld camera.} MaskRCNN generally fails to detect the moving robot as a single object and does not yield a segmentation mask with high confidence.}
  \label{fig:robot}
\end{figure*}

To nonetheless evaluate the performance of MaskRCNN in this challenging scenario, we captured an indoor scene featuring many static objects and a moving robot that we aim to segment with a hand-held camera.
Fig.~\ref{fig:robot} compares the detections and segmentation masks output by MaskRCNN for all MS-COCO classes with those obtained with our method. Because the custom robot cannot be associated with any existing MS-COCO category, MaskRCNN tends to split it into multiple objects. Obtaining a consistent mask of the robot would then require parsing these multiple detections. By contrast, our self-supervised approach naturally generalizes to such a previously-unseen object.
\renewcommand{\arraystretch}{1}
\renewcommand{\tabcolsep}{2mm}
\begin{table}[t]
\small
\begin{center}
\resizebox{\columnwidth}{!}{
\begin{tabular}{ccc}
\begin{tabular}{@{}lcc@{}}
&\multicolumn{2}{c}{\ski}\\
\toprule
Method              &J Measure &F Measure\\
\midrule
MaskRCNN  \cite{He17a}						  &\bf{0.73}	   &0.77	\\
ARP \cite{Koh17b}									&0.72	&0.82  \\
Unsup-Mov-Obj \cite{Yang19c}						& 0.66 	& 0.76 \\
Ours w/ optical flow + CRF								&\bf{0.73}		&\bf{0.83} \\
\bottomrule
\end{tabular}&

\begin{tabular}{@{}lcccc@{}}
\multicolumn{2}{c}{\handheld}\\
\toprule
& J measure & F measure \\
\midrule
  & \bf{0.83}  & \bf{0.95}	   	\\
  &0.60	  &0.68			\\
& 0.75			&0.83			\\
& 0.76			&0.85			\\

\bottomrule
\end{tabular}&

\begin{tabular}{@{}lcccc@{}}
	\multicolumn{2}{c}{\iceskating}\\
	\toprule
	& J measure & F measure \\
	\midrule
	&\bf{0.87}  & \bf{0.96}	   	\\
	&0.56	&0.69		\\
	& 0.68		&0.85			\\
	& 0.71		&0.86			\\
	
	\bottomrule
\end{tabular}\\
\end{tabular}
}
\end{center}
\caption{\textbf{MaskRCNN segmentation results on the \ski{}, \handheld{} and \iceskating{} datasets}. The direct application of off-the-shelf MaskRCNN on \handheld{} and \iceskating{} datasets outperforms the self-supervised methods in Table~\ref{tbl:results_ski_quantitative} whereas on \ski{} dataset with unusual motions, our method reaches the maximum F score and is on par with MaskRCNN in J score. This outcome is expected since MaskRCNN is trained on MS-COCO dataset that includes person class as one of the training categories.}
\label{tbl:maskrcnn}
\end{table}

\subsection{Ablation Study}

\label{sec:ski_dataset_analysis}
In Table~\ref{tbl:ski_ablation_regularizer_pretraining}, we investigate the effectiveness of different mask priors introduced in Section~\ref{section:extension} and ImageNet pre-training on the validation part of the \ski{} dataset. Although $\loss_1$ yields better segmentation masks than $\loss_2$, it tends to suppress the mask values too strictly, which causes convergence problems. This is mitigated by our $\loss_v$ prior, which achieves the highest scores in all measures, with consistently reliable results. This demonstrates that imposing regularization on the segmentation masks allows us to obtain sharper masks, removing the noise around the foreground object. We repeated the \ski{} experiment without optical flow extension four times with the best-performing configuration
and computed the mean and std on the validation sequences; the J- and F-measure are consistent, respectively, $0.67 \pm 0.004, 0.73 \pm 0.006$.

Table~\ref{tbl:ski_ablation_regularizer_pretraining} also shows the comparison of using ImageNet or self-supervised weights for network initialization, with only a small performance drop for the latter.

\renewcommand{\arraystretch}{1}
\renewcommand{\tabcolsep}{2mm}

\begin{table}
	\small
	\begin{center}
		\resizebox{1\columnwidth}{!}{
			\begin{tabular}{cc}
				\begin{tabular}{@{}lcccc@{}}
					&\multicolumn{2}{c}{\ski}\\
					\toprule
					Setting             				&J Measure &F Measure\\
					\midrule

					Ours w/o optical flow w/o prior						  &  0.51    &  0.53    \\
					Ours w/o optical flow w/ $\loss_2$ prior					 &0.61    &0.69  \\
					Ours w/o optical flow w/ $\loss_1$ prior					 &0.62    &0.69    \\
					Ours w/o optical flow w/ $\loss_v$ prior					 & \bf 0.67   & \bf 0.73    \\
					\midrule
					No ImageNet pre-training, $\loss_v$ prior    &0.60    &0.63      \\
					Unsupervised pre-training~\cite{Wu18c}, $\loss_v$ prior   &0.62   & 0.68    \\
					\bottomrule
				\end{tabular}

			\end{tabular}
		}
	\end{center}

	\caption{\textbf{Analysis of the mask prior effect and ImageNet pre-training on the \ski{} validation sequences}. We demonstrate the influence of using mask priors to supress the noise surrounding the foreground object and have clear-cut masks. At the bottom part of the table we show the results of using random weights and features from ~\cite{Wu18c} instead of using weights from ImageNet pre-training.}
	\label{tbl:ski_ablation_regularizer_pretraining}
\end{table}

Furthermore, Table~\ref{tbl:ski_ablation_bbox_size_lambda} compares the performance of our method for different values of hyper-parameters, where the subscript of $\vb$ corresponds to the minimum and maximum size of the bounding box and $\lambda$ used in our $\loss_v$ prior is the percentage of the pixels that should be activated in the segmentation mask.

\renewcommand{\arraystretch}{1}
\renewcommand{\tabcolsep}{2mm}
\begin{table}
	\small
	\begin{center}
		\resizebox{0.75\columnwidth}{!}{
			\begin{tabular}{cc}
				\begin{tabular}{@{}lcccc@{}}
					&\multicolumn{2}{c}{\ski}\\
					\toprule
					Setting             																		&J Measure &F Measure\\
					\midrule

					$\vb_{[0.1,0.5]}$, $\loss_{v, \lambda=0.0005}$									& 0.55    & 0.55   \\
					$\vb_{[0.1,0.5]}$, $\loss_{v, \lambda=0.001}$								   & 0.57    & 0.62 \\
					$\vb_{[0.1,0.5]}$, $\loss_{v, \lambda=0.005}$									   & 0.54    &  0.56    \\
					\midrule

					$\vb_{[0.20,0.5]}$, $\loss_{v, \lambda=0.0005}$								  &0.61   &0.70  \\
					$\vb_{[0.20,0.5]}$, $\loss_{v, \lambda=0.001}$								  &  0.60    &  0.65    \\
					$\vb_{[0.20,0.5]}$, $\loss_{v, \lambda=0.005}$								   &  0.61   &  0.67    \\
					\midrule

					$\vb_{[0.30,0.5]}$, $\loss_{v, \lambda=0.0005}$									&  0.57    &  0.64    \\
					$\vb_{[0.30,0.5]}$, $\loss_{v, \lambda=0.001}$									&  0.57    &  0.64    \\
					$\vb_{[0.30,0.5]}$, $\loss_{v, \lambda=0.005}$									 &  0.57    &  0.63    \\
					\midrule
					$\vb_{[0.20,0.60]}$, $\loss_{v, \lambda=0.0005}$									&  0.61    &  0.69    \\
					$\vb_{[0.20,0.60]}$, $\loss_{v, \lambda=0.001}$									&  0.61    &  0.68    \\
					$\vb_{[0.20,0.60]}$, $\loss_{v, \lambda=0.005}$									&  0.62    &  0.68    \\
					\midrule
					$\vb_{[0.20,0.70]}$, $\loss_{v, \lambda=0.0005}$									&  0.62    &  0.67    \\
					$\vb_{[0.20,0.70]}$, $\loss_{v, \lambda=0.001}$									&  0.59    &  0.66    \\
					$\vb_{[0.20,0.70]}$, $\loss_{v, \lambda=0.005}$									&  0.62    &  0.65    \\
					\midrule
					$\vb_{[0.20,0.80]}$, $\loss_{v, \lambda=0.0005}$									& 0.61    &  0.68    \\
					$\vb_{[0.20,0.80]}$, $\loss_{v, \lambda=0.001}$									&  \bf0.67    &  \bf0.73    \\
					$\vb_{[0.20,0.80]}$, $\loss_{v, \lambda=0.005}$									&  0.61    &  0.66    \\
					\bottomrule
				\end{tabular}

			\end{tabular}
		}
	\end{center}
	\caption{\textbf{Hyper-parameter study on the \ski{} validation sequences}. In this table we analyze the effectiveness of our hyper-parameter choice for the minimum and maximum bounding box sizes (given in square brackets as $\vb_{[scale_{min},scale_{max}]}$) as well as the threshold $\lambda$ for the $\loss_v$ loss. We conduct these experiments using our approach without optical flow. }
	\label{tbl:ski_ablation_bbox_size_lambda}
\end{table}

In the appendix, we show additional detection and segmentation results from the test set of the \ski{}, \handheld{} and \iceskating{} datasets with and without the optical flow extension and CRF post-processing.

\parag{People in a Controlled Environment.}
We evaluate different aspects of our approach using the
\textbf{Human3.6m} dataset~\cite{Ionescu14a} that comprises 3.6 million frames and 15 motion classes. It features 5 subjects for training and 2 for validation, seen from different viewpoints against a static background and with good illumination.

\begin{figure*}
  \centering
  \begin{tabular}{@{}cccccc@{}}
   \hspace{-3mm}
\includegraphics[width=0.1707\textwidth]{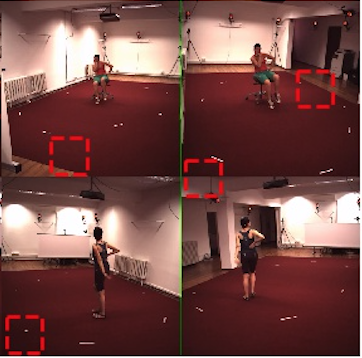} &\hspace{-4.5mm}
\includegraphics[width=0.167\textwidth]{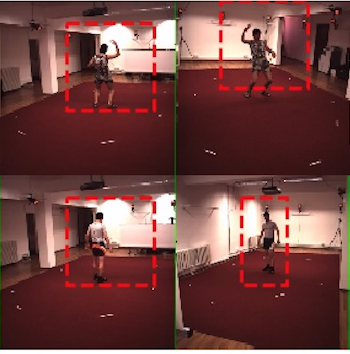} &\hspace{-4.5mm}
  \includegraphics[width=0.1705\textwidth]{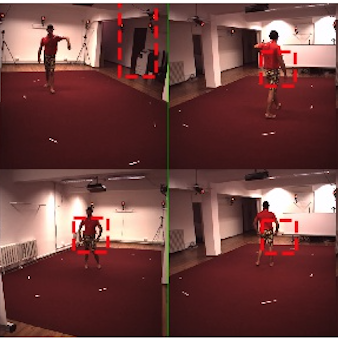} &\hspace{-4.5mm}
  \includegraphics[width=0.171\textwidth]{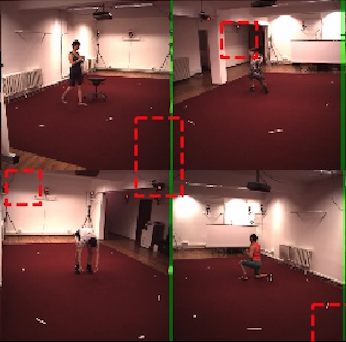} &\hspace{-4.5mm}
    \includegraphics[width=0.17\textwidth]{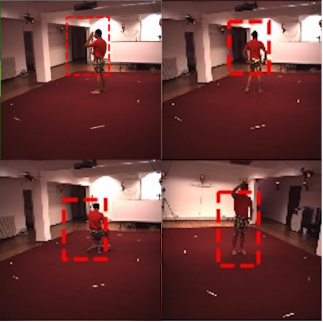} &\hspace{-4.5mm}
  \includegraphics[width=0.1694\textwidth]{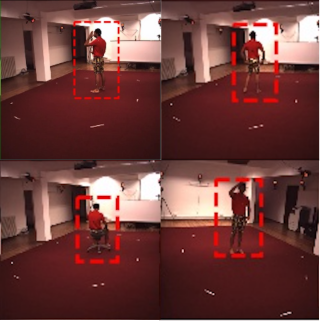} \\
  {\small (a) } & {\small (b) }& {\small (c) } & {\small (d) } & {\small (e) } & {\small (f) } \\ 
  \end{tabular}
  \caption{\textbf{Ablation study on H36M.} (a) Uniform sampling does not converge. (b) Joint training of $\objFG$ and $\objBG$ (c) only $\objBG$ (d) direct regression of a single bounding box using $\objFG$ and $\objBG$ (e) Gumbel-Softmax (f) Ours. }
  \label{fig:h36m ablation}
\end{figure*} %

On this dataset, we first study the importance of our model choices for training and probabilistic inference.
As shown in Fig.~\ref{fig:h36m ablation}(a), using uniform sampling instead of importance sampling does not converge. Fig.~\ref{fig:h36m ablation}(b) illustrates that
joint training of $\cD$ with $\objFG$ and $\objBG$, instead of our separate one, produces bounding boxes that are too large. Fig.~\ref{fig:h36m ablation}(c) shows that using only the background objective leads to small detections that miss the subject and (d) that direct regression without multiple candidates diverges. These failure cases are representative of the behavior on the whole dataset. To explore an alternative strategy to Monte Carlo-based sampling, we replaced the importance sampling in our method with the categorical reparameterization used in~\cite{Crawford19}. Since both strategies approximate the same objective, they had similar outcomes with a difference in the convergence speed and detection performance. To this end, we tried Gumbel-Softmax distribution~\cite{Jang16}. We found out that setting the temperature to $0.1$ yielded the best results. Increasing this value has a similar effect as increasing the $\epsilon$ in Eq.~\eqref{eq:epsilon} and approaches uniform sampling. Our experiments show that Gumbel-Softmax based categorical reparameterization did not lead to faster convergence and in fact degraded the detection performance as shown in Fig.~\ref{fig:h36m ablation}(e). Our method delivers a mAP$_{0.5}$ score of $0.58$ which is significantly higher than the mAP$_{0.5}$ score of $0.30$ obtained by using Gumbel-Softmax as our sampling strategy. Furthermore, our importance sampling approach is simpler than~\cite{Crawford19} and is an unbiased estimator. It does not need custom layers that behave differently in the forward and backwards passes during optimization, which is the case for the Gumbel-Softmax categorical reparameterization. Please note that direct comparison to~\cite{Crawford19} is not possible since it requires monochromatic backgrounds. Therefore, it does not apply to the \ski{}, \handheld{} and \iceskating{} datasets and was demonstrated only on simple synthetic cases, such as MNIST and Atari games, with multiple objects that go beyond the scope of our current approach. Finally, Fig.~\ref{fig:h36m ablation}(f) demonstrates that our full model using the separate training strategy and importance sampling can accurately detect the person and estimate tighter bounding boxes.

\subsection{Discussion}

\begin{figure}
  \centering
  \begin{tabular}{@{}ccccc@{}}

   \includegraphics[width=0.095\textwidth]{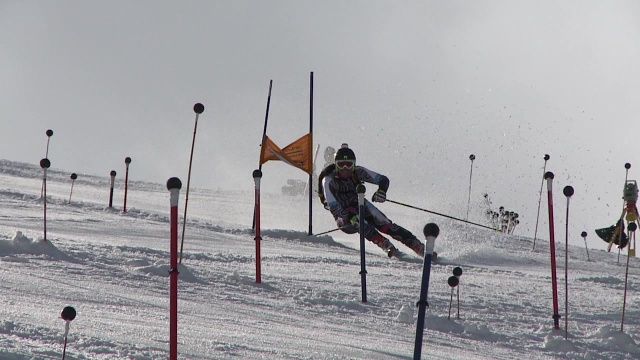} & \hspace{-5mm}
   \includegraphics[width=0.095\textwidth]{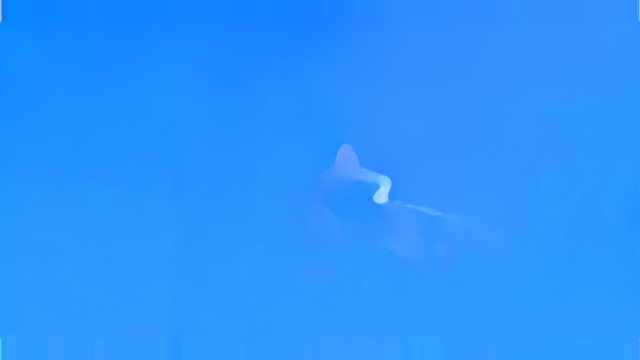} &\hspace{-5mm}
   \includegraphics[width=0.095\textwidth]{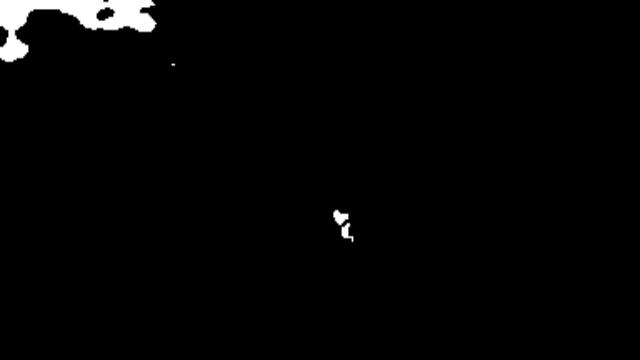} &\hspace{-4.5mm}
   \includegraphics[width=0.095\textwidth]{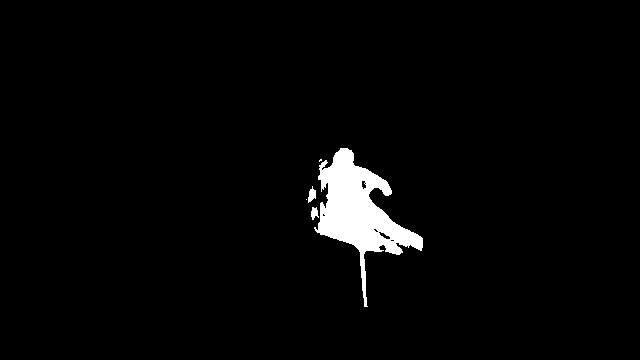} &\hspace{-5.5mm}
   \includegraphics[width=0.095\textwidth]{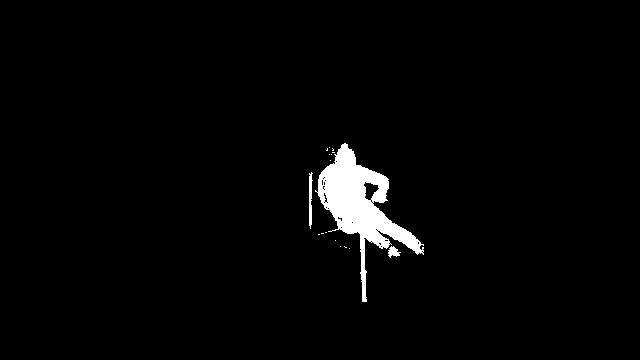} \\

  \includegraphics[width=0.095\textwidth]{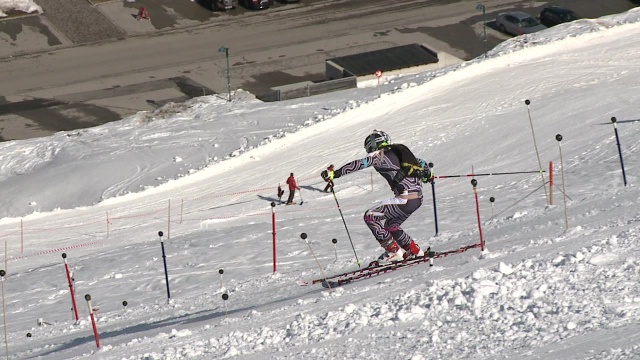} &\hspace{-5mm}
  \includegraphics[width=0.095\textwidth]{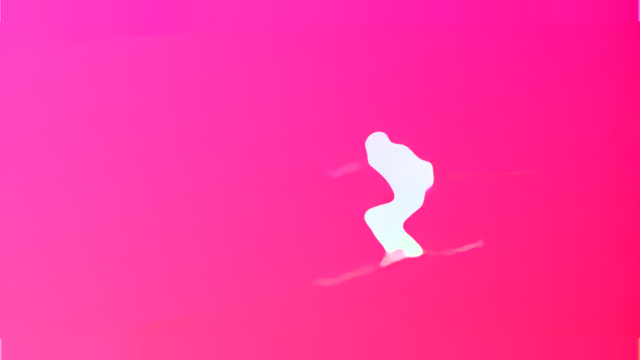} &\hspace{-5mm}
  \includegraphics[width=0.095\textwidth]{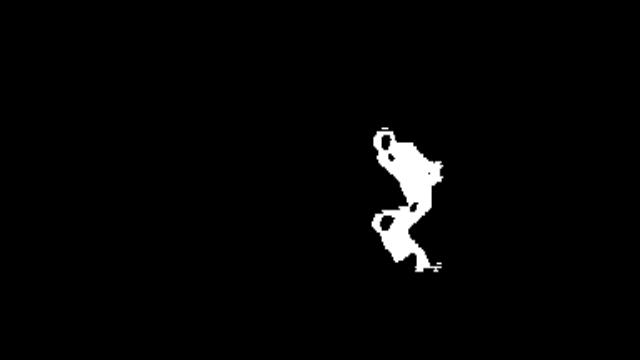} &\hspace{-4.5mm}
  \includegraphics[width=0.095\textwidth]{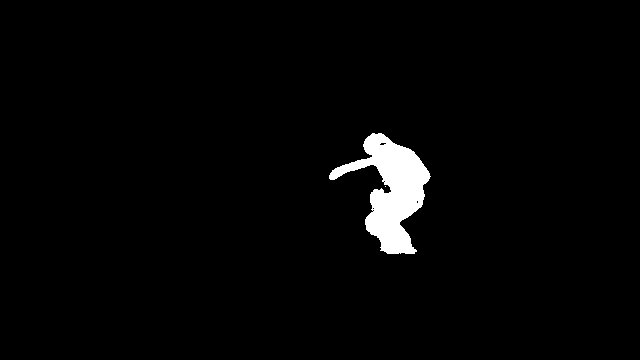} &\hspace{-5.5mm}
  \includegraphics[width=0.095\textwidth]{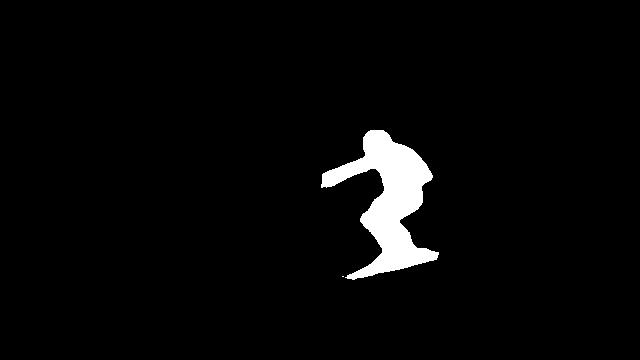} \\ \\
  
  {\small (a)Input} & \hspace{-6mm}{\small (b)Optical flow} &  \hspace{-4mm}{\small (c)~\cite{Yang19c}} &  \hspace{-4mm}{\small (d)Ours} &  \hspace{-6.5mm}{\small (e)Ours w/ flow}
  \end{tabular}

  \caption{\textbf{Optical flow failure.} When the optical flow image does not cover the entire object due to the limited movement, our method can still segment the moving object from a single RGB image whereas~\cite{Yang19c} might yield poor results. Note that here, we present the raw segmentation predictions of~\cite{Yang19c} and ours before the post-processing step in order to show the effect of using the optical flow. Top row: only the upper left body of the skier stands against the background in the optical flow image causing~\cite{Yang19c} to fail. Bottom row: the left arm of the skier has a similar movement as the background and therefore it is not recovered by~\cite{Yang19c} while our method can accurately detect it.}
  \label{fig:optical_flow_failure}
\end{figure}

\parag{Optical flow.}
As noted in~\cite{Yang19c}, the motion-based segmentation methods that require computing the optical flow between consecutive images can be error-prone due to the irregular or insufficient movement of the object. This gives us leverage against optical flow based approaches since our method can reliably detect the foreground object from single RGB images and uses optical flow only as an extension during training time. In Fig.~\ref{fig:optical_flow_failure}, we present possible failure cases that can occur when the optical flow partially covers the object due to its static parts. It can be seen that our method can accurately segment the object in this case.

\begin{figure}
  \centering

\begin{tabular}{@{}ccc@{}}%
\includegraphics[width=0.28\linewidth]{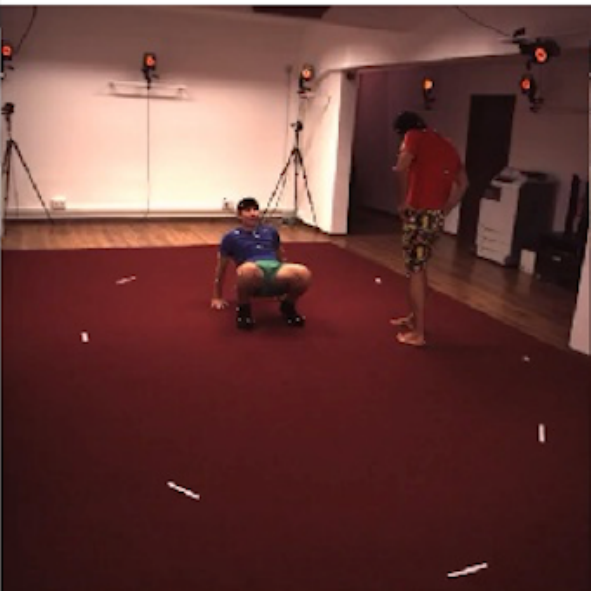} & 
\includegraphics[width=0.28\linewidth]{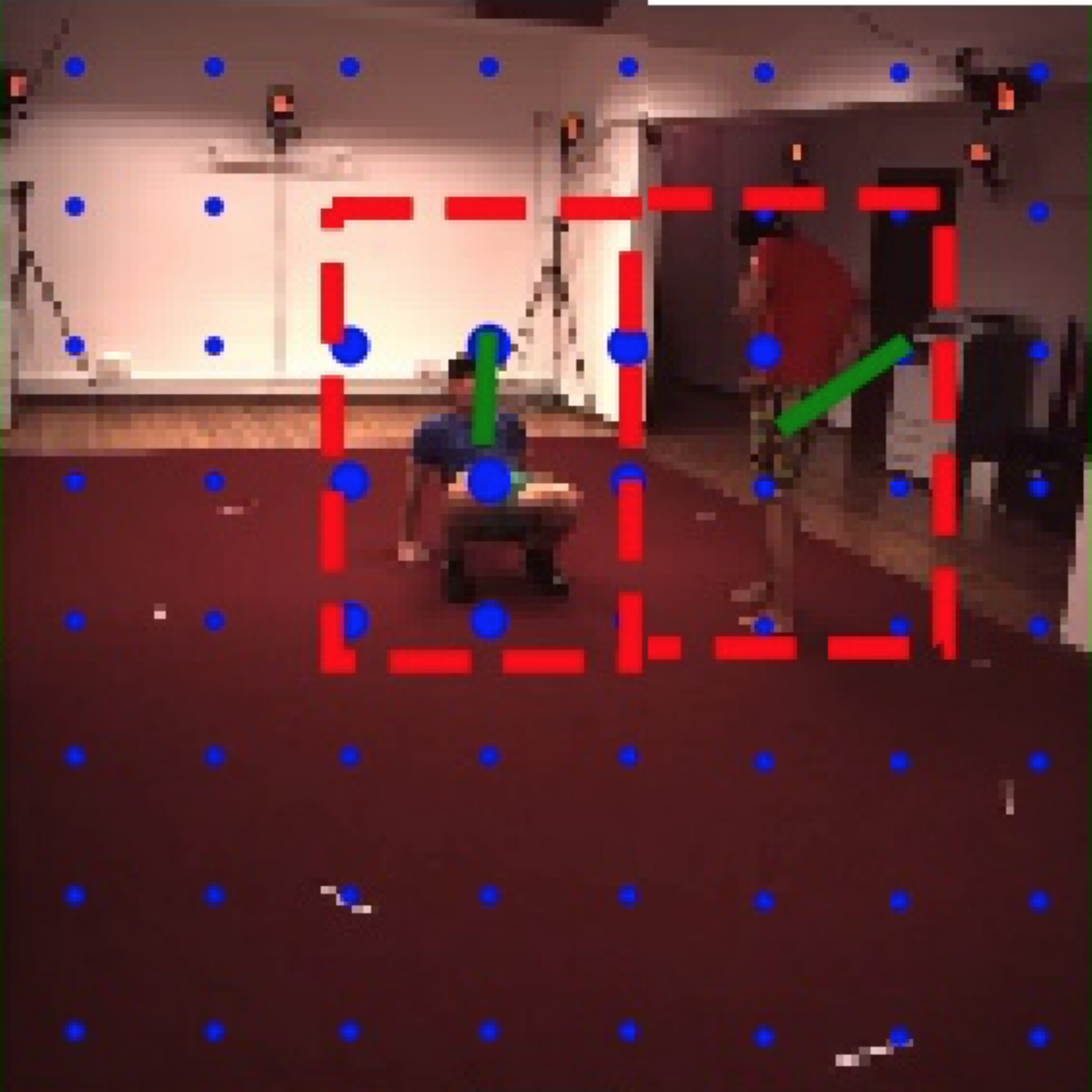} &
\includegraphics[width=0.28\linewidth]{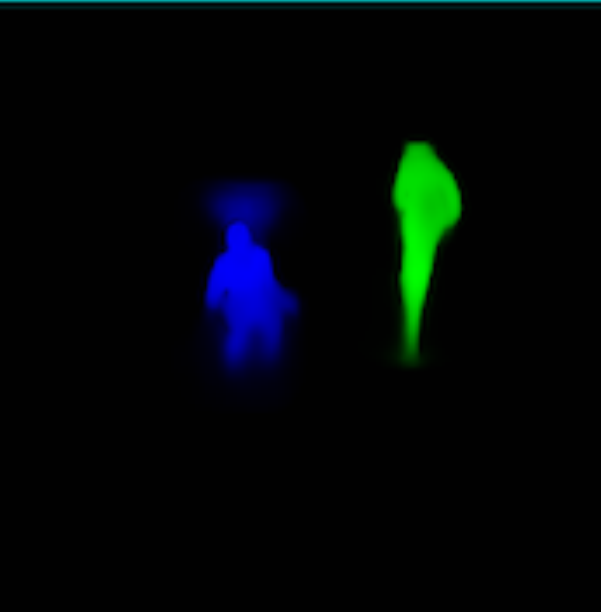} \\
\includegraphics[width=0.28\linewidth]{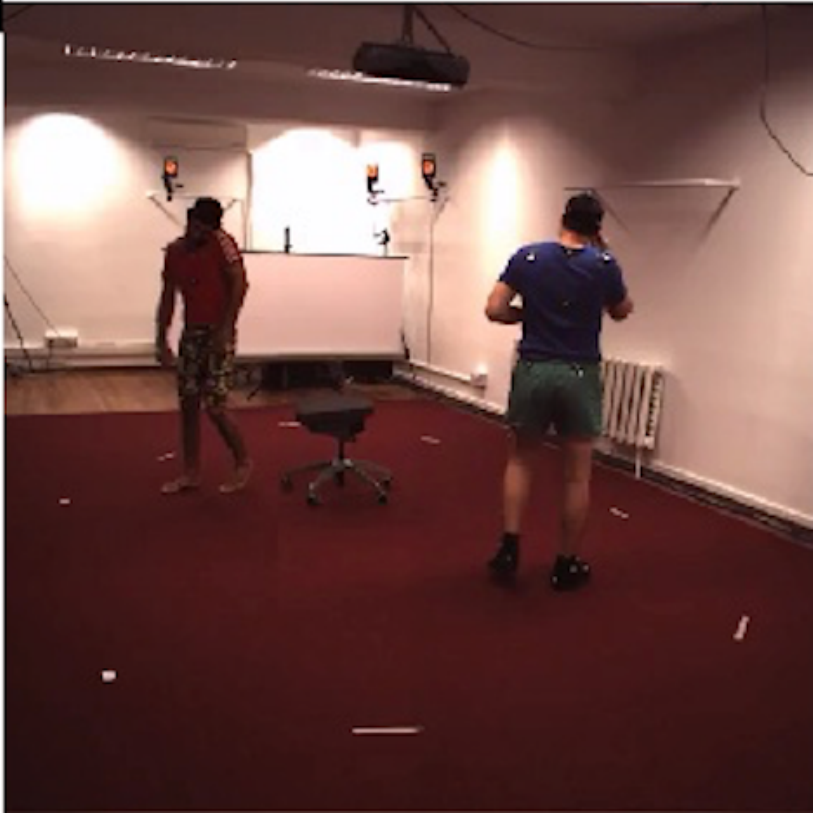} &
\includegraphics[width=0.28\linewidth]{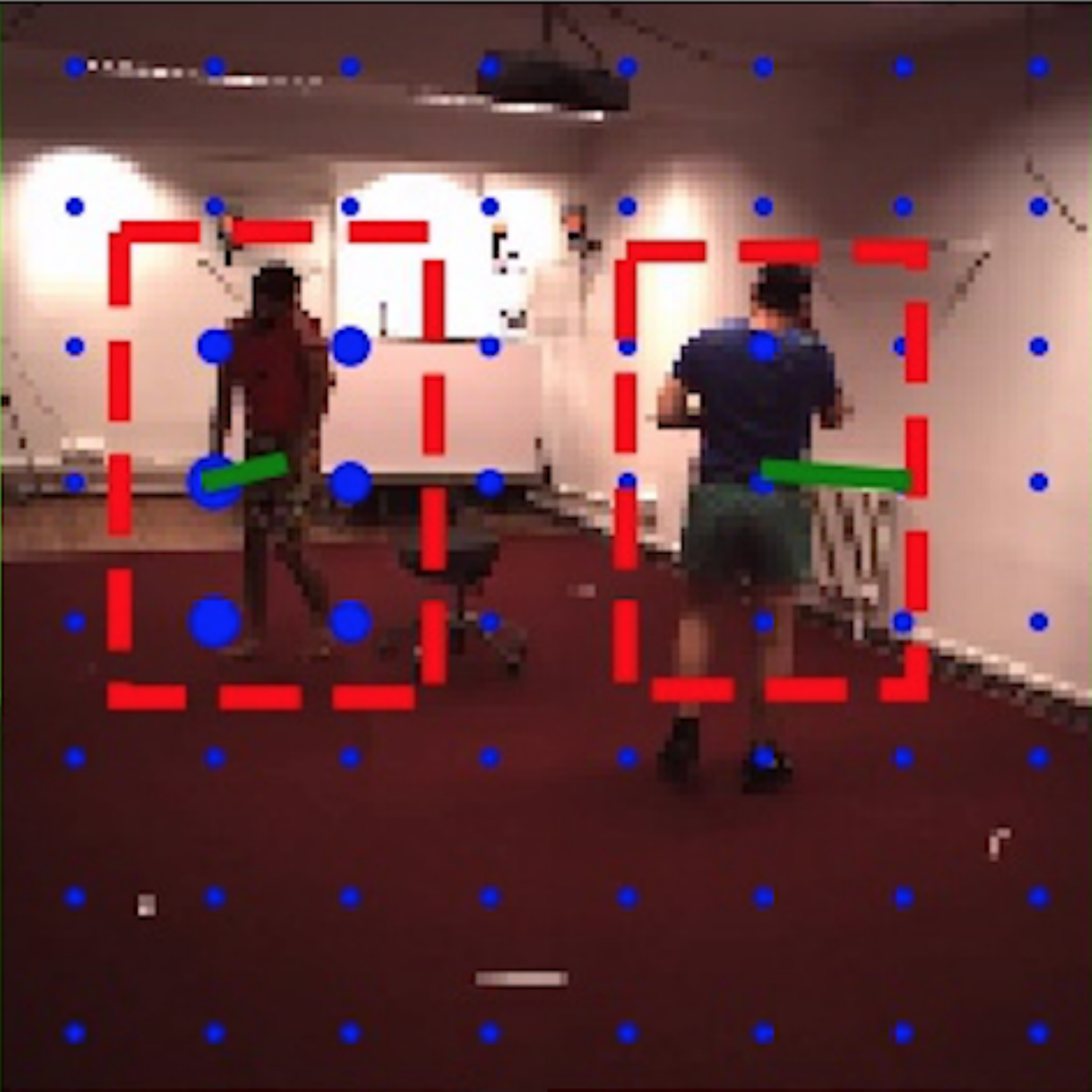} &
\includegraphics[width=0.28\linewidth]{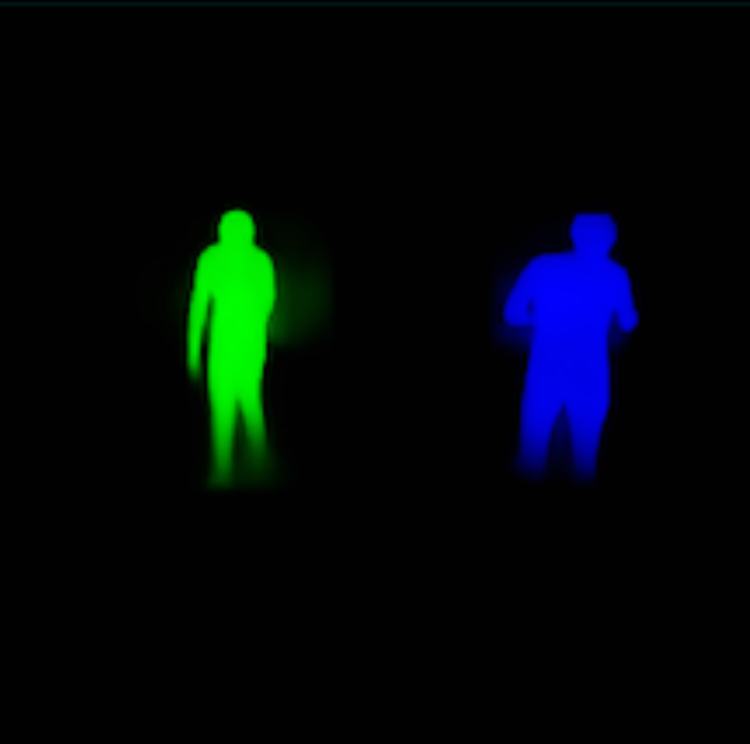} \\ 
{\small (a) Input} & {\small (b) Detection }& {\small (c) Segmentation}\\\\
  \end{tabular}
\vspace{-4mm}
  \caption{\textbf{Multi-person detection and segmentation results} generated by sampling our model multiple times. As the model is trained on single persons this only works for non-intersecting cases.}
  \label{fig:multi_person}
\end{figure}

\parag{Multiple people.} Although our focus is on handling single objects or persons, our probabilistic framework can handle several at test time by sampling more than once. Fig.~\ref{fig:multi_person} shows the predicted cell probability as blue dots whose size is proportional to the probability. The fully-convolutional architecture operates locally and thereby predicts a high person probability close to both subjects. As a result, both the detection and segmentation results remain accurate as long as the individuals are sufficiently separated. Note that the model used for this experiment was still trained on single subjects. In future work, we will attempt self-supervised training of multiple interacting people, which has so far only been established in controlled environments.

\begin{figure}
  \centering
  \begin{tabular}{@{}ccc@{}}

   \includegraphics[width=0.16\textwidth]{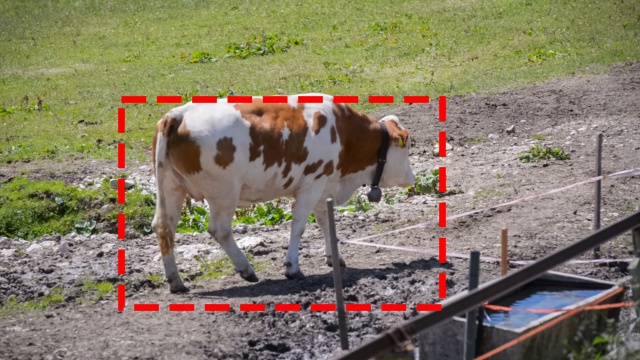} &\hspace{-4mm}
   \includegraphics[width=0.16\textwidth]{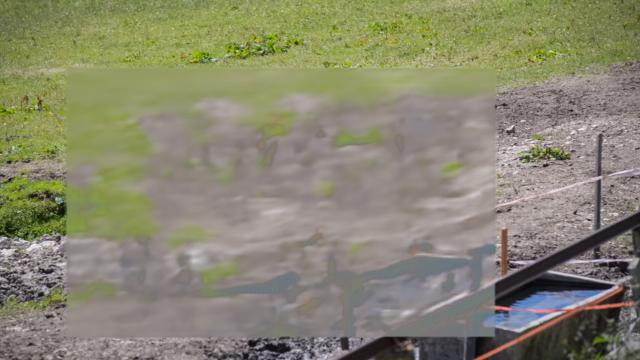} &\hspace{-4mm}
   \includegraphics[width=0.16\textwidth]{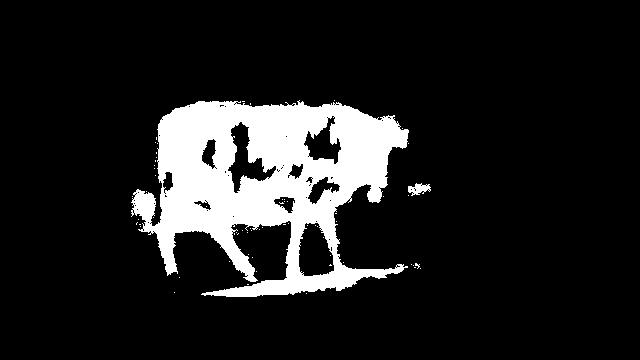} \\
\includegraphics[width=0.16\textwidth]{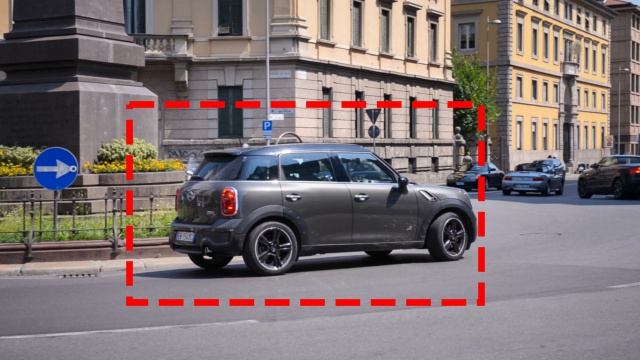} &\hspace{-4mm}
\includegraphics[width=0.16\textwidth]{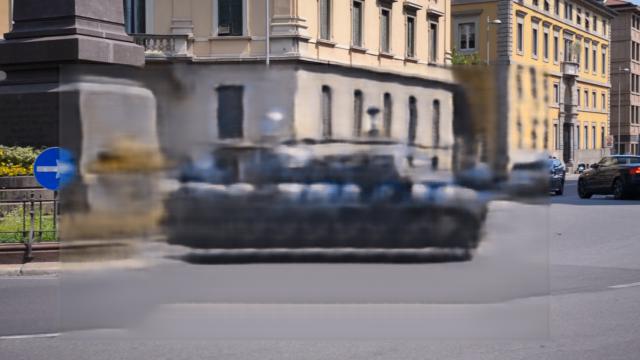} &\hspace{-4mm}
  \includegraphics[width=0.16\textwidth]{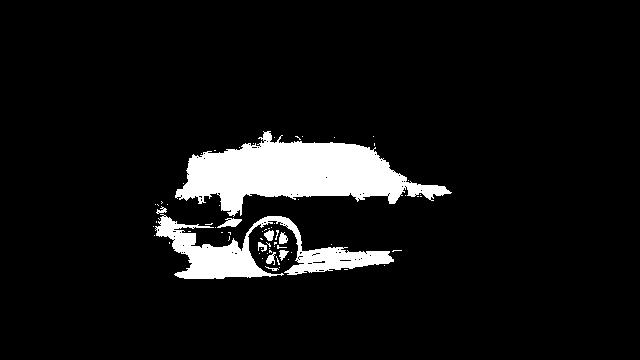}\\
  \includegraphics[width=0.16\textwidth]{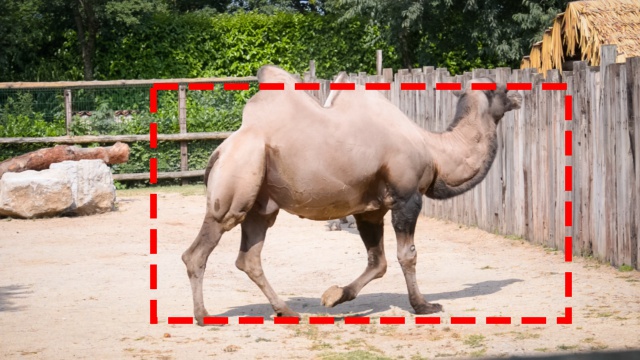} &\hspace{-4mm}
  \includegraphics[width=0.16\textwidth]{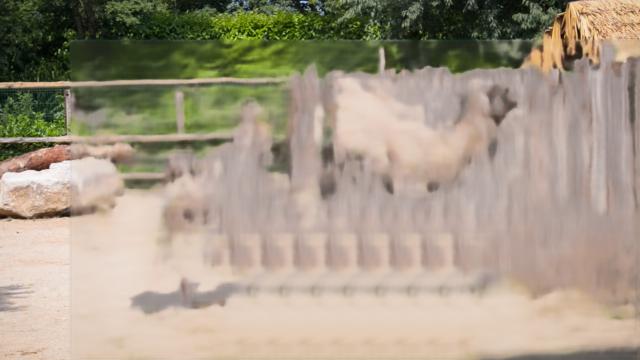} &\hspace{-4mm}
  \includegraphics[width=0.16\textwidth]{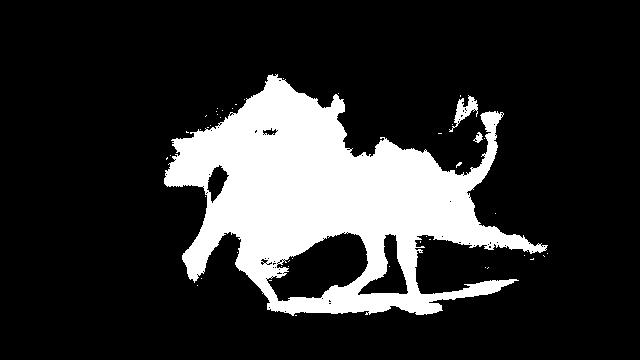} \\ \\
  
  {\small (a)Input/Ours } & {\small (b)Inpainting result} & {\small (c)Segmentation }
  \end{tabular}

  \caption{\textbf{Examples of qualitative results on DAVIS2016~\cite{Perazzi16} validation sequences.} Top row: successful segmentation of the moving object. Middle row: partially successful case in which the background scene carries information about the moving object's location. Bottom row: poor segmentation result that occurs when our inpainting network can resonstruct certain parts of the moving object due to its slow motion.}
  \label{fig:davis16}
\end{figure}

\parag{Other object categories.} In this section, we investigate the applicability of our method to standard benchmarks with other object categories. The existing object detection datasets SegTrackV2 and FBMS59 comprise multiple objects, which we do not support. Therefore, we demonstrate the qualitative performance of our method on the standard DAVIS2016~\cite{Perazzi16} benchmark that consists of various object categories such as car, cow and goat. DAVIS contains 30 training and 20 testing sequences, which are very short compared to other benchmarks suitable for deep-learning based methods. We follow the standard procedure and use the validation sequences for evaluation. Since our method does not require any annotations, we train and test on the validation sequences with an average of 70 frames per video. So far, we have evaluated our method on datasets with human subjects. Therefore we pick non-human object categories in the DAVIS2016 validation dataset to show that our method is not specific to a particular object type. As shown in Fig.~\ref{fig:davis16}, our performance on DAVIS2016 varies, depending on the length and footage of the sequence. However, we do not expect our method to compete with approaches tuned for short video snippets. Many of these short sequences include objects that move slowly, remaining mostly in the same image region. This makes them easy to inpaint, thus violating our assumptions A1 and A2 (Section~\ref{sec:separation}). Fig.~\ref{fig:davis16}(top) shows a successful segmentation result on a longer sequence with a moving object. Fig.~\ref{fig:davis16}(middle) illustrates a partially successful case that occurs when the location of the object changes with the background elements and the content of the scene in a short video clip provides significant clue about the reconstruction of the foreground object. In Fig.~\ref{fig:davis16}(bottom), we present a failure case that occurs when our method is applied to very short videos with negligible object displacement. In this case, our inpainting network can reconstruct the foreground object together with the background region, which causes holes in the regions of the segmentation mask that are already reconstructed by the inpainting network. 

In short, DAVIS2016 features only few videos per category, with each video being short, making them ill-suited to deep-learning based self-supervised approaches that exploit large unlabeled video collections. By contrast, we contribute new benchmarks with manual annotations for quantitative evaluation and three very different settings with significantly more and longer training videos that can be used to evaluate future self-supervised segmentation methods.

\section{Conclusion}

We have proposed a self-supervised method for object detection and segmentation that lends itself for application in domains where general purpose detectors fail. Our core contributions are the Monte Carlo-based optimization of proposal-based detection, new foreground and background objectives, and their joint training on unlabeled videos captured by static, rotating and handheld cameras. Our experiments demonstrate that, even if trained only on single persons, our approach generalizes to multi-person detection, as long as the persons are sufficiently separated.
In contrast to many existing solutions~\cite{Barnich11,Russell14,Stretcu15,Koh17b}, our
approach does not exploit temporal cues at test time. In the future, we will integrate temporal dependencies explicitly, which will facilitate addressing the scenario where multiple people interact closely, by incorporating physics-inspired constraints enforcing plausible motion.

\clearpage

\bibliographystyle{IEEEtran}
\bibliography{string,vision}

\begin{thebibliography}{10}
\providecommand{\url}[1]{#1}
\csname url@samestyle\endcsname
\providecommand{\newblock}{\relax}
\providecommand{\bibinfo}[2]{#2}
\providecommand{\BIBentrySTDinterwordspacing}{\spaceskip=0pt\relax}
\providecommand{\BIBentryALTinterwordstretchfactor}{4}
\providecommand{\BIBentryALTinterwordspacing}{\spaceskip=\fontdimen2\font plus
\BIBentryALTinterwordstretchfactor\fontdimen3\font minus
  \fontdimen4\font\relax}
\providecommand{\BIBforeignlanguage}[2]{{%
\expandafter\ifx\csname l@#1\endcsname\relax
\typeout{** WARNING: IEEEtran.bst: No hyphenation pattern has been}%
\typeout{** loaded for the language `#1'. Using the pattern for}%
\typeout{** the default language instead.}%
\else
\language=\csname l@#1\endcsname
\fi
#2}}
\providecommand{\BIBdecl}{\relax}
\BIBdecl

\bibitem{Lin14a}
T.-Y. Lin, M.~Maire, S.~Belongie, J.~Hays, P.~Perona, D.~Ramanan,
  P.~Doll{\'a}r, and C.~Zitnick, ``{Microsoft COCO: Common Objects in
  Context},'' in \emph{European Conference on Computer Vision}, 2014, pp.
  740--755.

\bibitem{He17a}
K.~He, G.~Gkioxari, P.~Doll{\'a}r, and R.~B. Girshick, ``{Mask R-CNN},'' in
  \emph{International Conference on Computer Vision}, 2017.

\bibitem{Bielski19}
A.~Bielski and P.~Favaro, ``{Emergence of Object Segmentation in Perturbed
  Generative Models},'' in \emph{Advances in Neural Information Processing
  Systems}, 2019.

\bibitem{Crawford19}
E.~Crawford and J.~Pineau, ``{Spatially Invariant Unsupervised Object Detection
  with Convolutional Neural Networks},'' in \emph{Conference on Artificial
  Intelligence}, 2019.

\bibitem{Croitoru19}
I.~Croitoru, S.~V. Bogolin, and M.~Leordeanu, ``{Unsupervised Learning of
  Foreground Object Segmentation},'' \emph{International Journal of Computer
  Vision}, vol. 127, pp. 1279--1302, 2019.

\bibitem{Eslami16}
S.~Eslami, N.~Heess, T.~Weber, Y.~Tassa, D.~Szepesvari, K.~Kavukcuoglu, and
  G.~Hinton, ``{Attend, Infer, Repeat: Fast Scene Understanding with Generative
  Models},'' in \emph{Advances in Neural Information Processing Systems}, 2016.

\bibitem{Rhodin19a}
H.~Rhodin, V.~Constantin, I.~Katircioglu, M.~Salzmann, and P.~Fua, ``{Neural
  Scene Decomposition for Human Motion Capture},'' in \emph{Conference on
  Computer Vision and Pattern Recognition}, 2019.

\bibitem{Redmon16}
J.~Redmon, S.~Divvala, R.~Girshick, and A.~Farhadi, ``{You Only Look Once:
  Unified, Real-Time Object Detection},'' in \emph{Conference on Computer
  Vision and Pattern Recognition}, 2016.

\bibitem{Rhodin18a}
H.~Rhodin, J.~Spoerri, I.~Katircioglu, V.~Constantin, F.~Meyer, E.~Moeller,
  M.~Salzmann, and P.~Fua, ``{Learning Monocular 3D Human Pose Estimation from
  Multi-View Images},'' in \emph{Conference on Computer Vision and Pattern
  Recognition}, 2018.

\bibitem{Cheng17a}
J.~Cheng, Y.~H. Tsai, S.~Wang, and M.~H. Yang, ``{Segflow: Joint Learning for
  Video Object Segmentation and Optical Flow},'' in \emph{International
  Conference on Computer Vision}, 2017.

\bibitem{Song18}
H.~Song, W.~Wang, S.~Zhao, J.~Shen, and K.-M. Lam, ``{Pyramid Dilated Deeper
  ConvLSTM for Video Salient Object Detection},'' in \emph{European Conference
  on Computer Vision}, 2018.

\bibitem{Perazzi15}
F.~Perazzi, O.~Wang, M.~Gross, and A.~S.-Hornung, ``{Fully Connected Object
  Proposals for Video Segmentation},'' in \emph{International Conference on
  Computer Vision}, 2015.

\bibitem{Hu18b}
Y.~T. Hu, J.~B. Huang, and A.~G. Schwing, ``{Unsupervised Video Object
  Segmentation Using Motion Saliency-Guided Spatio-Temporal Propagation},'' in
  \emph{European Conference on Computer Vision}, 2018.

\bibitem{Jain17}
S.~D. Jain, B.~Xiong, and K.~Grauman, ``{Fusionseg: Learning to Combine Motion
  and Appearance for Fully Automatic Segmentation of Generic Objects in
  Videos},'' in \emph{Conference on Computer Vision and Pattern Recognition},
  2017.

\bibitem{Li18g}
S.~Li, B.~Seybold, A.~Vorobyov, A.~Fathi, Q.~Huang, and C.-C.~J. Kuo,
  ``{Instance Embedding Transfer to Unsupervised Video Object Segmentation},''
  in \emph{Conference on Computer Vision and Pattern Recognition}, 2018.

\bibitem{Li18j}
S.~Li, B.~Seybold, A.~Vorobyov, X.~Lei, and C.-C.~J. Kuo, ``{Unsupervised Video
  Object Segmentation with Motion-Based Bilateral Networks},'' in
  \emph{European Conference on Computer Vision}, 2018.

\bibitem{Lu19}
X.~Lu, W.~Wang, C.~Ma, J.~Shen, L.~Shao, and F.~Porikli, ``{See More, Know
  More: Unsupervised Video Object Segmentation with Co-Attention Siamese
  Networks},'' in \emph{Conference on Computer Vision and Pattern Recognition},
  2019.

\bibitem{Yang19b}
Z.~Yang, Q.~Wang, L.~Bertinetto, W.~Hu, S.~Bai, and P.~H.~S. Torr, ``{Anchor
  Diffusion for Unsupervised Video Object Segmentation},'' in
  \emph{International Conference on Computer Vision}, 2019.

\bibitem{Wang19g}
X.~Wang, A.~Jabri, and A.~Efros, ``{Learning Correspondence from the
  Cycle-Consistency of Time},'' in \emph{Conference on Computer Vision and
  Pattern Recognition}, 2019.

\bibitem{Koh17b}
Y.~J. Koh and C.-S. Kim, ``{Primary Object Segmentation in Videos Based on
  Region Augmentation and Reduction},'' in \emph{Conference on Computer Vision
  and Pattern Recognition}, 2017.

\bibitem{Cho15}
M.~Cho, S.~Kwak, C.~Schmid, and J.~Ponce, ``{Unsupervised Object Discovery and
  Localization in the Wild: Part-Based Matching with Bottom-Up Region
  Proposals},'' in \emph{Conference on Computer Vision and Pattern
  Recognition}, 2015.

\bibitem{Wei17}
X.-S. Wei, C.-L. Zhang, J.~Wu, C.~Shen, and Z.-H. Zhou, ``{Unsupervised Object
  Discovery and Co-Localization by Deep Descriptor Transforming},'' in
  \emph{arXiv Preprint}, 2017.

\bibitem{Tokmakov17b}
P.~Tokmakov, K.~Alahari, and C.~Schmid, ``{Learning Motion Patterns in
  Videos},'' in \emph{Conference on Computer Vision and Pattern Recognition},
  2017.

\bibitem{Tokmakov17a}
------, ``{Learning Video Object Segmentation with Visual Memory},'' in
  \emph{International Conference on Computer Vision}, 2017.

\bibitem{Lee11}
Y.~Lee, J.~Kim, and A.~K. Grauman, ``{Key-Segments for Video Object
  Segmentation},'' in \emph{International Conference on Computer Vision}, 2011.

\bibitem{Papazoglou13}
A.~Papazoglou and V.~Ferrari, ``{Fast Object Segmentation in Unconstrained
  Video},'' in \emph{International Conference on Computer Vision}, 2013, pp.
  1777--1784.

\bibitem{Factor14}
A.~Faktor and M.~Irani, ``{Video Segmentation by Non-Local Consensus Voting},''
  in \emph{British Machine Vision Conference}, 2014.

\bibitem{Wang15d}
W.~Wang, J.~Shen, and F.~Porikli, ``{Saliency-Aware Geodesic Video Object
  Segmentation},'' in \emph{Conference on Computer Vision and Pattern
  Recognition}, 2015.

\bibitem{Barnich11}
O.~Barnich and M.~V. Droogenbroeck, ``{Vibe: A Universal Background Subtraction
  Algorithm for Video Sequences},'' \emph{IEEE Transactions on Image
  processing}, vol.~20, no.~6, pp. 1709--1724, 2011.

\bibitem{Stretcu15}
O.~Stretcu and M.~Leordeanu, ``{Multiple Frames Matching for Object Discovery
  in Video},'' in \emph{British Machine Vision Conference}, 2015.

\bibitem{Russell14}
C.~Russell, R.~Yu, and L.~Agapito, ``{Video Pop-Up: Monocular 3D Reconstruction
  of Dynamic Scenes},'' in \emph{European Conference on Computer Vision}, 2014.

\bibitem{Lu20}
X.~Lu, W.~Wang, J.~Shen, Y.~Tai, D.~Crandall, and S.~Hoi, ``{Learning Video
  Object Segmentation from Unlabeled Videos},'' in \emph{Conference on Computer
  Vision and Pattern Recognition}, 2020.

\bibitem{Yang19c}
Y.~Yang, A.~Loquercio, D.~Scaramuzza, and S.~Soatto, ``{Unsupervised Moving
  Object Detection via Contextual Information Separation},'' in
  \emph{Conference on Computer Vision and Pattern Recognition}, 2019.

\bibitem{Sun18a}
D.~Sun, X.~Yang, M.~Liu, and J.~Kautz, ``{Pwc-Net: CNNs for Optical Flow Using
  Pyramid, Warping, and Cost Volume},'' in \emph{Conference on Computer Vision
  and Pattern Recognition}, 2018.

\bibitem{Jaderberg15}
M.~Jaderberg, K.~Simonyan, A.~Zisserman, and K.~Kavukcuoglu, ``{Spatial
  Transformer Networks},'' in \emph{Advances in Neural Information Processing
  Systems}, 2015, pp. 2017--2025.

\bibitem{Pathak17}
D.~Pathak, R.~Girshick, P.~Doll\'{a}r, T.~Darrell, and B.~Hariharan,
  ``{Learning Features by Watching Objects Move},'' in \emph{Conference on
  Computer Vision and Pattern Recognition}, 2017.

\bibitem{Chen19a}
M.~Chen, T.~Artieres, and L.~Denoyer, ``{Unsupervised Object Segmentation by
  Redrawing},'' in \emph{Advances in Neural Information Processing Systems},
  2019.

\bibitem{Arandjelovic19}
R.~Arandjelovic and A.~Zisserman, ``{Object Discovery with a Copy-Pasting
  GAN},'' in \emph{arXiv Preprint}, 2019.

\bibitem{Baque17b}
P.~Baqu{\'e}, F.~Fleuret, and P.~Fua, ``{Deep Occlusion Reasoning for
  Multi-Camera Multi-Target Detection},'' in \emph{International Conference on
  Computer Vision}, 2017.

\bibitem{Pathak16}
D.~Pathak, P.~Kr{\"a}henb{\"u}hl, J.~Donahue, T.~Darrell, and A.~A. Efros,
  ``{Context Encoders: Feature Learning by Inpainting},'' in \emph{Conference
  on Computer Vision and Pattern Recognition}, 2016.

\bibitem{Yu2018}
J.~Yu, Z.~Lin, J.~Yang, X.~Shen, X.~Lu, and T.~S. Huang, ``{Generative Image
  Inpainting with Contextual Attention},'' in \emph{Conference on Computer
  Vision and Pattern Recognition}, 2018.

\bibitem{Glynn90}
P.~M. Glynn, ``{Likelihood Ratio Gradient Estimation for Stochastic Systems},''
  \emph{Communications of the ACM}, vol.~33, no.~10, pp. 75--84, 1990.

\bibitem{Williams92}
R.~Williams, ``{Simple Statistical Gradient-Following Algorithms for
  Connectionist Reinforcement Learning},'' \emph{Machine Learning}, 1992.

\bibitem{Ilg17}
E.~Ilg, N.~Mayer, T.~Saikia, M.~Keuper, A.~Dosovitskiy, and T.~Brox, ``{Flownet
  2.0: Evolution of Optical Flow Estimation with Deep Networks},'' in
  \emph{Conference on Computer Vision and Pattern Recognition}, 2017.

\bibitem{Kraehenbuehl11}
P.~Kr{\"a}henb{\"u}hl and V.~Koltun, ``{Efficient Inference in Fully Connected
  CRFs with Gaussian Edge Potentials},'' in \emph{Advances in Neural
  Information Processing Systems}, 2011.

\bibitem{Perazzi16}
F.~Perazzi, J.~Pont-Tuset, B.~McWilliams, L.~V. Gool, M.~Gross, and
  A.~Sorkine-Hornung, ``{A Benchmark Dataset and Evaluation Methodology for
  Video Object Segmentation},'' in \emph{Conference on Computer Vision and
  Pattern Recognition}, 2016.

\bibitem{Wu18c}
Z.~Wu, Y.~Xiong, X.~Y. Stella, and D.~Lin, ``{Unsupervised Feature Learning via
  Non-Parametric Instance Discrimination},'' in \emph{Conference on Computer
  Vision and Pattern Recognition}, 2018.

\bibitem{Ionescu14a}
C.~Ionescu, I.~Papava, V.~Olaru, and C.~Sminchisescu, ``{{Human3.6M}: Large
  Scale Datasets and Predictive Methods for 3D Human Sensing in Natural
  Environments},'' \emph{IEEE Transactions on Pattern Analysis and Machine
  Intelligence}, 2014.

\bibitem{Jang16}
E.~Jang, S.~Gu, and B.~Poole, ``{Categorical Reparameterization with
  Gumbel-Softmax},'' in \emph{International Conference on Learning
  Representations}, 2017.

\end{thebibliography}

\end{document}